\documentclass[10pt,twocolumn,letterpaper]{article}

\usepackage{iccv}
\usepackage{times}
\usepackage{epsfig}
\usepackage{url}            
\usepackage{graphbox}
\usepackage{booktabs}       
\usepackage{amsfonts}       
\usepackage{nicefrac}       
\usepackage{microtype}      
\usepackage{comment}
\usepackage{amsmath,amssymb} 
\usepackage[font=small]{caption}
\usepackage{mathtools}
\usepackage{graphicx}
\usepackage{soul}
\usepackage[subrefformat=parens]{subfig}
\usepackage{enumitem}

\usepackage{xspace}
\usepackage{multirow}
\usepackage{array}
\usepackage{layouts}
\usepackage{algorithm}
\usepackage{bbding}
\usepackage{listings}
\usepackage{pifont}
\usepackage{wasysym}
\usepackage{float}
\usepackage{soul}
\usepackage{footnote}
\makesavenoteenv{tabular}
\makesavenoteenv{table}
\makesavenoteenv{figure}
\usepackage[dvipsnames]{xcolor}
\usepackage{pythonhighlight}

\usepackage{xcolor,colortbl}
\definecolor{Gray}{gray}{0.90}
\newcolumntype{g}{>{\columncolor{Gray}}c}
\definecolor{ffe1da}{RGB}{255,225,218}
\definecolor{F7E0D5}{RGB}{247,224,213}
\definecolor{darkF7E0D5}{RGB}{209,154,128}
\colorlet{Light}{White!0!F7E0D5}

\usepackage[pagebackref=true,breaklinks=true,letterpaper=true,colorlinks,bookmarks=false]{hyperref}
\usepackage{cleveref}

\newcommand\blfootnote[1]{%
  \begingroup
  \renewcommand\thefootnote{}\footnote{#1}%
  \addtocounter{footnote}{-1}%
  \endgroup
}

\iccvfinalcopy 

\def\iccvPaperID{7530} 

\ificcvfinal\fi

\def \pzo {\phantom{0}} 

\newcommand{\xmark}{\ding{55}}
\newcommand{\gray}[1]{\textcolor{gray}{#1}}
\def\OURS{DINO\xspace}

\newcommand{\rownumber}[1]{\textcolor{darkF7E0D5}{#1}}

\begin{document}

\title{Emerging Properties in Self-Supervised Vision Transformers}

\author{
Mathilde Caron$^{1,2}$~~~~~~~Hugo Touvron$^{1,3}$~~~~~~~Ishan Misra$^{1}$~~~~~~~Herv\'e Jegou$^{1}$\\
Julien Mairal$^{2}$~~~~~~~Piotr Bojanowski$^{1}$~~~~~~~Armand Joulin$^{1}$
\vspace{1em}
\\
$^1$ Facebook AI Research~~~~~~~~~~$^2$ Inria$^*$~~~~~~~~~~$^3$ Sorbonne University
}

\ificcvfinal\thispagestyle{empty}\fi


\twocolumn[{%
\renewcommand\twocolumn[1][]{#1}%
\maketitle
\begin{center}
  \centering
  \captionsetup{type=figure}
  \includegraphics[width=\linewidth]{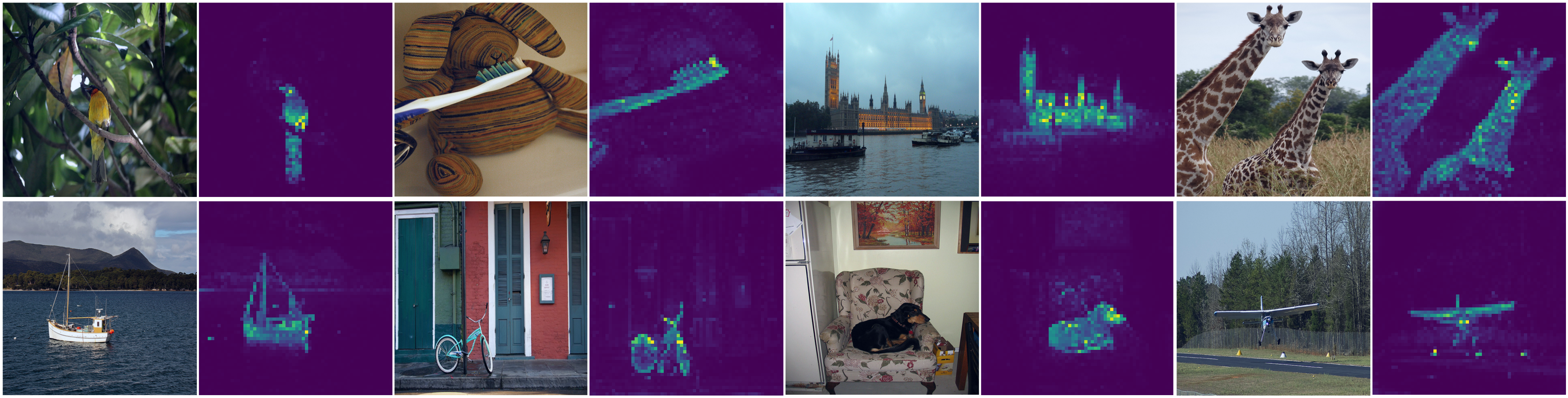}
  \captionof{figure}{
    \textbf{Self-attention from a Vision Transformer with $8\times 8$ patches trained with no supervision.}
    We look at the self-attention of the \texttt{[CLS]} token on the heads of the last layer. 
    This token is not attached to any label nor supervision.
    These maps show that the model automatically learns class-specific features leading to unsupervised object segmentations.
  }
  \label{fig:att}
\end{center}%
}]

\begin{abstract}
\blfootnote{
$^*$Univ. Grenoble Alpes, Inria, CNRS, Grenoble INP, LJK, 38000 Grenoble, France. \\
Correspondence: mathilde@fb.com \\
Code: \url{https://github.com/facebookresearch/dino}
}
In this paper, we question if self-supervised learning provides new properties to Vision Transformer~(ViT)~\cite{dosovitskiy2020image} that stand out compared to convolutional networks (convnets).
Beyond the fact that adapting self-supervised methods to this architecture works particularly well, we make the following observations:
first, self-supervised ViT features contain explicit information about the semantic segmentation of an image,
which does not emerge as clearly with supervised ViTs, nor with convnets.
Second, these features are also excellent $k$-NN classifiers, reaching 78.3\% top-1 on ImageNet with a small ViT. 
Our study also underlines the importance of momentum encoder~\cite{he2020momentum}, multi-crop training~\cite{caron2020unsupervised}, and the use of small patches with ViTs.
We implement our findings into a simple self-supervised method, called \OURS, which we interpret as a form of self-\textbf{di}stillation with \textbf{no} labels.
We show the synergy between \OURS and ViTs by achieving 80.1\% top-1 on ImageNet in linear evaluation with ViT-Base.



   
\end{abstract}



\section{Introduction}
Transformers~\cite{vaswani2017attention} have recently emerged as an alternative to convolutional neural networks (convnets) for visual recognition~\cite{dosovitskiy2020image,touvron2020training,zhao2020exploring}.
Their adoption has been coupled with a training strategy inspired by natural language processing (NLP), that is, pretraining on large quantities of data and finetuning on the target dataset~\cite{devlin2018bert,radford2019language}. 
The resulting Vision Transformers~(ViT)~\cite{dosovitskiy2020image} are competitive with convnets but, they have not yet delivered clear benefits over them: 
they are computationally more demanding, require more training data, and their features do not exhibit unique properties. 

In this paper, we question whether the muted success of Transformers in vision can be explained by the use of supervision in their pretraining.
Our motivation is that one of the main ingredients for the success of Transformers in NLP was the use of self-supervised pretraining, in the form of close procedure in BERT~\cite{devlin2018bert} or language modeling in GPT~\cite{radford2019language}.
These self-supervised pretraining objectives use the words in a sentence to create pretext tasks that provide a richer learning signal than the supervised objective of predicting a single label per sentence.
Similarly, in images, image-level supervision often reduces the rich visual information contained in an image to a single concept selected from a predefined set of a few thousand categories of objects~\cite{russakovsky2015imagenet}.

While the self-supervised pretext tasks used in NLP are text specific, many existing self-supervised methods have shown their potential on images with convnets~\cite{caron2020unsupervised, chen2020simple,grill2020bootstrap,he2020momentum}.
They typically share a similar structure but with different components designed to avoid trivial solutions (collapse) or to improve performance~\cite{chen2020exploring}.
In this work, inspired from these methods, we study the impact of self-supervised pretraining on ViT features.
Of particular interest, we have identified several interesting properties that do not emerge with supervised ViTs, nor with convnets:

\begin{itemize}
\item
Self-supervised ViT features explicitly contain the scene layout and, in particular, object boundaries, as shown in Figure~\ref{fig:att}.
This information is directly accessible in the self-attention modules of the last block.
\item
Self-supervised ViT features perform particularly well with a basic nearest neighbors classifier ($k$-NN)~\emph{without any finetuning, linear classifier nor data augmentation}, achieving 78.3\% top-1 accuracy on ImageNet.
\end{itemize}

The emergence of segmentation masks seems to be a property shared across self-supervised methods.
However, the good performance with $k$-NN only emerge when combining certain components such as momentum encoder~\cite{he2020momentum} and multi-crop augmentation~\cite{caron2020unsupervised}.
Another finding from our study is the importance of using smaller patches with ViTs to improve the quality of the resulting features. 

Overall, our findings about the importance of these components lead us to design a simple self-supervised approach that can be interpreted as a form of knowledge \textbf{di}stillation~\cite{hinton2015distilling} with \textbf{no} labels.
The resulting framework, \OURS, simplifies self-supervised training by directly predicting the output of a teacher network---built with a momentum encoder---by using a standard cross-entropy loss.
Interestingly, our method can work with only a centering and sharpening of the teacher output to avoid collapse, while
other popular components such as predictor~\cite{grill2020bootstrap}, advanced normalization~\cite{caron2020unsupervised} or contrastive loss~\cite{he2020momentum} add little benefits in terms of stability or performance.
Of particular importance, our framework is flexible and works on both convnets and ViTs without the need to modify the architecture, nor adapt internal normalizations~\cite{richemond2020byol}. 

We further validate the synergy between \OURS and ViT by outperforming previous self-supervised features on the ImageNet linear classification benchmark with 80.1\% top-1 accuracy with a ViT-Base with small patches.
We also confirm that \OURS works with convnets by matching the state of the art with a ResNet-50 architecture.
Finally, we discuss different scenarios to use \OURS with ViTs in case of limited computation and memory capacity. 
In particular, training \OURS with ViT takes just two 8-GPU servers over 3 days to achieve $76.1\%$ on ImageNet linear benchmark, which outperforms self-supervised systems based on convnets of comparable sizes with significantly reduced compute requirements~\cite{caron2020unsupervised,grill2020bootstrap}.

\begin{figure}[t]
\centering
 \includegraphics[width=0.6\linewidth]{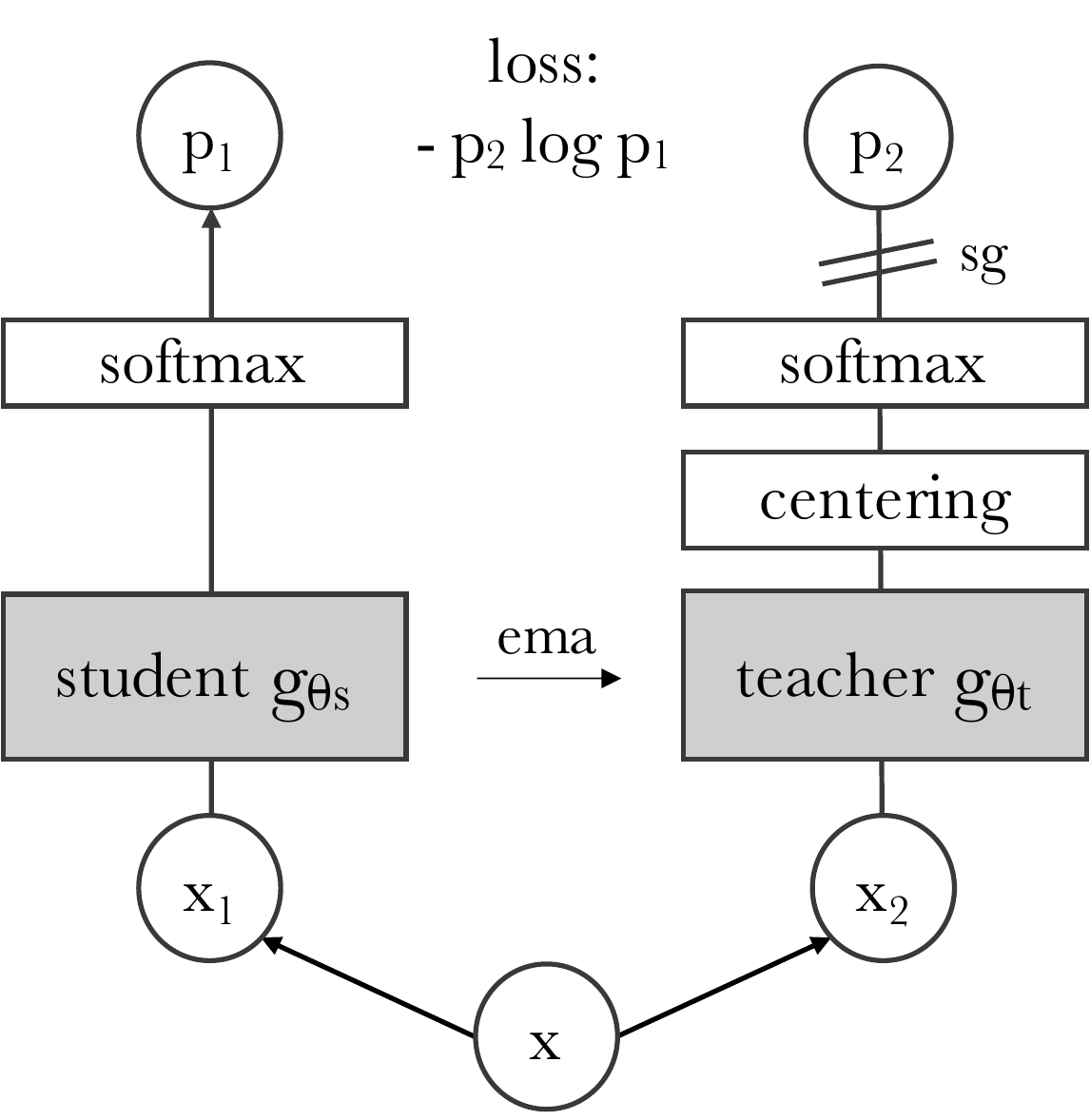}
\caption{
\textbf{Self-distillation with no labels.}
We illustrate \OURS in the case of one single pair of views ($x_1$, $x_2$) for simplicity.
The model passes two different random transformations of an input image to the student and teacher networks.
Both networks have the same architecture but different parameters. 
The output of the teacher network is centered with a mean computed over the batch.
Each networks outputs a $K$ dimensional feature that is normalized with a temperature softmax over the feature dimension.
Their similarity is then measured with a cross-entropy loss.
We apply a stop-gradient (sg) operator on the teacher to propagate gradients only through the student.
The teacher parameters are updated with an exponential moving average (ema) of the student parameters.
}
\vspace{-1em}
\label{fig:model}
\end{figure}

\section{Related work}


\paragraph{Self-supervised learning.}
A large body of work on self-supervised learning focuses on discriminative approaches coined \emph{instance classification}~\cite{chen2020simple,dosovitskiy2016discriminative,he2020momentum,wu2018unsupervised},
which considers each image a different class and trains the model by discriminating them up to data augmentations.
However, explicitly learning a classifier to discriminate between all images~\cite{dosovitskiy2016discriminative} does not scale well with the number of images. 
Wu~\etal~\cite{wu2018unsupervised} propose to use a noise contrastive estimator (NCE)~\cite{gutmann2010noise} to compare instances instead of classifying them.
A caveat of this approach is that it requires comparing features from a large number of images simultaneously.
In practice, this requires large batches~\cite{chen2020simple} or memory banks~\cite{he2020momentum, wu2018unsupervised}.
Several variants allow automatic grouping of instances in the form of clustering~\cite{asano2019self,caron2018deep,caron2019unsupervised,huang2019unsupervised,junnan2021prototypical,xie2016unsupervised,yang2016joint,zhuang2019local}.

Recent works have shown that we can learn unsupervised features without discriminating between images.
Of particular interest, Grill~\etal~\cite{grill2020bootstrap} propose a metric-learning formulation called BYOL, where features are trained by matching them to representations obtained with a momentum encoder.
Methods like BYOL work even without a momentum encoder, at the cost of a drop of performance~\cite{chen2020exploring,grill2020bootstrap}.
Several other works echo this direction, showing that one can match more elaborate representations~\cite{gidaris2020learning,gidaris2020obow}, train features matching them to a uniform distribution~\cite{bojanowski2017unsupervised} or by using whitening~\cite{ermolov2020whitening,zbontar2021barlow}.
Our approach takes its inspiration from BYOL but operates with a different similarity matching loss and uses the exact same architecture for the student and the teacher.
That way, our work completes the interpretation initiated in BYOL of self-supervised learning as a form of Mean Teacher self-distillation~\cite{tarvainen2017mean} with no labels.

\paragraph{Self-training and knowledge distillation.}
Self-training aims at improving the quality of features by propagating a small initial set of annotations to a large set of unlabeled instances.
This propagation can either be done with hard assignments of labels~\cite{lee2013pseudo,xu2020iterative,yalniz2019billion} or with a soft assignment~\cite{xie2020self}.
When using soft labels, the approach is often referred to as knowledge distillation~\cite{bucilua2006model,hinton2015distilling} and has been primarily designed to train a small network to mimic the output of a larger network to compress models.
Xie~\etal~\cite{xie2020self} have shown that distillation could be used to propagate soft pseudo-labels to unlabelled data in a self-training pipeline, drawing an essential connection between self-training and knowledge distillation.
Our work builds on this relation and extends knowledge distillation to the case where no labels are available.
Previous works have also combined self-supervised learning and knowledge distillation~\cite{fang2021seed,shen2021s2,chen2020big,noroozi2018boosting}, enabling self-supervised model compression and performance gains.
However, these works rely on a \emph{pre-trained} fixed teacher while our teacher is dynamically built during training.
This way, knowledge distillation, instead of being used as a post-processing step to self-supervised pre-training, is directly cast as a self-supervised objective.
Finally, our work is also related to \textit{codistillation}~\cite{anil2018large} where student and teacher have the same architecture and use distillation during training.
However, the teacher in \textit{codistillation} is also distilling from the student, while it is updated with an average of the student in our work.

\section{Approach}
\subsection{SSL with Knowledge Distillation}
\label{sec:method}

The framework used for this work, \OURS, shares the same overall structure as recent self-supervised approaches~\cite{caron2020unsupervised,chen2020exploring,chen2020simple,grill2020bootstrap,he2020momentum}.
However, our method shares also similarities with knowledge distillation~\cite{hinton2015distilling} and we present it under this angle.
We illustrate \OURS in Figure~\ref{fig:model} and propose a pseudo-code implementation in Algorithm~\ref{algo:DINO}.

Knowledge distillation is a learning paradigm where we train a student network $g_{\theta_s}$ to match the output of a given teacher network $g_{\theta_t}$, parameterized by $\theta_s$ and $\theta_t$ respectively.
Given an input image $x$, both networks output probability distributions over $K$ dimensions denoted by $P_s$ and $P_t$.
The probability $P$ is obtained by normalizing the output of the network $g$ with a softmax function. More precisely,
\begin{equation}
  P_s(x)^{(i)} = \frac{\exp(g_{\theta_s}(x)^{(i)} / \tau_s)}{\sum_{k=1}^K \exp(g_{\theta_s}(x)^{(k)} / \tau_s)},
\end{equation}                                                                                                                                                                                        
with $\tau_s>0$ a temperature parameter that controls the sharpness of the output distribution, and a similar formula holds for~$P_t$ with temperature $\tau_t$.
Given a fixed teacher network~$g_{\theta_t}$, we learn to match these distributions by minimizing the cross-entropy loss w.r.t. the parameters of the student network $\theta_s$:
\begin{equation}
        \min_{\theta_s} H(P_t(x), P_s(x)),
  \label{eq:kd}                                                                                                                                                                                       
\end{equation}
where $H(a, b) = - a \log b$.

In the following, we detail how we adapt the problem in Eq.~(\ref{eq:kd}) to self-supervised learning.
First, we construct different distorted views, or crops, of an image with multi-crop strategy~\cite{caron2020unsupervised}.
More precisely, from a given image, we generate a set $V$ of different views.
This set contains two \emph{global} views, $x^{g}_1$ and $x^{g}_2$ and several \emph{local} views of smaller resolution.
All crops are passed through the student while only the \emph{global} views are passed through the teacher, therefore encouraging ``local-to-global'' correspondences.
We minimize the loss:
\begin{equation}
	\min_{\theta_s} \sum_{x \in \{x^{g}_1, x^{g}_2\}} \quad \sum_{\substack{x' \in V\\x'\neq \, x}} \quad H(P_t(x), P_s(x')).
\label{eq:loss}
\end{equation}

\begin{algorithm}[tb]
   \caption{\OURS PyTorch pseudocode w/o multi-crop.}
   \label{algo:DINO}
    \definecolor{codeblue}{rgb}{0.25,0.5,0.5}
    \lstset{
      basicstyle=\fontsize{7.2pt}{7.2pt}\ttfamily\bfseries,
      commentstyle=\fontsize{7.2pt}{7.2pt}\color{codeblue},
      keywordstyle=\fontsize{7.2pt}{7.2pt},
    }
\begin{lstlisting}[language=python]
# gs, gt: student and teacher networks
# C: center (K)
# tps, tpt: student and teacher temperatures
# l, m: network and center momentum rates
gt.params = gs.params
for x in loader: # load a minibatch x with n samples
    x1, x2 = augment(x), augment(x) # random views

    s1, s2 = gs(x1), gs(x2) # student output n-by-K
    t1, t2 = gt(x1), gt(x2) # teacher output n-by-K

    loss = H(t1, s2)/2 + H(t2, s1)/2
    loss.backward() # back-propagate

    # student, teacher and center updates
    update(gs) # SGD
    gt.params = l*gt.params + (1-l)*gs.params
    C = m*C + (1-m)*cat([t1, t2]).mean(dim=0)

def H(t, s):
    t = t.detach() # stop gradient
    s = softmax(s / tps, dim=1)
    t = softmax((t - C) / tpt, dim=1) # center + sharpen
    return - (t * log(s)).sum(dim=1).mean()


\end{lstlisting}
\end{algorithm}
This loss is general and can be used on any number of views, even only $2$.
However, we follow the standard setting for multi-crop by using 2 global views at resolution $224^2$ covering a large (for example greater than $50\%$) area of the original image, and several local views of resolution $96^2$ covering only small areas (for example less than $50\%$) of the original image.
We refer to this setting as the basic parametrization of \OURS, unless mentioned otherwise.

Both networks share the same architecture $g$ with different sets of parameters $\theta_s$ and $\theta_t$.
We learn the parameters $\theta_s$ by minimizing Eq.~(\ref{eq:loss}) with stochastic gradient descent.

\paragraph{Teacher network.}
Unlike knowledge distillation, we do not have a teacher~$g_{\theta_t}$ given \emph{a priori} and hence, we build it from past iterations of the student network.
We study different update rules for the teacher in Section~\ref{sec:teacher} and show that freezing the teacher network over an epoch works surprisingly well in our framework, while copying the student weight for the teacher fails to converge.
Of particular interest, using an exponential moving average (EMA) on the student weights, i.e., a momentum encoder~\cite{he2020momentum}, is particularly well suited for our framework. 
The update rule is $ \theta_t \leftarrow \lambda \theta_t + (1-\lambda) \theta_s,$ with $\lambda$ following a cosine schedule from $0.996$ to $1$ during training~\cite{grill2020bootstrap}.
Originally the momentum encoder has been introduced as a substitute for a queue in contrastive learning~\cite{he2020momentum}.
However, in our framework, its role differs since we do not have a queue nor a contrastive loss, and may be closer to the role of the mean teacher used in self-training~\cite{tarvainen2017mean}.
Indeed, we observe that this teacher performs a form of model ensembling similar to Polyak-Ruppert averaging with an exponential decay~\cite{polyak1992acceleration, ruppert1988efficient}.
Using Polyak-Ruppert averaging for model ensembling is a standard practice to improve the performance of a model~\cite{jean2014using}.
We observe that this teacher has better performance than the student throughout the training, and hence, guides the training of the student by providing target features of higher quality.
This dynamic was not observed in previous works~\cite{grill2020bootstrap, richemond2020byol}.
\begin{table}[t]
\centering
\caption{\textbf{Networks configuration.}
``Blocks'' is the number of Transformer blocks, ``dim'' is channel dimension and ``heads'' is the number of heads in multi-head attention.
``\# tokens'' is the length of the token sequence when considering $224^2$ resolution inputs, ``\# params'' is the total number of parameters (without counting the projection head) and ``im/s'' is the inference time on a NVIDIA V100 GPU with 128 samples per forward.
}
	\label{tab:archs}
  \setlength{\tabcolsep}{2pt}
  \begin{tabular}{@{}l c c c c c c@{}}
\toprule
	  model & blocks & dim & heads & \#tokens & \#params & im/s \\
    \midrule
	  ResNet-50 & -- & 2048 & -- & -- & 23M & 1237 \\
	  ViT-S/16 & 12 & 384 & 6 & 197 & 21M & 1007 \\
	  ViT-S/8 & 12 & 384 & 6 & 785 & 21M & 180 \\
	  ViT-B/16 & 12 & 768 & 12 & 197 & 85M & 312 \\
	  ViT-B/8 & 12 & 768 & 12 & 785 & 85M & 63 \\
\bottomrule
  \end{tabular}
\end{table}

\paragraph{Network architecture.}
The neural network $g$ is composed of a backbone $f$ (ViT~\cite{dosovitskiy2020image} or ResNet~\cite{he2016deep}), and of a projection head $h$: $g = h \circ f$.
The features used in downstream tasks are the backbone $f$ output.
The projection head consists of a 3-layer multi-layer perceptron (MLP) with hidden dimension $2048$ followed by $\ell_2$ normalization and a weight normalized fully connected layer~\cite{salimans2016weight} with $K$ dimensions, which is similar to the design from SwAV~\cite{caron2020unsupervised}.
We have tested other projection heads and this particular design appears to work best for \OURS (Appendix~\ref{ap:projhead}).
We do not use a predictor~\cite{grill2020bootstrap, chen2020exploring}, resulting in the exact same architecture in both student and teacher networks.
Of particular interest, we note that unlike standard convnets, ViT architectures do not use batch normalizations (BN) by default.
Therefore, when applying \OURS to ViT we do not use any BN also in the projection heads, making the system \emph{entirely BN-free}.

\paragraph{Avoiding collapse.} 
Several self-supervised methods differ by the operation used to avoid collapse, either through contrastive loss~\cite{wu2018unsupervised}, clustering constraints~\cite{caron2018deep,caron2020unsupervised}, predictor~\cite{grill2020bootstrap} or batch normalizations~\cite{grill2020bootstrap,richemond2020byol}.
While our framework can be stabilized with multiple normalizations~\cite{caron2020unsupervised}, it can also work with only a centering and sharpening of the momentum teacher outputs to avoid model collapse.
As shown experimentally in Section~\ref{sec:collapse}, centering prevents one dimension to dominate but encourages collapse to the uniform distribution, while the sharpening has the opposite effect.
Applying both operations balances their effects which is sufficient to avoid collapse in presence of a momentum teacher.
Choosing this method to avoid collapse trades stability for less dependence over the batch:
the centering operation only depends on first-order batch statistics and can be interpreted as adding a bias term $c$ to the teacher: $g_{t}(x) \leftarrow g_{t}(x) + c$.
The center $c$ is updated with an exponential moving average, which allows the approach to work well across different batch sizes as shown in Section~\ref{sec:smallbs}:
\begin{equation}
c \leftarrow m c + (1-m) \frac{1}{B} \sum_{i=1}^B g_{\theta_t}(x_i),
\label{eq:center_update}
\end{equation}
where $m>0$ is a rate parameter and $B$ is the batch size.
Output sharpening is obtained by using a low value for the temperature $\tau_t$ in the teacher softmax normalization.

\subsection{Implementation and evaluation protocols}
In this section, we provide the implementation details to train with \OURS and present the evaluation protocols used in our experiments.

\paragraph{Vision Transformer.}
We briefly describe the mechanism of the Vision Transformer~(ViT)~\cite{dosovitskiy2020image,vaswani2017attention} and refer to Vaswani~\etal~\cite{vaswani2017attention} for details about Transformers and to Dosovitskiy~\etal~\cite{dosovitskiy2020image} for its adaptation to images.
We follow the implementation used in \texttt{DeiT}~\cite{touvron2020training}.
We summarize the configuration of the different networks used in this paper in Table~\ref{tab:archs}.
The ViT architecture takes as input a grid of non-overlapping contiguous image patches of resolution $N \times N$.
In this paper we typically use $N=16$ (``/16'') or $N=8$ (``/8'').
The patches are then passed through a linear layer to form a set of embeddings.
We add an extra learnable token to the sequence~\cite{devlin2018bert,dosovitskiy2020image}.
The role of this token is to aggregate information from the entire sequence and we attach the projection head $h$ at its output.
We refer to this token as the class token \texttt{[CLS]} for consistency with previous works\cite{devlin2018bert,dosovitskiy2020image,touvron2020training}, even though it is not attached to any label nor supervision in our case.
The set of patch tokens and \texttt{[CLS]} token are fed to a standard Transformer network with a ``pre-norm'' layer normalization~\cite{chen2018best,klein2017opennmt}.
The Transformer is a sequence of self-attention and feed-forward layers, paralleled with skip connections.
The self-attention layers update the token representations by looking at the other token representations with an attention mechanism~\cite{bahdanau2014neural}.

\paragraph{Implementation details.}
We pretrain the models on the ImageNet dataset~\cite{russakovsky2015imagenet} without labels.
We train with the adamw optimizer~\cite{loshchilov2018fixing} and a batch size of $1024$, distributed over $16$ GPUs when using ViT-S/16.
The learning rate is linearly ramped up during the first $10$ epochs to its base value determined with the following linear scaling rule~\cite{goyal2017accurate}: $lr = 0.0005 * \text{batchsize} / 256$.
After this warmup, we decay the learning rate with a cosine schedule~\cite{loshchilov2016sgdr}.
The weight decay also follows a cosine schedule from $0.04$ to $0.4$.
The temperature $\tau_s$ is set to $0.1$ while we use a linear warm-up for $\tau_t$ from $0.04$ to $0.07$ during the first $30$ epochs.
We follow the data augmentations of BYOL~\cite{grill2020bootstrap} (color jittering, Gaussian blur and solarization) and multi-crop~\cite{caron2020unsupervised} with a bicubic interpolation to adapt the position embeddings to the scales~\cite{dosovitskiy2020image, touvron2020training}.
The code and models to reproduce our results is publicly available.

\paragraph{Evaluation protocols.}
Standard protocols for self-supervised learning are to either learn a linear classifier on frozen features~\cite{zhang2016colorful,he2020momentum} or to finetune the features on downstream tasks.
For linear evaluations, we apply random resize crops and horizontal flips augmentation during training, and report accuracy on a central crop.
For finetuning evaluations, we initialize networks with the pretrained weights and adapt them during training.
However, both evaluations are sensitive to hyperparameters, and we observe a large variance in accuracy between runs when varying the learning rate for example. 
We thus also evaluate the quality of features with a simple weighted nearest neighbor classifier ($k$-NN) as in~\cite{wu2018unsupervised}. 
We freeze the pretrain model to compute and store the features of the training data of the downstream task.
The nearest neighbor classifier then matches the feature of an image to the $k$ nearest stored features that votes for the label. 
We sweep over different number of nearest neighbors and find that $20$ NN is consistently working the best for most of our runs.
This evaluation protocol does not require any other hyperparameter tuning, nor data augmentation and can be run with only one pass over the downstream dataset, greatly simplifying the feature evaluation.


\begin{table}[t]
    \caption{
      \textbf{Linear and $k$-NN classification on ImageNet.} 
We report top-1 accuracy for linear and $k$-NN evaluations on the validation set of ImageNet for different self-supervised methods.
We focus on ResNet-50 and ViT-small architectures, but also report the best results obtained across architectures.
$^*$ are run by us.
We run the $k$-NN evaluation for models with official released weights.
	The throughput (im/s) is calculated on a NVIDIA V100 GPU with 128 samples per forward.
Parameters (M) are of the feature extractor.
}
\centering
\small
  \setlength{\tabcolsep}{4pt}
    \begin{tabular}{@{} l l c c c c @{}}
      \toprule
	    Method      & Arch. & Param. & im/s & Linear & $k$-NN  \\
      \midrule
	    \gray{Supervised} & \gray{RN50} & \gray{23} & \gray{1237} & \gray{79.3} & \gray{79.3} \\
	    SCLR~\cite{chen2020simple}     & RN50  & 23 & 1237 & 69.1 & 60.7 \\
	    MoCov2~\cite{chen2020improved}   & RN50  & 23 & 1237 & 71.1 & 61.9 \\
	    InfoMin~\cite{tian2020makes}   & RN50  & 23 & 1237 & 73.0  & 65.3 \\
	    BarlowT~\cite{zbontar2021barlow}   & RN50  & 23 & 1237 & 73.2 & 66.0 \\
	    OBoW~\cite{gidaris2020obow}   & RN50  & 23 & 1237 & 73.8 & 61.9 \\
	    BYOL~\cite{grill2020bootstrap}   & RN50  & 23 & 1237 & 74.4 & 64.8 \\
	    DCv2~\cite{caron2020unsupervised}   & RN50  & 23 & 1237 & 75.2 & 67.1 \\
	    SwAV~\cite{caron2020unsupervised} & RN50 & 23 & 1237 & \bf 75.3 & 65.7 \\
\rowcolor{Light}
	    \OURS   		       & RN50 & 23 & 1237  & \bf 75.3 & \bf 67.5 \\
\midrule
	    \gray{Supervised} & \gray{ViT-S} & \gray{21} & \gray{1007} & \gray{79.8} & \gray{79.8} \\
	    BYOL$^*$~\cite{grill2020bootstrap} & ViT-S & 21 & 1007 & 71.4 & 66.6 \\
	    MoCov2$^*$~\cite{chen2020improved} & ViT-S & 21 & 1007 & 72.7 & 64.4 \\
	    SwAV$^*$~\cite{caron2020unsupervised} & ViT-S & 21 & 1007 & 73.5 & 66.3 \\
\rowcolor{Light}
	    \OURS   		       & ViT-S  & 21 & 1007 & \bf 77.0 & \bf 74.5 \\
\midrule
\midrule
\multicolumn{5}{@{}l}{\textit{Comparison across architectures}}\\
	    SCLR~\cite{chen2020simple}             & RN50w4 & 375 & 117 & 76.8 & 69.3 \\
	    SwAV~\cite{caron2020unsupervised}   & RN50w2  & 93 & 384 & 77.3 & 67.3 \\
	    BYOL~\cite{grill2020bootstrap}   & RN50w2  & 93 & 384 & 77.4 & -- \\
\rowcolor{Light}
	    \OURS   		       & ViT-B/16  & 85 & 312 & 78.2 & 76.1 \\
	    SwAV~\cite{caron2020unsupervised}             & RN50w5 & 586 & 76 & 78.5 & 67.1 \\
	    BYOL~\cite{grill2020bootstrap}             & RN50w4 & 375 & 117 & 78.6 & -- \\
	    BYOL~\cite{grill2020bootstrap}   & RN200w2  & 250 & 123 & 79.6 & 73.9 \\
\rowcolor{Light}
	    \OURS   		       & ViT-S/8  & 21 & 180 & 79.7 & \bf 78.3 \\
	    SCLRv2~\cite{chen2020big}  & RN152w3+SK & 794 & 46 & 79.8 & 73.1 \\
\rowcolor{Light}
	    \OURS   		       & ViT-B/8  & 85 & 63 & \bf 80.1 & 77.4 \\
      \bottomrule
 \end{tabular}
    \label{tab:sota}
\end{table}

\section{Main Results}
We first validate the \OURS framework used in this study with the standard self-supervised benchmark on ImageNet.
We then study the properties of the resulting features for retrieval, object discovery and transfer-learning.

\subsection{Comparing with SSL frameworks on ImageNet}
We consider two different settings: comparison with the same architecture and across architectures.

\paragraph{Comparing with the same architecture.}
In top panel of Table~\ref{tab:sota}, we compare \OURS with other self-supervised methods with the same architecture, either a ResNet-50~\cite{he2016deep} or a ViT-small (which follows the design of DeiT-S~\cite{touvron2020training}).
The choice of ViT-S is motivated by its similarity with ResNet-50 along several axes: number of parameters (21M vs 23M), throughput (1237/sec VS 1007 im/sec) and supervised performance on ImageNet with the training procedure of~\cite{touvron2020training} (79.3\% VS 79.8\%).
We explore variants of ViT-S in Appendix~\ref{ap:ablations}.
First, we observe that \OURS performs on par with the state of the art on ResNet-50, validating that \OURS works in the standard setting.
When we switch to a ViT architecture, \OURS outperforms BYOL, MoCov2 and SwAV by +3.5\% with linear classification and by +7.9\% with $k$-NN evaluation.
More surprisingly, the performance with a simple $k$-NN classifier is almost on par with a linear classifier (74.5\% versus 77.0\%).
This property emerges only when using \OURS with ViT architectures, and does not appear with other existing self-supervised methods nor with a ResNet-50.

\paragraph{Comparing across architectures.}
On the bottom panel of Table~\ref{tab:sota}, we compare the best performance obtained across architectures.
The interest of this setting is not to compare methods directly, but to evaluate the limits of a ViT trained with \OURS when moving to larger architectures.
While training a larger ViT with \OURS improves the performance, reducing the size of the patches (``/8'' variants) has a bigger impact on the performance.
While reducing the patch size do not add parameters, it still leads to a significant reduction of running time, and larger memory usage.
Nonetheless, a base ViT with $8\times 8$ patches trained with \OURS achieves 80.1\% top-1 in linear classification and 77.4\% with a $k$-NN classifier with $10\times$ less parameters and $1.4\times$ faster run time than previous state of the art~\cite{chen2020big}.

\subsection{Properties of ViT trained with SSL}

We evaluate properties of the \OURS features in terms of nearest neighbor search, retaining information about object location and transferability to downstream tasks.

\begin{table}
\centering
\caption{\textbf{Image retrieval.}
We compare the performance in retrieval of off-the-shelf features pretrained with supervision or with \OURS on ImageNet and Google Landmarks v2 (GLDv2) dataset.
We report mAP on revisited Oxford and Paris.
Pretraining with \OURS on a landmark dataset performs particularly well.
For reference, we also report the best retrieval method with off-the-shelf features~\cite{revaud2019learning}. 
}\label{tab:retrieval}
\small
\setlength{\tabcolsep}{4pt}
\begin{tabular}{@{}l l c cc cc@{}}
\toprule
&& & \multicolumn{2}{c}{$\mathcal{R}$Ox} & \multicolumn{2}{c@{}}{$\mathcal{R}$Par} \\
\cmidrule{4-5}
\cmidrule{6-7}
Pretrain &Arch. & Pretrain    & M & H & M & H\\
\midrule
Sup.~\cite{revaud2019learning}  & RN101+R-MAC & ImNet & 49.8 & 18.5 & 74.0 & \bf 52.1 \\
\midrule
Sup.  & ViT-S/16 & ImNet & 33.5 & 8.9 & 63.0 & 37.2 \\
\rowcolor{Light}
\OURS & ResNet-50 & ImNet & 35.4 & 11.1 & 55.9 & 27.5 \\
\rowcolor{Light}
\OURS & ViT-S/16 & ImNet & 41.8 & 13.7 & 63.1 & 34.4 \\
\rowcolor{Light}
\OURS & ViT-S/16  & GLDv2 & \bf 51.5 & \bf 24.3 & \bf 75.3 & 51.6 \\
\bottomrule
\end{tabular}
\end{table}

\subsubsection{Nearest neighbor retrieval with \OURS ViT}
The results on ImageNet classification have exposed the potential of our features for tasks relying on nearest neighbor retrieval.
In this set of experiments, we further consolidate this finding on landmark retrieval and copy detection tasks.

\paragraph{Image Retrieval.}
We consider the revisited~\cite{radenovic2018revisiting} Oxford and Paris image retrieval datasets~\cite{philbin2008lost}.
They contain 3 different splits of gradual difficulty with query/database pairs.
We report the Mean Average Precision (mAP) for the Medium (M) and Hard (H) splits.
In Table~\ref{tab:retrieval}, we compare the performance of different \emph{off-the-shelf} features obtained with either supervised or \OURS training.
We freeze the features and directly apply $k$-NN for retrieval.
We observe that \OURS features outperform those trained on ImageNet with labels.

An advantage of SSL approaches is that they can be trained on any dataset, without requiring any form of annotations.
We train \OURS on the 1.2M clean set from Google Landmarks v2 (GLDv2)~\cite{weyand2020google}, a dataset of landmarks designed for retrieval purposes.
\OURS ViT features trained on GLDv2 are remarkably good, outperforming previously published methods based on off-the-shelf descriptors~\cite{tolias2015particular,revaud2019learning}.

\begin{table}
\centering
\caption{\textbf{Copy detection.}
We report the mAP performance in copy detection on Copydays ``strong'' subset~\cite{douze2009evaluation}.
For reference, we also report the performance of the multigrain model~\cite{berman2019multigrain}, trained specifically for particular object retrieval.
}\label{tab:copy}
\small
\begin{tabular}{@{}l l c c c@{}}
\toprule
	Method & Arch. & Dim. & Resolution & mAP \\
\midrule
Multigrain~\cite{berman2019multigrain}  & ResNet-50 & 2048 & $224^2$ & 75.1 \\
Multigrain~\cite{berman2019multigrain}  & ResNet-50 & 2048 & largest side 800 & 82.5 \\
\midrule
Supervised~\cite{touvron2020training}  & ViT-B/16 & 1536 & $224^2$ & 76.4 \\
\rowcolor{Light}
\OURS  & ViT-B/16 & 1536 & $224^2$ & 81.7 \\
\rowcolor{Light}
\OURS  & ViT-B/8 & 1536 & $320^2$ & \bf 85.5 \\
\bottomrule
\end{tabular}
\end{table}

\paragraph{Copy detection.}
We also evaluate the performance of ViTs trained with \OURS on a copy detection task.
We report the mean average precision on the ``strong'' subset of the INRIA Copydays dataset~\cite{douze2009evaluation}.
The task is to recognize images that have been distorted by blur, insertions, print and scan, etc. 
Following prior work~\cite{berman2019multigrain}, we add 10k distractor images randomly sampled from the YFCC100M dataset~\cite{thomee2015yfcc100m}.
We perform copy detection directly with cosine similarity on the features obtained from our pretrained network.
The features are obtained as the concatenation of the output \texttt{[CLS]} token and of the GeM pooled~\cite{radenovic2018fine} output patch tokens.
This results in a 1536d descriptor for ViT-B.
Following~\cite{berman2019multigrain}, we apply whitening on the features.
We learn this transformation on an extra 20K random images from YFCC100M, distincts from the distractors.
Table~\ref{tab:copy} shows that ViT trained with \OURS is very competitive on copy detection.

\begin{table}[h]
\centering
	\caption{\textbf{DAVIS 2017 Video object segmentation.}
We evaluate the quality of frozen features on video instance tracking.
We report mean region similarity $\mathcal{J}_m$ and mean contour-based accuracy $\mathcal{F}_m$.
We compare with existing self-supervised methods and a supervised ViT-S/8 trained on ImageNet.
Image resolution is 480p.
  }
  \setlength{\tabcolsep}{6pt}
\small
\begin{tabular}{@{}lllccc@{}}
\toprule
Method & Data & Arch. & $ (\mathcal{J}$\&$\mathcal{F})_m$ & $\mathcal{J}_m$ & $\mathcal{F}_m$\\
\midrule
\multicolumn{4}{@{}l}{\textit{Supervised}}\\
ImageNet & INet & ViT-S/8 & 66.0 & 63.9 & 68.1\\
STM~\cite{oh2019video} & I/D/Y & RN50 & 81.8 & 79.2 & 84.3 \\
\midrule
\multicolumn{4}{@{}l}{\textit{Self-supervised}}\\
CT~\cite{wang2019learning} & VLOG & RN50 & 48.7 & 46.4 & 50.0\\
MAST~\cite{lai2020mast} & YT-VOS & RN18 & 65.5 & 63.3 & 67.6 \\
STC~\cite{jabri2020space} & Kinetics & RN18 & 67.6 & 64.8 & 70.2\\
\rowcolor{Light}
\OURS & INet & ViT-S/16 & 61.8 & 60.2 & 63.4 \\
\rowcolor{Light}
\OURS & INet & ViT-B/16 & 62.3 & 60.7 & 63.9 \\
\rowcolor{Light}
\OURS & INet & ViT-S/8 & \bf 69.9 & \bf 66.6 & \bf 73.1\\
\rowcolor{Light}
\OURS & INet & ViT-B/8 & \bf 71.4 & \bf 67.9 & \bf 74.9\\
\bottomrule
\end{tabular}
\label{tab:dense}
\end{table}

\begin{figure}[t]
\centering
  \setlength{\tabcolsep}{0.5pt}
\begin{tabular}{cc ccc cc}
	\includegraphics[width=.24\linewidth]{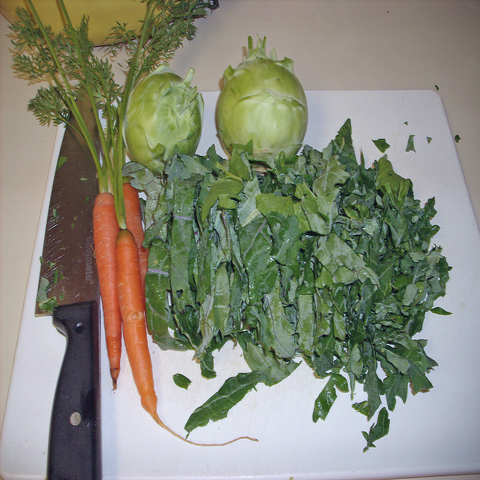}&
	\includegraphics[width=.24\linewidth]{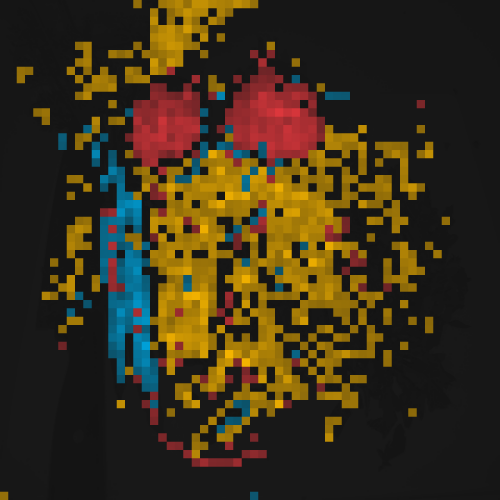}&&&&
	\includegraphics[width=.24\linewidth]{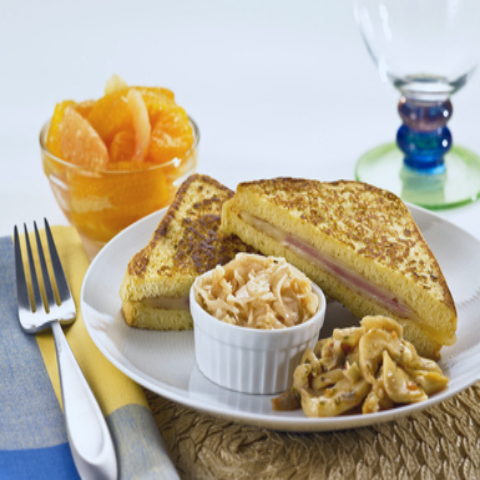}&
	\includegraphics[width=.24\linewidth]{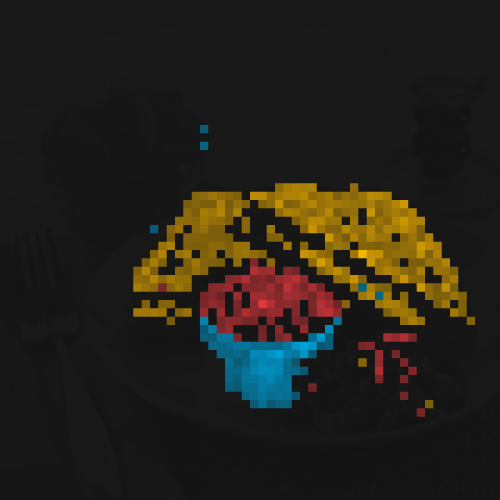}
\\
	\includegraphics[width=.24\linewidth]{3017.png}&
	\includegraphics[width=.24\linewidth]{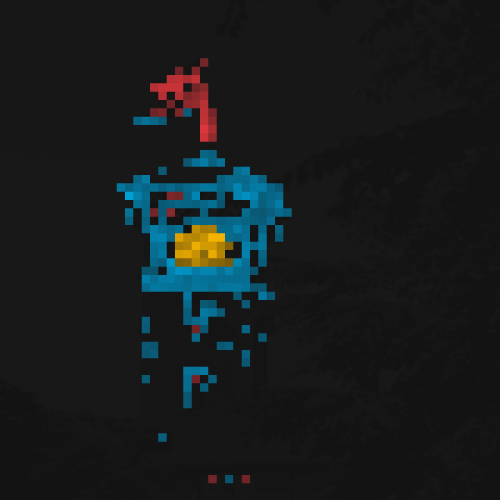}&&&&
	\includegraphics[width=.24\linewidth]{3425.png}&
	\includegraphics[width=.24\linewidth]{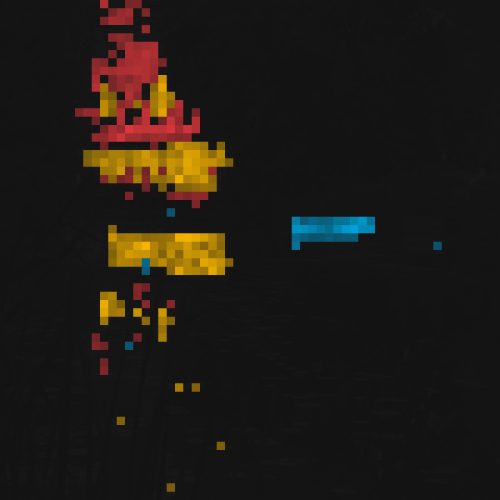}
\\
	\includegraphics[width=.24\linewidth]{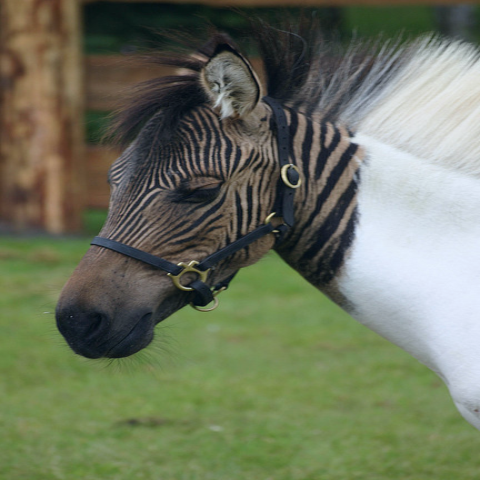}&
	\includegraphics[width=.24\linewidth]{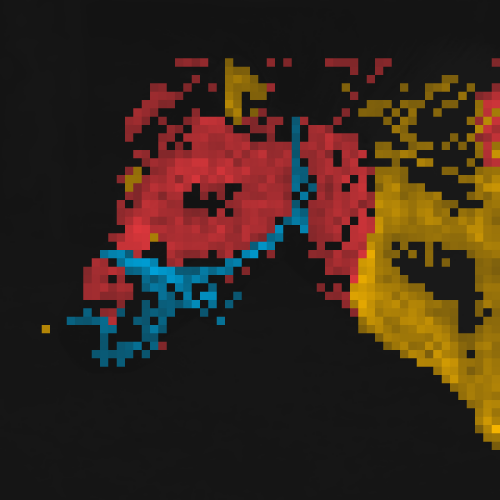}&&&&
	\includegraphics[width=.24\linewidth]{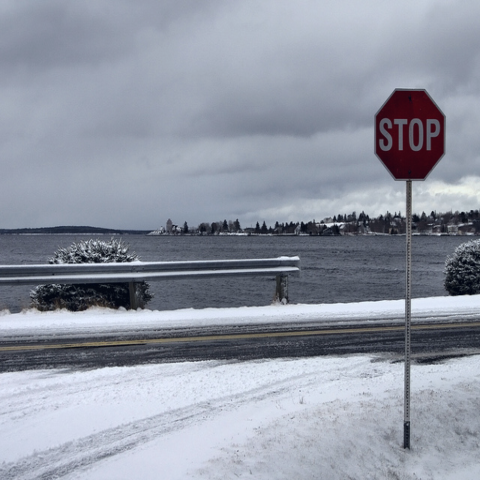}&
	\includegraphics[width=.24\linewidth]{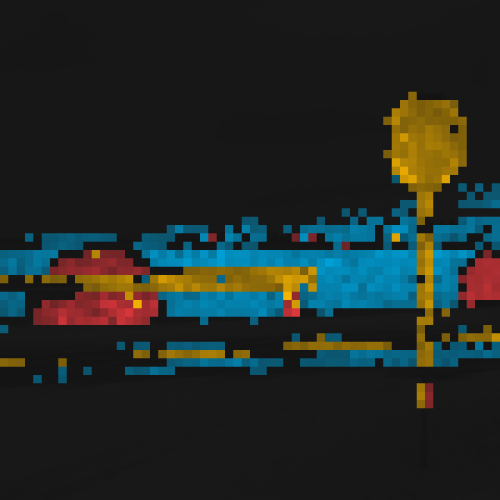}
\\
	\includegraphics[width=.24\linewidth]{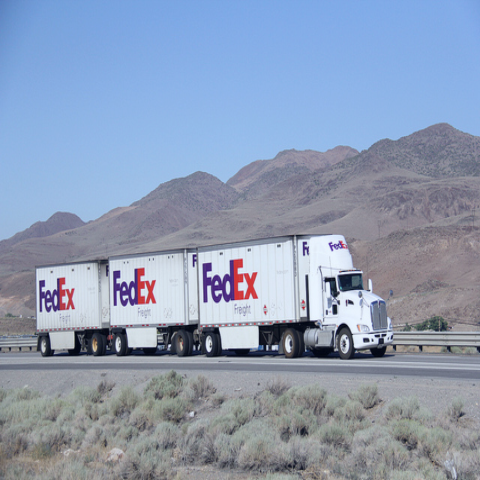}&
	\includegraphics[width=.24\linewidth]{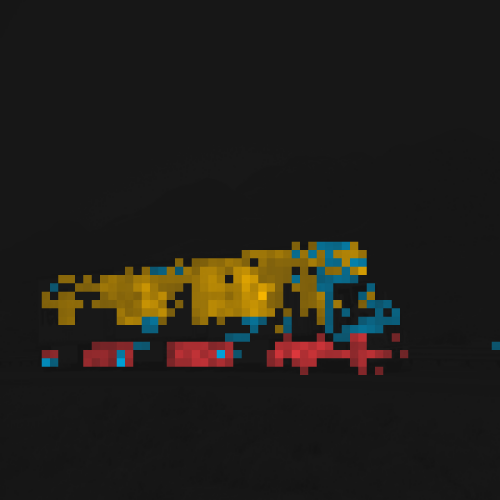}&&&&
	\includegraphics[width=.24\linewidth]{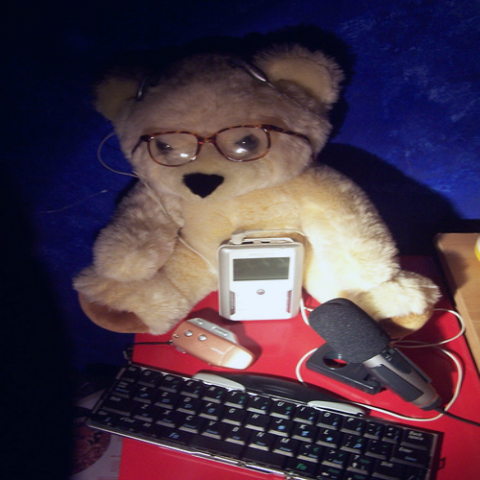}&
	\includegraphics[width=.24\linewidth]{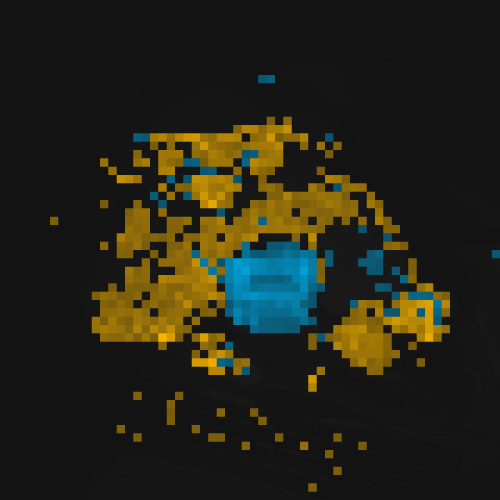}
\end{tabular}
\caption{\textbf{Attention maps from multiple heads.}
We consider the heads from the last layer of a ViT-S/8 trained with \OURS and display the self-attention for \texttt{[CLS]} token query.
Different heads, materialized by different colors, focus on different locations that represents different objects or parts
	(more examples in Appendix).}
\label{fig:multihead}
\end{figure}

\subsubsection{Discovering the semantic layout of scenes}

As shown qualitatively in Figure~\ref{fig:att}, our self-attention maps contain information about the segmentation of an image. 
In this study, we measure this property on a standard benchmark as well as by directly probing the quality of masks generated from these attention maps.

\paragraph{Video instance segmentation.}
In Tab.~\ref{tab:dense}, we evaluate the output patch tokens on the DAVIS-2017 video instance segmentation benchmark~\cite{pont20172017}.
We follow the experimental protocol in Jabri~\etal~\cite{jabri2020space} and segment scenes with a nearest-neighbor between consecutive frames; we thus do not train any model on top of the features, nor finetune any weights for the task.
We observe in Tab.~\ref{tab:dense} that even though our training objective nor our architecture are designed for dense tasks, the performance is competitive on this benchmark.
Since the network is not finetuned, the output of the model must have retained some spatial information.
Finally, for this dense recognition task, the variants with small patches (``/8'') perform much better (+$9.1\%$ $(\mathcal{J}$\&$\mathcal{F})_m$ for ViT-B).

\begin{figure}[t]
\centering
  \setlength{\tabcolsep}{0.8pt}
\begin{tabular}{@{}lccccc}
	\small \textit{Supervised} &&&&\\
	\includegraphics[width=.195\linewidth]{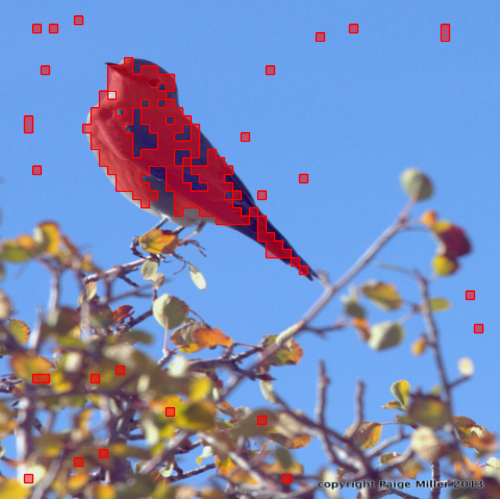}&
	\includegraphics[width=.195\linewidth]{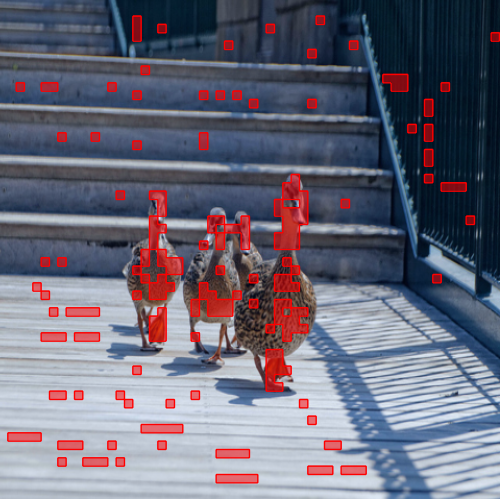}&
	\includegraphics[width=.195\linewidth]{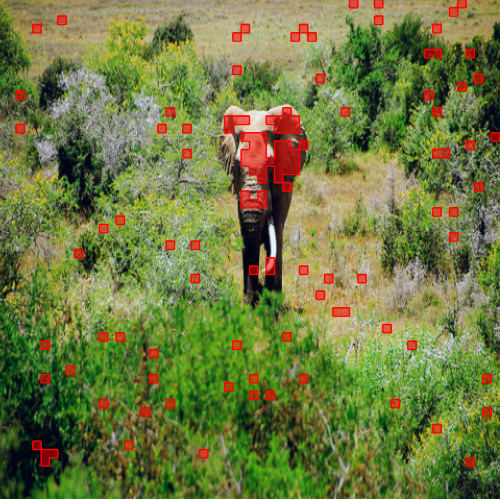}&
	\includegraphics[width=.195\linewidth]{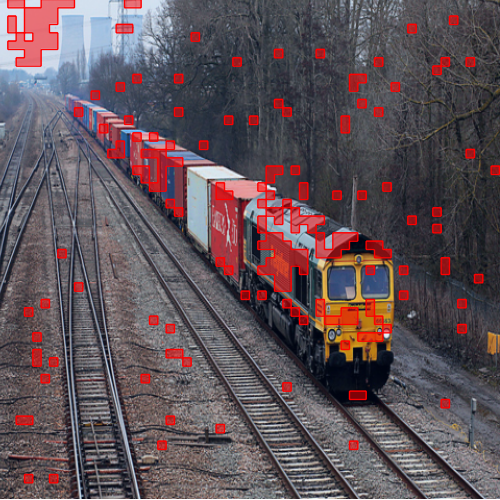}&
	\includegraphics[width=.195\linewidth]{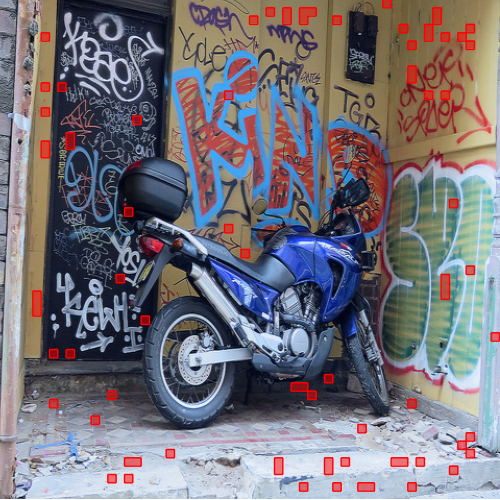}\\
	\small \textit{\OURS} &&&&\\
	\includegraphics[width=.195\linewidth]{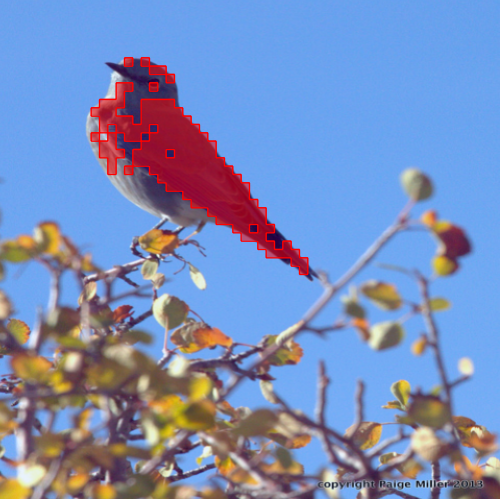}&
	\includegraphics[width=.195\linewidth]{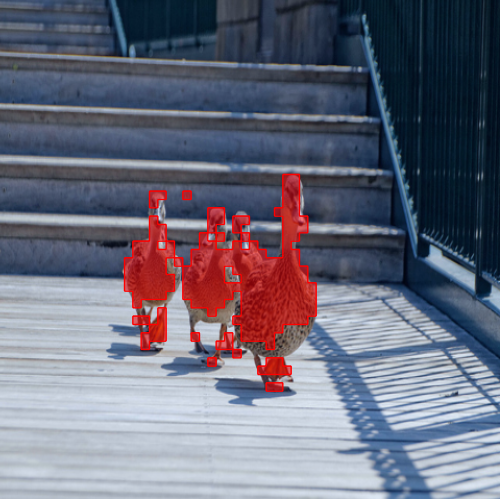}&
	\includegraphics[width=.195\linewidth]{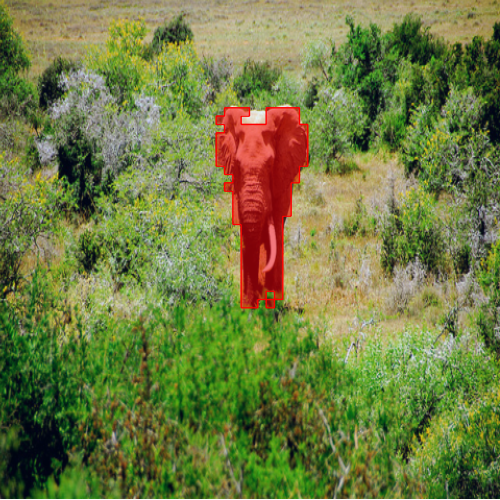}&
	\includegraphics[width=.195\linewidth]{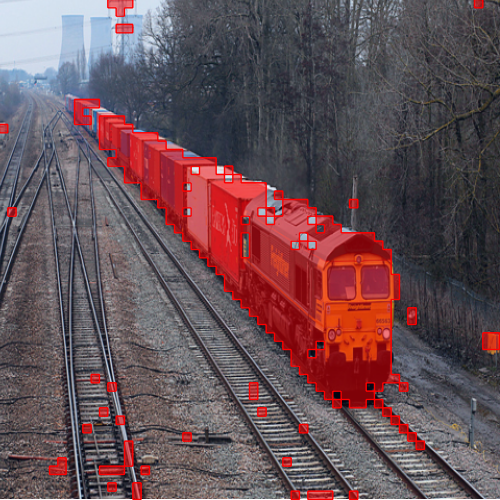}&
	\includegraphics[width=.195\linewidth]{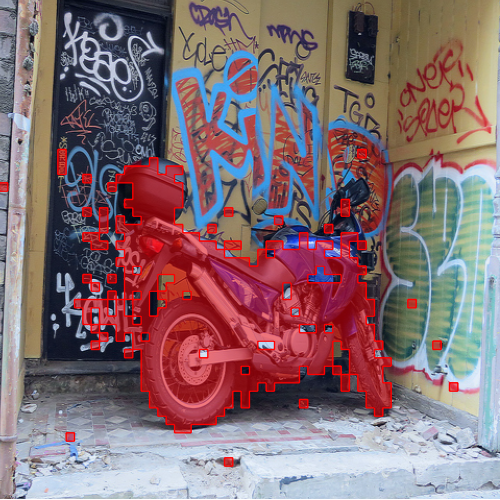}\\
\end{tabular}
  \setlength{\tabcolsep}{6pt}
\small
\begin{tabular}{lccc}
\toprule
  & Random & Supervised  & \OURS  \\
\midrule
ViT-S/16 &  22.0 & 27.3 & \colorbox{Light}{45.9}\\
ViT-S/8  & 21.8 & 23.7 & \colorbox{Light}{44.7}\\
\bottomrule
\end{tabular}
\caption{\textbf{Segmentations from supervised versus \OURS.}
We visualize masks obtained by thresholding the self-attention maps to keep 60\% of the mass.
On top, we show the resulting masks for a ViT-S/8 trained with supervision and \OURS. 
We show the best head for both models. 
The table at the bottom compares the Jaccard similarity between the ground truth and these masks on the validation images of PASCAL VOC12 dataset.
}
\label{fig:vssup}
\end{figure}

\paragraph{Probing the self-attention map.}
In Fig.~\ref{fig:multihead}, we show that different heads can attend to different semantic regions of an image,  even when they are occluded (the bushes on the third row) or small (the flag on the second row).
Visualizations are obtained with $480$p images, resulting in sequences of 3601 tokens for ViT-S/8.
In Fig.~\ref{fig:vssup}, we show that a supervised ViT does not attend well to objects in presence of clutter both qualitatively and quantitatively.
We report the Jaccard similarity between the ground truth and segmentation masks obtained by thresholding the self-attention map to keep 60\% of the mass.
Note that the self-attention maps are smooth and not optimized to produce a mask. 
Nonetheless, we see a clear difference between the supervised or \OURS models with a significant gap in terms of Jaccard similarities.
Note that self-supervised convnets also contain information about segmentations but it requires dedicated methods to extract it from their weights~\cite{gur2020visualization}.

\subsubsection{Transfer learning on downstream tasks}
In Tab.~\ref{tab:transfers}, we evaluate the quality of the features pretrained with \OURS on different downstream tasks.
We compare with features from the same architectures trained with supervision on ImageNet.
We follow the protocol used in Touvron~\etal~\cite{touvron2020training} and finetune the features on each downstream task.
We observe that for ViT architectures, self-supervised pretraining transfers better than features trained with supervision, which is consistent with observations made on convolutional networks~\cite{caron2020unsupervised,he2020momentum,sariyildiz2020concept}.
Finally, self-supervised pretraining greatly improves results on ImageNet (+1-2\%). 

\begin{table}[t]
  \centering
	\caption{\textbf{Transfer learning by finetuning pretrained models on different datasets.}
We report top-1 accuracy.
Self-supervised pretraining with \OURS transfers better than supervised pretraining.}
\small
  \setlength{\tabcolsep}{3pt}
  \begin{tabular}{@{}l ccccccc@{}}
    \toprule
	   & Cifar$_{\text{10}}$ & Cifar$_{\text{100}}$ & INat$_{\text{18}}$ & INat$_{\text{19}}$ & Flwrs & Cars & INet \\
    \midrule
\textit{\small ViT-S/16}\\
	  Sup.~\cite{touvron2020training}           & \bf 99.0 & 89.5 & 70.7     & 76.6     & 98.2     & 92.1 & 79.9 \\
\rowcolor{Light}
	  \OURS 			            & \bf 99.0 & \bf 90.5 & \bf 72.0 & \bf 78.2 & \bf 98.5 & \bf 93.0 & \bf 81.5 \\
    \midrule
\textit{\small ViT-B/16}\\
	  Sup.~\cite{touvron2020training}           & 99.0 & 90.8 & \bf 73.2 & 77.7 & 98.4 & 92.1 & 81.8 \\
\rowcolor{Light}
	  \OURS                                     & \bf 99.1 & \bf 91.7 &  72.6 & \bf 78.6 & \bf 98.8 & \bf 93.0 & \bf 82.8 \\
    \bottomrule
  \end{tabular}
  \label{tab:transfers}
\end{table}

\section{Ablation Study of \OURS}
In this section, we empirically study \OURS applied to ViT.
The model considered for this entire study is ViT-S.
We also refer the reader to Appendix for additional studies.

\subsection{Importance of the Different Components}
\label{sec:comp}
We show the impact of adding different components from self-supervised learning on ViT trained with our framework. 

\begin{table}[t]
  \centering
	\caption{
    \textbf{Important component for self-supervised ViT pretraining.} 
Models are trained for 300 epochs with ViT-S/16.
We study the different components that matter for the $k$-NN and linear (``Lin.'') evaluations. 
For the different variants, we \colorbox{Light}{highlight} the differences from the default \OURS setting.
The best combination is the momentum encoder with the multicrop augmentation and the cross-entropy loss.
We also report results with BYOL~\cite{grill2020bootstrap}, MoCo-v2~\cite{chen2020improved} and SwAV~\cite{caron2020unsupervised}.
  }
\small
  \setlength{\tabcolsep}{4pt}
  \begin{tabular}{@{}p{.8em}@{}l ccccccc@{}}
    \toprule
	  & Method & Mom. & SK & MC & Loss & Pred. & $k$-NN & Lin. \\
    \midrule
	  \rownumber{1}&	  \OURS & \checkmark & \xmark & \checkmark & \texttt{CE} & \xmark & 72.8 & 76.1 \\
	  \rownumber{2}&	   & \colorbox{Light}{\xmark} & \xmark & \checkmark & \texttt{CE} & \xmark & 0.1 & 0.1 \\
	  \rownumber{3}&	   & \checkmark & \colorbox{Light}{\checkmark} & \checkmark & \texttt{CE} & \xmark & 72.2 & 76.0\\
	  \rownumber{4}&	   & \checkmark & \xmark & \colorbox{Light}{\xmark} & \texttt{CE} & \xmark & 67.9 & 72.5\\
	  \rownumber{5}&	   & \checkmark & \xmark &  \checkmark & \colorbox{Light}{\texttt{MSE}} &\xmark & 52.6 & 62.4\\
	  \rownumber{6}&	   & \checkmark & \xmark &  \checkmark & \texttt{CE} & \colorbox{Light}{\checkmark} & 71.8 & 75.6 \\
    \midrule
	  \rownumber{7}&	   BYOL & \checkmark & \xmark &  \xmark & \texttt{MSE} & \checkmark & 66.6 & 71.4\\
	  \rownumber{8}&	   MoCov2 & \checkmark & \xmark &  \xmark & \texttt{INCE} & \xmark & 62.0 & 71.6 \\
	  \rownumber{9}&	   SwAV & \xmark & \checkmark &  \checkmark & \texttt{CE} & \xmark & 64.7 & 71.8 \\
    \bottomrule
\multicolumn{9}{c}{SK: Sinkhorn-Knopp, MC: Multi-Crop, Pred.: Predictor}\\
\multicolumn{9}{c}{CE: Cross-Entropy, MSE: Mean Square Error, INCE: InfoNCE}\\
  \end{tabular}
\label{tab:components}
\end{table}

In Table~\ref{tab:components}, we report different model variants as we add or remove components.
First, we observe that in the absence of momentum, our framework does not work (row \rownumber{2}) and more advanced operations, SK for example, are required to avoid collapse (row \rownumber{9}).
However, with momentum, using SK has little impact (row \rownumber{3}).
In addtition, comparing rows \rownumber{3} and \rownumber{9} highlights the importance of the momentum encoder for performance.
Second, in rows \rownumber{4} and \rownumber{5}, we observe that multi-crop training and the cross-entropy loss in \OURS are important components to obtain good features.
We also observe that adding a predictor to the student network has little impact (row \rownumber{6}) while it is critical in BYOL to prevent collapse~\cite{chen2020exploring,grill2020bootstrap}.
For completeness, we propose in Appendix~\ref{ap:comp} an extended version of this ablation study.

\begin{figure}[t]
  \begin{minipage}{0.6\linewidth}
	\centering
  \includegraphics[width=\linewidth]{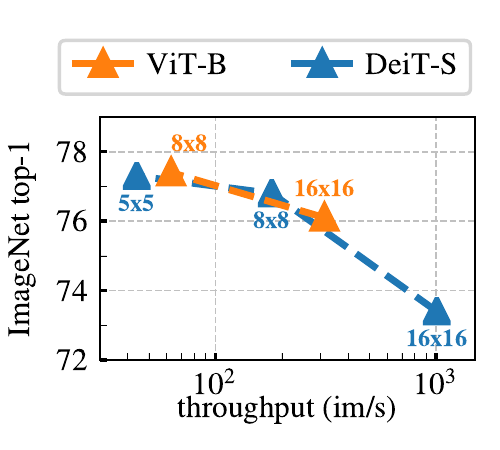}
\end{minipage}
  \begin{minipage}{0.35\linewidth}
	\caption{
    \textbf{Effect of Patch Size.}
     $k$-NN evaluation as a function of the throughputs for different input patch sizes with ViT-B and ViT-S. Models are trained for 300 epochs.
}
  \label{fig:patch-size}
\end{minipage}
\end{figure}

\paragraph{Importance of the patch size.}
In Fig.~\ref{fig:patch-size}, we compare the $k$-NN classification performance of ViT-S models trained with different patch sizes, $16\times 16$, $8\times 8$ and $5\times 5$.
We also compare to ViT-B with $16 \times 16$ and $8\times 8$ patches.
All the models are trained for 300 epochs.
We observe that the performance greatly improves as we decrease the size of the patch.
It is interesting to see that performance can be greatly improved without adding additional parameters.
However, the performance gain from using smaller patches comes at the expense of throughput: when using 5$\times$5 patches, the throughput falls to 44 im/s, \emph{vs} 180 im/s for 8$\times$8 patches.

\subsection{Impact of the choice of Teacher Network}
\label{sec:teacher}
In this ablation, we experiment with different teacher network to understand its role in \OURS. 
We compare models trained for $300$ epochs using the $k$-NN protocol.

\begin{figure}[t]
  \begin{minipage}{0.55\linewidth}
	\centering
    \includegraphics[width=\linewidth]{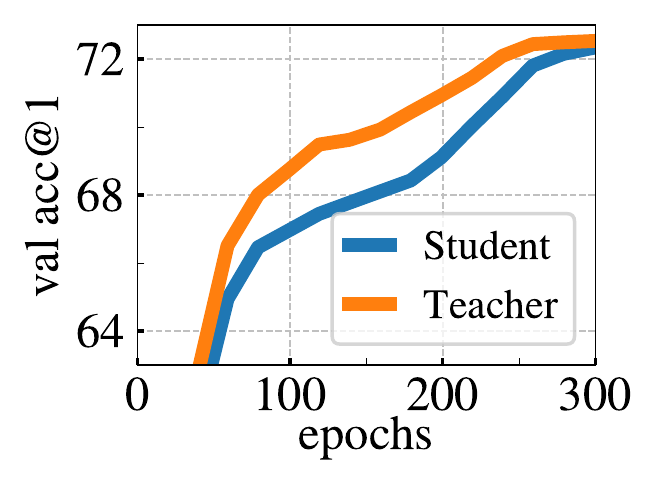}
  \end{minipage}
\hspace{.5em}
  \begin{minipage}[t]{0.4\linewidth}
    \centering
\small
  \begin{tabular}{@{}l c@{}}
    \toprule
    Teacher & Top-1 \\
    \midrule
    Student copy & 0.1 \\
    Previous iter & 0.1\\
    Previous epoch & 66.6 \\
    Momentum & 72.8 \\
  \bottomrule
\phantom{latex!}
  \end{tabular}
  \end{minipage}
	\caption{
Top-1 accuracy on ImageNet validation with $k$-NN classifier.
\textbf{(left)} Comparison between the performance of the momentum teacher and the student during training.
\textbf{(right)} Comparison between different types of teacher network.
The momentum encoder leads to the best performance but is not the only viable option.
	}
	\label{fig:mom}
\end{figure}

\paragraph{Building different teachers from the student.}
In Fig.~\ref{fig:mom}(right), we compare different strategies to build the teacher from previous instances of the student besides the momentum teacher.
First we consider using the student network from a previous epoch as a teacher.
This strategy has been used in a memory bank~\cite{wu2018unsupervised} or as a form of clustering hard-distillation~\cite{caron2018deep,asano2019self,chen2020unsupervised}.
Second, we consider using the student network from the previous iteration, as well as a copy of the student for the teacher.
In our setting, using a teacher based on a recent version of the student does not converge.
This setting requires more normalizations to work.
Interestingly, we observe that using a teacher from the previous epoch does not collapse, providing performance in the $k$-NN evaluation competitive with existing frameworks such as MoCo-v2 or BYOL. 
While using a momentum encoder clearly provides superior performance to this naive teacher, this finding suggests that there is a space to investigate alternatives for the teacher.

\paragraph{Analyzing the training dynamic.}
To further understand the reasons why a momentum teacher works well in our framework, we study its dynamic during the training of a ViT in the left panel of Fig.~\ref{fig:mom}.
A key observation is that this teacher constantly outperforms the student during the training, and we observe the same behavior when training with a ResNet-50 (Appendix~\ref{ap:ablations}).
This behavior has not been observed by other frameworks also using momentum~\cite{he2020momentum,grill2020bootstrap}, nor when the teacher is built from the previous epoch.
We propose to interpret the momentum teacher in \OURS as a form of Polyak-Ruppert averaging~\cite{polyak1992acceleration, ruppert1988efficient} with an exponentially decay.
Polyak-Ruppert averaging is often used to simulate model ensembling to improve the performance of a network at the end of the training~\cite{jean2014using}. 
Our method can be interpreted as applying Polyak-Ruppert averaging during the training to constantly build a model ensembling that has superior performances.
This model ensembling then guides the training of the student network~\cite{tarvainen2017mean}. 

\subsection{Avoiding collapse}
\label{sec:collapse}

\begin{figure}[t]
	\centering
  \includegraphics[width=\linewidth]{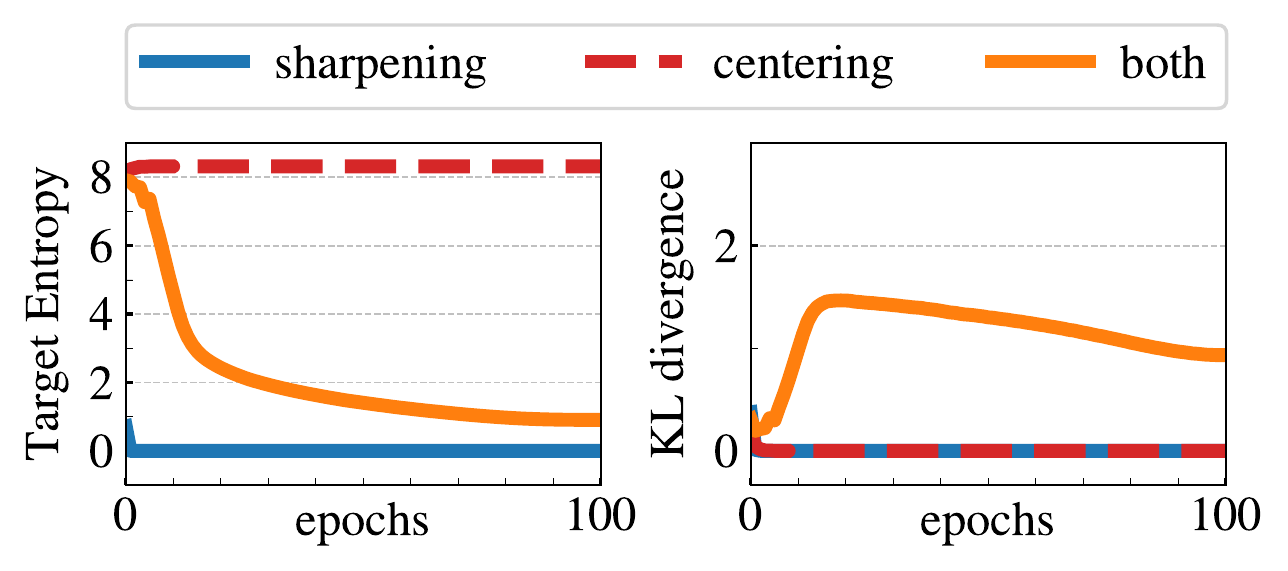}
	\caption{
\textbf{Collapse study.}
    \textbf{(left)}: evolution of the teacher's target entropy along training epochs;
    \textbf{(right)}: evolution of KL divergence between teacher and student outputs.
	}
	\label{fig:collapse}
\end{figure}

We study the complementarity role of centering and target sharpening to avoid collapse.
There are two forms of collapse: regardless of the input, the model output is uniform along all the dimensions or dominated by one dimension. 
The centering avoids the collapse induced by a dominant dimension, but encourages an uniform output.
Sharpening induces the opposite effect.
We show this complementarity by decomposing the cross-entropy $H$ into an entropy $h$ and the Kullback-Leibler divergence (``KL'') $D_{KL}$:
\begin{equation}
  H(P_t, P_s) = h(P_t) + D_{KL}(P_t | P_s).
\end{equation}
A KL equal to zero indicates a constant output, and hence a collapse.
In Fig.~\ref{fig:collapse}, we plot the entropy and KL during training with and without centering and sharpening.
If one operation is missing, the KL converges to zero, indicating a collapse.
However, the entropy $h$ converges to different values: $0$ with no centering and $-\log(1/K)$ with no sharpening, indicating that both operations induce different form of collapse.
Applying both operations balances these effects (see study of the sharpening parameter $\tau_t$ in Appendix~\ref{ap:ablations}).

\subsection{Compute requirements}
\begin{table}[t]
\centering
  \caption{
    \textbf{Time and memory requirements.}
We show total running time and peak memory per GPU (``mem.'') when running ViT-S/16 \OURS models on two 8-GPU machines.
We report top-1 ImageNet val acc with linear evaluation for several variants of multi-crop, each having a different level of compute requirement.
  }
  \label{tab:compute}
  \setlength{\tabcolsep}{3pt}
\begin{tabular}{ll g r c g r r}
	\toprule
	&& \multicolumn{2}{c}{100 epochs} && \multicolumn{2}{c}{300 epochs} &   \\
    \cmidrule{3-4}
    \cmidrule{6-7}
	multi-crop && top-1 & time && top-1 & time & mem. \\
    \midrule
	$2\!\times\!224^2$ && 67.8 & 15.3h && 72.5 & 45.9h & 9.3G \\
	$2\!\times\!224^2 + \phantom{0}2\!\times\!96^2$ && 71.5 & 17.0h && 74.5 & 51.0h & 10.5G \\
	$2\!\times\!224^2 + \phantom{0}6\!\times\!96^2$ && 73.8 & 20.3h && 75.9 & 60.9h & 12.9G \\
	$2\!\times\!224^2 + 10\!\times\!96^2$ && 74.6 & 24.2h && 76.1 & 72.6h & 15.4G \\
\bottomrule
  \end{tabular}
\end{table}
In Tab.~\ref{tab:compute}, we detail the time and GPU memory requirements when running ViT-S/16 \OURS models on two $8$-GPU machines.
We report results with several variants of multi-crop training, each having a different level of compute requirement.
We observe in Tab.~\ref{tab:compute} that using multi-crop improves the accuracy / running-time tradeoff for \OURS runs.
For example, the performance is $72.5\%$ after $46$ hours of training without multi-crop (i.e. $2\!\times\!224^2$) while \OURS in $2\!\times\!224^2 + 10\!\times\!96^2$ crop setting reaches $74.6\%$ in $24$ hours only.
This is an improvement of $+2\%$ while requiring $2\!\times$ less time, though the memory usage is higher ($15.4G$ versus $9.3G$).
We observe that the performance boost brought with multi-crop cannot be caught up by more training in the $2\!\times\!224^2$ setting, which shows the value of the ``local-to-global'' augmentation.
Finally, the gain from adding more views diminishes (+.2\% form $6\!\times$ to $10\!\times$ $96^2$ crops) for longer trainings.

Overall, training \OURS with Vision Transformers achieves $76.1$ top-1 accuracy using two 8-GPU servers for 3 days.
This result outperforms state-of-the-art self-supervised systems based on convolutional networks of comparable sizes with a significant reduction of computational requirements~\cite{grill2020bootstrap,caron2020unsupervised}.
Our code is available to train self-supervised ViT on a limited number of GPUs.

\subsection{Training with small batches}
\label{sec:smallbs}
\begin{table}[h]
  \centering
\begin{minipage}{.5\linewidth}
\small
  \setlength{\tabcolsep}{4pt}
  \begin{tabular}{@{}l c c c c@{}}
    \toprule
	  bs & 128 & 256 & 512 & 1024 \\
    \midrule
	  top-1 & 57.9 & 59.1 & 59.6 & 59.9  \\
    \bottomrule
\phantom{latex!}
  \end{tabular}
\end{minipage}
\hspace{1em}
\begin{minipage}{.4\linewidth}
  \caption{
	  \textbf{Effect of batch sizes.} Top-1 with $k$-NN for models trained for 100 epochs without multi-crop.
  }
  \label{tab:bs}
\end{minipage}
\vspace{-1em}
\end{table}
In Tab.~\ref{tab:bs}, we study the impact of the batch size on the features obtained with \OURS.
We also study the impact of the smooth parameter $m$ used in the centering update rule of Eq.~\ref{eq:center_update} in Appendix~\ref{ap:ablations}.
We scale the learning rate linearly with the batch size~\cite{goyal2017accurate}: $lr = 0.0005 * \text{batchsize} / 256 $.
Tab.~\ref{tab:bs} confirms that we can train models to high performance with small batches.
Results with the smaller batch sizes ($bs=128$) are slightly below our default training setup of $bs=1024$, and would certainly require to re-tune hyperparameters like the momentum rates for example.
Note that the experiment with batch size of $128$ runs on only $1$ GPU.
We have explored training a model with a batch size of $8$, reaching $35.2\%$ after $50$ epochs, showing the potential for training large models that barely fit an image per GPU.

\section{Conclusion}

In this work, we have shown the potential of self-supervised pretraining a standard ViT model, achieving performance that are comparable with the best convnets specifically designed for this setting.
We have also seen emerged two properties that can be leveraged in future applications:
the quality of the features in $k$-NN classification has a potential for image retrieval where ViT are already showing promising results~\cite{el2021training}.
The presence of information about the scene layout in the features can also benefit weakly supervised image segmentation.
However, the main result of this paper is that we have evidences that self-supervised learning could be the key to developing a BERT-like model based on ViT.
In the future, we plan to explore if pretraining a large ViT model with \OURS on random uncurated images could push the limits of visual features~\cite{goyal2021self}.

\paragraph{Acknowledgement.}
We thank Mahmoud Assran, Matthijs Douze, Allan Jabri, Jure Zbontar, Alaaeldin El-Nouby, Y-Lan Boureau, Kaiming He, Thomas Lucas as well as the Thoth and FAIR teams for their help, support and discussions around this project.
Julien Mairal was funded by the ERC grant number 714381 (SOLARIS project) and by ANR 3IA MIAI@Grenoble Alpes (ANR-19-P3IA-0003).

{\small
\bibliographystyle{ieee_fullname}
\bibliography{egbib}
}

\renewcommand{\thesubsection}{\Alph{subsection}}
\newcommand\tab[1][5mm]{\hspace*{#1}}
\section*{Appendix}

\subsection{Additional Results}
\label{ap:more_results}
\paragraph{$k$-NN classification.}
\label{sec:knn}
In Tab.~\ref{tab:knn}, we evaluate the frozen representations given by ResNet-50 or ViT-small pre-trained with \OURS with two evaluation protocols: linear or $k$-NN.
For both evaluations, we extract representations from a pre-trained network without using any data augmentation.
Then, we perform classification either with weighted $k$-NN or with a linear regression learned with \texttt{cyanure} library~\cite{mairal2019cyanure}.
In Tab.~\ref{tab:knn} we see that ViT-S accuracies are better than accuracies obtained with RN50 both with a linear or a $k$-NN classifier.
However, the performance gap when using the $k$-NN evaluation is much more significant than when considering linear evaluation.
For example on ImageNet 1\%, ViT-S outperforms ResNet-50 by a large margin of $+14.1\%$ with $k$-NN evaluation.
This suggests that transformers architectures trained with \OURS might offer more model flexibility that benefits the $k$-NN evaluation.
$K$-NN classifiers have the great advantage of being fast and light to deploy, without requiring any domain adaptation.
Overall, ViT trained with \OURS provides features that combine particularly well with $k$-NN classifiers.

\begin{table}[h]
  \centering
\small
  \setlength{\tabcolsep}{5pt}
\caption{\textbf{$k$-NN and linear evaluation for ViT-S/16 and ResNet-50 pre-trained with \OURS.}
	We use ImageNet-1k~\cite{russakovsky2015imagenet} (``Inet''), Places205~\cite{zhou2014learning}, PASCAL VOC~\cite{everingham2010pascal} and Oxford-102 flowers (``FLOWERS'')~\cite{nilsback2008automated}.
ViT trained with \OURS provides features that are particularly $k$-NN friendly.}
  \begin{tabular}{@{}l ccg c ccg@{}}
    \toprule
    & \multicolumn{3}{c}{Logistic} && \multicolumn{3}{c}{$k$-NN} \\
    \cmidrule{2-4}
    \cmidrule{6-8}
	  & RN50 & ViT-S & $\Delta$ && RN50 & ViT-S & $\Delta$ \\
    \midrule
	  Inet   100\% & 72.1 & 75.7 & 3.6 && 67.5 & 74.5 & 7.0 \\
	  Inet   10\% & 67.8 & 72.2 & 4.4 && 59.3 & 69.1 & 9.8 \\
	  Inet    1\% & 55.1 & 64.5 & 9.4 && 47.2 & 61.3 & 14.1 \\
    Pl. 10\% & 53.4 & 52.1 & -1.3  && 46.9 & 48.6 & 1.7 \\
    Pl.  1\% & 46.5 & 46.3 & -0.2 && 39.2 & 41.3 & 2.1 \\
    VOC07     & 88.9 & 89.2 & 0.3 && 84.9 & 88.0 & 3.1 \\
    FLOWERS     & 95.6 & 96.4 & 0.8 && 87.9 & 89.1 & 1.2 \\
    \midrule
    Average $\Delta$     & & & 2.4 && & & \bf 5.6 \\
    \bottomrule
  \end{tabular}
  \label{tab:knn}
\end{table}

\begin{table}[t]
  \centering
  \setlength{\tabcolsep}{.7em}
	\caption{
\textbf{ImageNet classification with different pretraining.}
Top-1 accuracy on ImageNet for supervised ViT-B/16 models using different pretrainings or using an additional pretrained convnet to guide the training.
The methods use different image resolution (``res.'') and training procedure (``tr. proc.''), i.e., data augmentation and optimization.
``MPP'' is \textit{Masked Patch Prediction}.
}
    \begin{tabular}{@{} lc c ccc @{}}
    \toprule
    \multicolumn{2}{c}{Pretraining} && & &  \\
    \cmidrule{1-2}
    method & data && res. & tr. proc. & Top-1\\
    \midrule
    \multicolumn{4}{@{}l}{\textit{Pretrain on additional data}}\\
	  MMP        &  JFT-300M &&384&\cite{dosovitskiy2020image}  & 79.9 \\
	  Supervised & JFT-300M  &&384&\cite{dosovitskiy2020image}  & 84.2 \\
    \midrule
    \multicolumn{4}{@{}l}{\textit{Train with additional model}}\\
	  Rand. init. & - &&224&\cite{touvron2020training}& 83.4 \\
    \midrule
    \multicolumn{4}{@{}l}{\textit{No additional data nor model}}\\
	  Rand. init. & -       &&224 &\cite{dosovitskiy2020image} & 77.9 \\
	  Rand. init. & -        &&224&\cite{touvron2020training}   & 81.8 \\
	  Supervised & ImNet &&224&\cite{touvron2020training}   & 81.9 \\
    \rowcolor{Light}
	  \OURS & ImNet      &&224&\cite{touvron2020training}   & 82.8 \\
    \bottomrule
  \end{tabular}
  \label{tab:finetune}
\end{table}

\paragraph{Self-supervised ImageNet pretraining of ViT.}
In this experiment, we study the impact of pretraining a supervised ViT model with our method.
In Tab.~\ref{tab:finetune}, we compare the performance of supervised ViT models that are initialized with different pretraining or guided during training with an additional pretrained convnet.
The first set of models are pretrained with and without supervision on the large curated dataset composed of 300M images.
The second set of models are trained with hard knowledge distillation from a pretrained supervised RegNetY~\cite{radosavovic2020designing}.
The last set of models do not use any additional data nor models, and are initialized either randomly or after a pretraining with \OURS on ImageNet.
Compare to random initialization, pretraining with \OURS leads to a performance gain of +1\%.
This is not caused by a longer training since pretraining with supervision instead of \OURS does not improve performance.
Using self-supervised pretraining reduces the gap with models pretrained on extra data or distilled from a convnet.

\paragraph{Low-shot learning on ImageNet.}
We evaluate the features obtained with \OURS applied on ViT-S on low-shot learning.
In Tab.~\ref{tab:semisup}, we report the validation accuracy of a logistic regression trained on frozen features (\textsc{frozen}) with 1\% and 10\% labels.
The logistic regression is trained with the \texttt{cyanure} library~\cite{mairal2019cyanure}.
When comparing models with a similar number of parameters and image/sec, we observe that our features are on par with state-of-the-art semi-supervised models.
Interestingly, this performance is obtained by training a multi-class logistic regression on \emph{frozen features, without data augmentation nor finetuning}.

\begin{table}[t]
	\caption{
\textbf{Low-shot learning on ImageNet with frozen ViT features.}
We train a logistic regression on frozen features (\textsc{frozen}).
Note that this \textsc{frozen} evaluation is performed \emph{without any finetuning nor data augmentation}.
We report top-1 accuracy.
For reference, we show previously published results that uses finetuning and semi-supervised learning.}
    \centering
\small
    \begin{tabular}{@{}l l c c c@{}}
	    \toprule
        & && \multicolumn{2}{c}{Top 1}\\
        Method & Arch & Param. &  1\% &  10\% \\
\midrule
\multicolumn{5}{@{}l@{}}{\textit{Self-supervised pretraining with finetuning}}\\
        UDA~\cite{xie2020unsupervised} & RN50 & 23 & -- & 68.1 \\
        SimCLRv2~\cite{chen2020big} & RN50 & 23 & 57.9 & 68.4 \\
        BYOL~\cite{grill2020bootstrap} & RN50 & 23 & 53.2 & 68.8 \\
        SwAV~\cite{caron2020unsupervised} & RN50 & 23 & 53.9 & 70.2 \\
	SimCLRv2~\cite{chen2020exploring} & RN50w4 & 375 & 63.0 & 74.4\\
        BYOL~\cite{grill2020bootstrap} & RN200w2 & 250 & 71.2 & 77.7 \\
\midrule
\multicolumn{5}{@{}l@{}}{\textit{Semi-supervised methods}}\\
        SimCLRv2+KD~\cite{chen2020big} & RN50 & 23 & 60.0 & 70.5 \\
        SwAV+CT~\cite{assran2020recovering} & RN50 & 23 & -- & 70.8 \\
        FixMatch~\cite{sohn2020fixmatch} & RN50 & 23 & -- & 71.5 \\
        MPL~\cite{pham2020meta} & RN50 & 23 & -- & 73.9 \\
	SimCLRv2+KD~\cite{chen2020big} & RN152w3+SK & 794 & 76.6 & 80.9 \\
	\midrule
\multicolumn{5}{@{}l@{}}{\textit{Frozen self-supervised features}}\\
        \rowcolor{Light}
        \OURS\textsc{-frozen} & ViT-S/16 & 21 & 64.5 & 72.2 \\
	\bottomrule
    \end{tabular}
    \label{tab:semisup}
\end{table}

\begin{figure*}[t]
\centering
  \setlength{\tabcolsep}{3pt}
\begin{tabular}{ccc}
\includegraphics[width=0.35\linewidth]{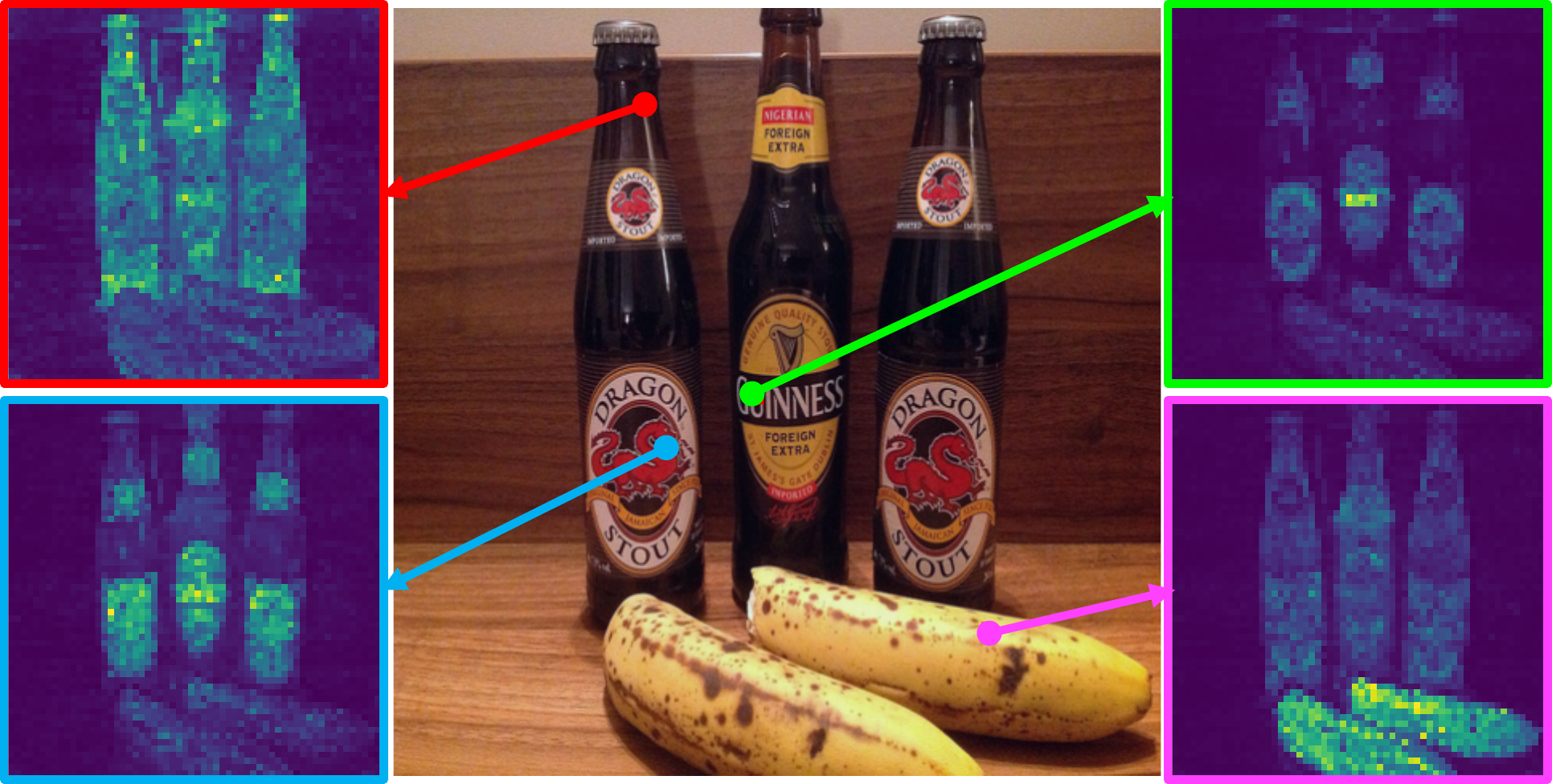}&
\includegraphics[width=0.35\linewidth]{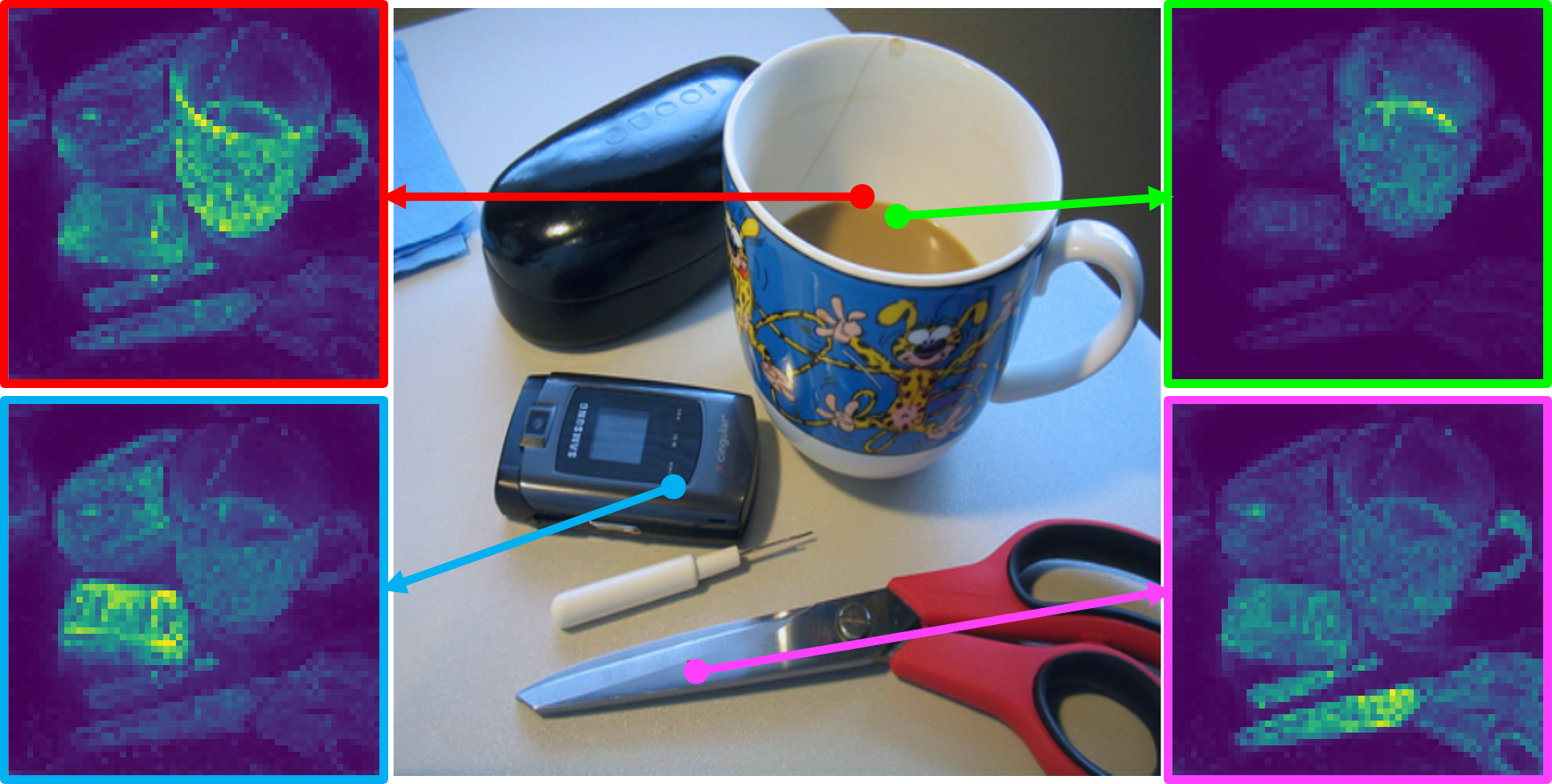}&
\includegraphics[width=0.35\linewidth]{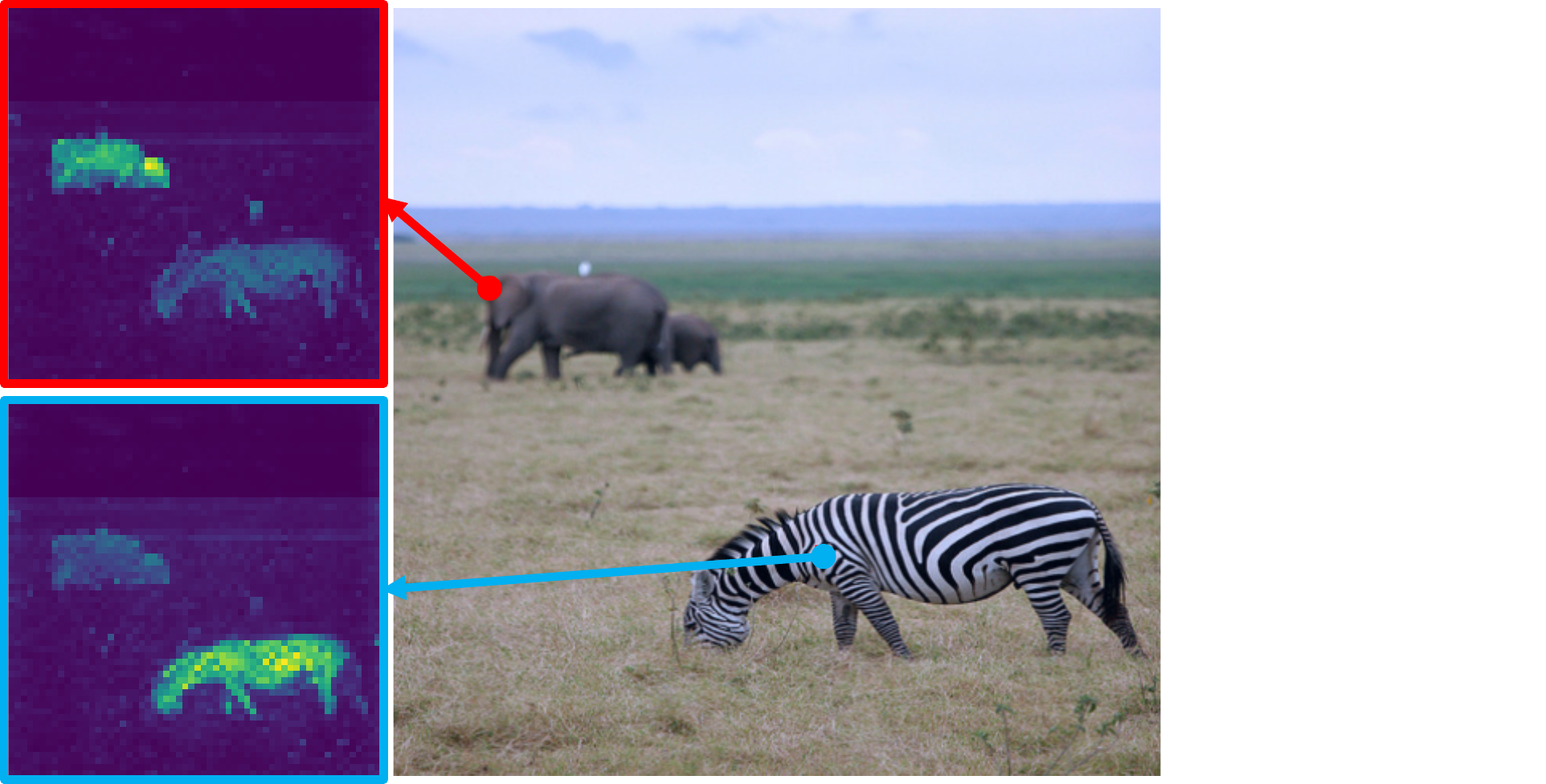}\\
\end{tabular}
\caption{
	\textbf{Self-attention for a set of reference points.}
We visualize the self-attention module from the last block of a ViT-S/8 trained with \OURS.
The network is able to separate objects, though it has been trained with no supervision at all.
}
\label{fig:pointing}
\end{figure*}

\subsection{Methodology Comparison}
\label{ap:comp}
We compare the performance of different self-supervised frameworks, MoCo-v2~\cite{chen2020improved}, SwAV~\cite{caron2020unsupervised} and BYOL~\cite{grill2020bootstrap} when using convnet or ViT.
In Tab.~\ref{tab:vit_cnn}, we see that when trained with ResNet-50 (convnet), \OURS performs on par with SwAV and BYOL.
However, \OURS unravels its potential with ViT, outperforming MoCo-v2, SwAV and BYOL by large margins (+4.3\% with linear and +6.2\% with k-NN evaluations).
In the rest of this section, we perform ablations to better understand the performance of \OURS applied to ViT.
In particular, we provide a detailed comparison with methods that either use a momentum encoder, namely MoCo-v2 and BYOL, and methods that use multi-crop, namely SwAV.
\begin{table}[t]
\centering
  \caption{
	  \textbf{Methodology comparison for DEIT-small and ResNet-50.}
    We report ImageNet linear and $k$-NN evaluations validation accuracy after 300 epochs pre-training.
    All numbers are run by us and match or outperform published results.
  }
  \begin{tabular}{@{}ll cc c cc@{}}
    \toprule
  && \multicolumn{2}{c}{ResNet-50} && \multicolumn{2}{c}{ViT-small} \\
    \cmidrule{3-4}
    \cmidrule{6-7}
	  Method && Linear & $k$-NN && Linear & $k$-NN \\
    \midrule
	  MoCo-v2 && 71.1 & 62.9 && 71.6 & 62.0 \\
	  BYOL && 72.7 & 65.4 && 71.4 & 66.6 \\
	  SwAV && 74.1 & 65.4 && 71.8 & 64.7 \\
    \midrule
	  \OURS && \bf 74.5 & \bf 65.6  && \bf 76.1 & \bf 72.8 \\
    \bottomrule
  \end{tabular}
  \label{tab:vit_cnn}
\end{table}

\paragraph{Relation to MoCo-v2 and BYOL.}
In Tab.~\ref{tab:byol}, we present the impact of ablating components that differ between \OURS, MoCo-v2 and BYOL: the choice of loss, the predictor in the student head, the centering operation, the batch normalization in the projection heads, and finally, the multi-crop augmentation.
The loss in \OURS is a cross-entropy on sharpened softmax outputs (\texttt{CE}) while MoCo-v2 uses the InfoNCE contrastive loss (\texttt{INCE}) and BYOL a mean squared error on l2-normalized outputs (\texttt{MSE}).
No sharpening is applied with the \texttt{MSE} criterion.
Though, \OURS surprisingly still works when changing the loss function to \texttt{MSE}, but this significantly alters the performance (see rows (\rownumber{1}, \rownumber{2}) and (\rownumber{4}, \rownumber{9})).
We also observe that adding a predictor has little impact (\rownumber{1}, \rownumber{3}).
However, in the case of BYOL, the predictor is critical to prevent collapse (\rownumber{7}, \rownumber{8}) which is consistent with previous studies~\cite{chen2020exploring,grill2020bootstrap}.
Interestingly, we observe that the teacher output centering avoids collapse without predictor nor batch normalizations in BYOL (\rownumber{7}, \rownumber{9}), though with a significant performance drop which can likely be explained by the fact that our centering operator is designed to work in combination with sharpening.
Finally, we observe that multi-crop works particularly well with \OURS and MoCo-v2, removing it hurts performance by $2-4\%$ (\rownumber{1} versus \rownumber{4} and, \rownumber{5} versus \rownumber{6}).
Adding multi-crop to BYOL does not work out-of-the-box (\rownumber{7}, \rownumber{10}) as detailed in Appendix~\ref{ap:mc} and further adaptation may be required.

\begin{table}[t]
  \centering
\small
  \setlength{\tabcolsep}{4.3pt}
	\caption{
    \textbf{Relation to MoCo-v2 and BYOL.}
	We ablate the components that differ between \OURS, MoCo-v2 and BYOL: the loss function (cross-entropy, \texttt{CE}, versus InfoNCE, \texttt{INCE}, versus mean-square error, \texttt{MSE}), the multi-crop training, the centering operator, the batch normalization in the projection heads and the student predictor. Models are run for 300 epochs with ViT-S/16. We report top-1 accuracy on ImageNet linear evaluation.
  }
  \begin{tabular}{@{}llcccccc@{}}
    \toprule
	  & Method & Loss & multi-crop & Center. & BN & Pred. & Top-1 \\
    \midrule
\rownumber{1}&	  \OURS & \texttt{CE} & \checkmark & \checkmark & & & 76.1 \\
\rownumber{2}&  -- & \texttt{MSE} & \checkmark & \checkmark & & & 62.4 \\
\rownumber{3}&  -- & \texttt{CE} & \checkmark & \checkmark & & \checkmark & 75.6 \\
\rownumber{4}&  -- & \texttt{CE} & & \checkmark & & & 72.5 \\
\arrayrulecolor{black!20}\midrule
\rownumber{5}&  MoCov2 & \texttt{INCE} & & & \checkmark &  & 71.4 \\
\rownumber{6}&   & \texttt{INCE} & \checkmark & & \checkmark &  & 73.4 \\
\arrayrulecolor{black!20}\midrule
\rownumber{7}&  BYOL & \texttt{MSE} & & & \checkmark & \checkmark & 71.4 \\
\rownumber{8}&  -- & \texttt{MSE} & & & \checkmark &  & \pzo0.1 \\
\rownumber{9}&  -- & \texttt{MSE} & & \checkmark &  &  & 52.6 \\
\rownumber{10}&  -- & \texttt{MSE} & \checkmark  & & \checkmark & \checkmark & 64.8 \\
   \arrayrulecolor{black} \bottomrule
  \end{tabular}
\label{tab:byol}
\end{table}

\begin{table}[t]
  \centering
	\caption{\textbf{Relation to SwAV.}
We vary the operation on the teacher output between centering, a softmax applied over the batch dimension and the Sinkhorn-Knopp algorithm. 
We also ablate the Momentum encoder by replacing it with a hard copy of the student with a stop-gradient as in SwAV.
Models are run for 300 epochs with ViT-S/16. We report top-1 accuracy on ImageNet linear evaluation.
	}
  \begin{tabular}{@{}llccc@{}}
    \toprule
	 & Method & Momentum & Operation & Top-1 \\
    \midrule
\rownumber{1}&  \OURS & \checkmark & \texttt{Centering} & 76.1 \\
\rownumber{2}&  -- & \checkmark & \texttt{Softmax(batch)} & 75.8 \\
\rownumber{3}&  -- & \checkmark & \texttt{Sinkhorn-Knopp} & 76.0 \\
\rownumber{4}&  -- & & \texttt{Centering} & \pzo0.1 \\
\rownumber{5}&  -- & & \texttt{Softmax(batch)} & 72.2 \\
\rownumber{6}&  SwAV & & \texttt{Sinkhorn-Knopp} & 71.8 \\
    \bottomrule
  \end{tabular}
  \label{tab:swav}
\end{table}

\paragraph{Relation to SwAV.}
In Tab.~\ref{tab:swav}, we evaluate the differences between \OURS and SwAV: the presence of the momentum encoder and the operation on top of the teacher output.
In absence of the momentum, a copy of the student with a stop-gradient is used.
We consider three operations on the teacher output: \texttt{Centering}, \texttt{Sinkhorn-Knopp} or a \texttt{Softmax} along the batch axis.
The \texttt{Softmax} is similar to a single Sinkhorn-Knopp iteration as detailed in the next paragraph.
First, these ablations show that using a momentum encoder significantly improves the performance for ViT (\rownumber{3} versus \rownumber{6}, and \rownumber{2} versus \rownumber{5}).
Second, the momentum encoder also avoids collapse when using only centering (row \rownumber{1}).
In the absence of momentum, centering the outputs does not work (\rownumber{4}) and more advanced operations are required (\rownumber{5}, \rownumber{6}).
Overall, these ablations highlight the importance of the momentum encoder, not only for performance but also to stabilize training, removing the need for normalization beyond centering.

\paragraph{Details on the \texttt{Softmax(batch)} variant.}
The iterative Sinkhorn-Knopp algorithm~\cite{cuturi2013sinkhorn} used in SwAV~\cite{caron2020unsupervised} is implemented simply with the following PyTorch style code.
\begin{python}
# x is n-by-K
# tau is Sinkhorn regularization param
x = exp(x / tau)
for _ in range(num_iters): # 1 iter of Sinkhorn
	# total weight per dimension (or cluster)
	c = sum(x, dim=0, keepdim=True) 
	x /= c

	# total weight per sample
	n = sum(x, dim=1, keepdim=True) 
	# x sums to 1 for each sample (assignment)
	x /= n 
\end{python}
When performing a single Sinkhorn iteration (\texttt{num\_iters=1}) the implementation can be highly simplified into only two lines of code, which is our \texttt{softmax(batch)} variant:
\begin{python}
x = softmax(x / tau, dim=0)
x /= sum(x, dim=1, keepdim=True) 
\end{python}
We have seen in Tab.~\ref{tab:swav} that this highly simplified variant of SwAV works competitively with SwAV.
Intuitively, the \texttt{softmax} operation on the batch axis allows to select for each dimension (or ``cluster'') its best matches in the batch.

\paragraph{Validating our implementation.}
We observe in Tab.~\ref{tab:vit_cnn} that our reproduction of BYOL, MoCo-v2, SwAV matches or outperforms the corresponding published numbers with ResNet-50.
Indeed, we obtain $72.7\%$ for BYOL while \cite{grill2020bootstrap} report $72.5\%$ in this $300$-epochs setting.
We obtain $71.1\%$ for MoCo after $300$ epochs of training while \cite{chen2020improved} report $71.1\%$ after $800$ epochs of training.
Our improvement compared to the implementation of \cite{chen2020improved} can be explained by the use of a larger projection head (3-layer, use of batch-normalizations and projection dimension of $256$).

\paragraph{Relation to other works.}
\OURS is also related to UIC~\cite{chen2020unsupervised} that use outputs from the previous epoch as hard pseudo-labels for ``unsupervised classification''.
However, we use centering to prevent collapse while UIC resorts to balance sampling techniques as in~\cite{caron2018deep}.
Our work can be interpreted as a soft UIC variant with momentum teacher.

The concurrent work CsMI~\cite{xu2021seed} also exhibits strong performance with simple k-NN classifiers on ImageNet, even with convnets.
As \OURS, CsMI combines a momentum network and multi-crop training, which we have seen are both crucial for good k-NN performance in our experiments with ViTs.
We believe studying this work would help us identifying more precisely the components important for good $k$-NN performance and leave this investigation for future work.

\subsection{Projection Head}
\label{ap:projhead}
Similarly to other self-supervised frameworks, using a projection head~\cite{chen2020simple} improves greatly the accuracy of our method.
The projection head starts with a $n$-layer multi-layer perceptron (MLP).
The hidden layers are 2048d and are with gaussian error linear units (GELU) activations.
The last layer of the MLP is without GELU.
Then we apply a $\ell_2$ normalization and a weight normalized fully connected layer~\cite{chen2020exploring,salimans2016weight} with $K$ dimensions.
This design is inspired from the projection head with a ``prototype layer'' used in SwAV~\cite{caron2020unsupervised}.
We do not apply batch normalizations.

\paragraph{BN-free system.}
Unlike standard convnets, ViT architectures do not use batch normalizations (BN) by default.
\begin{table}[h!]
\vspace{-0.8em}
  \centering
  \begin{tabular}{@{}l c c@{}}
	  ViT-S, 100 epochs & heads w/o BN & heads w/ BN \\
    \midrule
	  $k$-NN top-1 & \colorbox{Light}{69.7} & 68.6 \\
  \end{tabular}
\vspace{-0.8em}
\end{table}
Therefore, when applying \OURS to ViT we do not use any BN also in the projection heads.
In this table we evaluate the impact of adding BN in the heads.
We observe that adding BN in the projection heads has little impact, showing that BN is not important in our framework.
\emph{Overall, when applying \OURS to ViT, we do not use any BN anywhere, making the system entirely BN-free.}
This is a great advantage of \OURS + ViT to work at state-of-the-art performance without requiring any BN.
Indeed, training with BN typically slows down trainings considerably, especially when these BN modules need to be synchronized across processes~\cite{he2020momentum,caron2020unsupervised,caron2019unsupervised,grill2020bootstrap}.
\begin{figure}[h!]
\centering
\includegraphics[width=\linewidth]{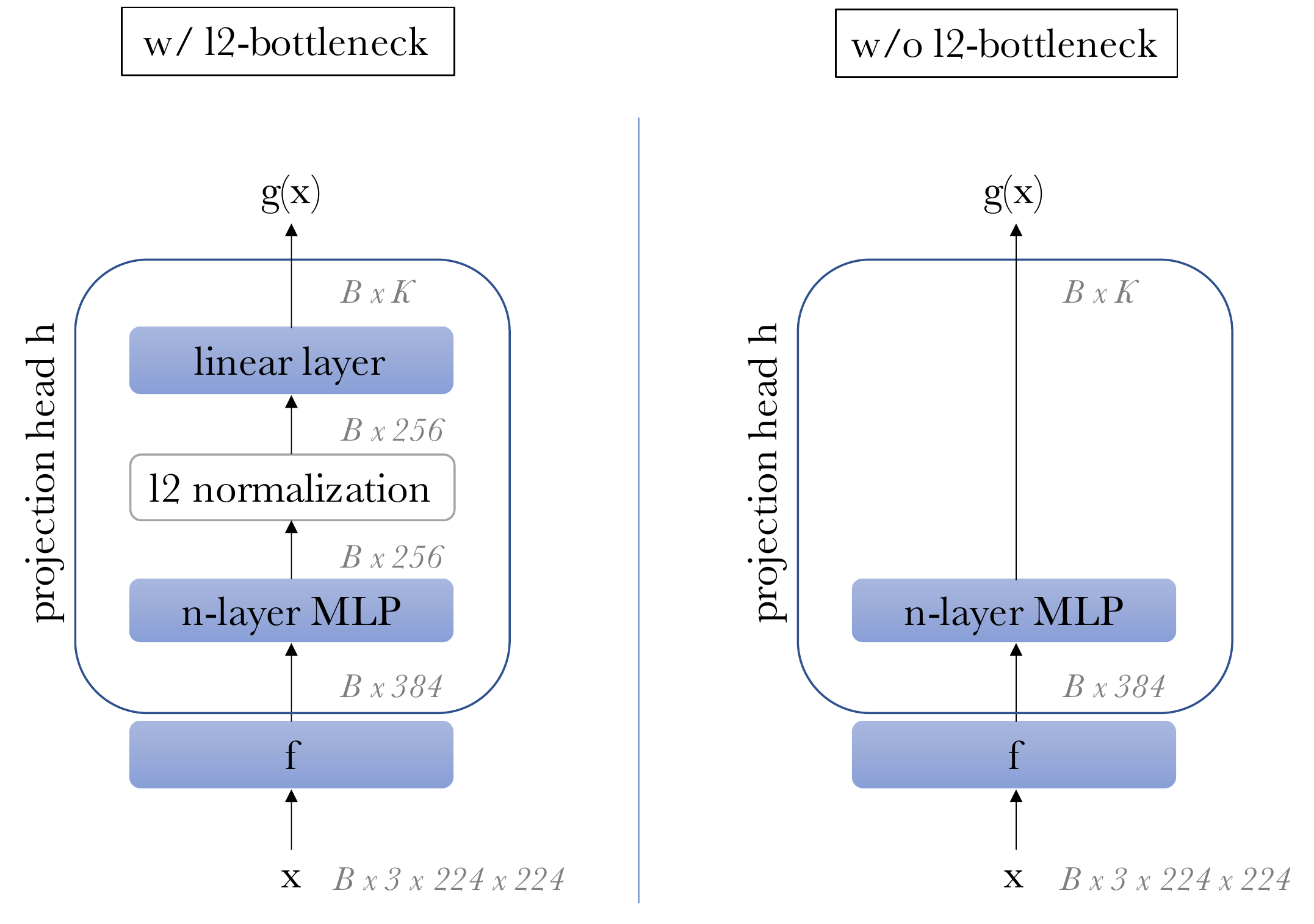}
	\caption{
    \textbf{Projection head design w/ or w/o l2-norm bottleneck.}
  }
\label{fig:proj}
\end{figure}

\paragraph{L2-normalization bottleneck in projection head.}
We illustrate the design of the projection head with or without l2-normalization bottleneck in Fig.~\ref{fig:proj}.
\begin{table}[h!]
\vspace{-0.8em}
  \centering
  \begin{tabular}{@{}l c c c c@{}}
	  \# proj. head linear layers & $1$ & $2$ & $3$ & $4$\\
    \midrule
	  w/ l2-norm bottleneck & -- & 62.2 & 68.0 & \colorbox{Light}{69.3} \\
	  w/o l2-norm bottleneck & 61.6 & 62.9 & 0.1 & 0.1 \\
  \end{tabular}
\vspace{-0.8em}
\end{table}
We evaluate the accuracy of \OURS models trained with or without l2-normalization bottleneck and we vary the number of linear layers in the projection head.
With l2 bottleneck, the total number of linear layers is $n + 1$ ($n$ from the MLP and $1$ from the weight normalized layer) while without bottleneck the total number of linear layers is $n$ in the head.
In this table, we report ImageNet top-1 $k$-NN evaluation accuracy after 100 epochs pre-training with ViT-S/16.
The output dimensionality $K$ is set to $4096$ in this experiment.
We observe that \OURS training fails without the l2-normalization bottleneck when increasing the depth of the projection head.
L2-normalization bottleneck stabilizes the training of \OURS with deep projection head.
We observe that increasing the depth of the projection head improves accuracy.
Our default is to use a total of 4 linear layers: 3 are in the MLP and one is after the l2 bottleneck.

\paragraph{Output dimension.}
In this table, we evaluate the effect of varying the output dimensionality $K$.
\begin{table}[h!]
\vspace{-0.8em}
  \centering
  \begin{tabular}{@{}l c c c c c@{}}
	  $K$ & 1024 & 4096 & 16384 & 65536 & 262144 \\
    \midrule
	  $k$-NN top-1 & 67.8 & 69.3 & 69.2 & \colorbox{Light}{69.7} & 69.1 \\
  \end{tabular}
\vspace{-0.8em}
\end{table}
We observe that a large output dimensionality improves the performance.
We note that the use of l2-normalization bottleneck permits to use a large output dimension with a moderate increase in the total number of parameters.
Our default is to use $K$ equals to 65536 and $d=256$ for the bottleneck.

\paragraph{GELU activations.}
By default, the activations used in ViT are gaussian error linear units (GELU).
\begin{table}[h!]
\vspace{-0.8em}
  \centering
  \begin{tabular}{@{}l c c@{}}
	  ViT-S, 100 epochs & heads w/ GELU & heads w/ ReLU \\
    \midrule
	  $k$-NN top-1 & \colorbox{Light}{69.7} & 68.9 \\
  \end{tabular}
\vspace{-0.8em}
\end{table}
Therefore, for consistency within the architecture, we choose to use GELU also in the projection head.
We evaluate the effect of using ReLU instead of GELU in this table and observe that changing the activation unit to ReLU has relatively little impact.

\subsection{Additional Ablations}
\label{ap:ablations}
We have detailed in the main paper that the combination of centering and sharpening is important to avoid collapse in \OURS.
We ablate the hyperparameters for these two operations in the following.
We also study the impact of training length and some design choices for the ViT networks.

\paragraph{Online centering.}
We study the impact of the smoothing parameters in the update rule for the center $c$ used in the output of the teacher network.
\begin{table}[h!]
\vspace{-0.8em}
  \centering
  \begin{tabular}{@{}l c c c c@{}}
	  $m$ & 0 & 0.9 & 0.99 & 0.999 \\
    \midrule
	  $k$-NN top-1 & 69.1 & \colorbox{Light}{69.7} & 69.4 & 0.1 \\
  \end{tabular}
\vspace{-0.8em}
\end{table}
The convergence is robust to a wide range of smoothing, and the model only collapses when the update is too slow, i.e., $m = 0.999$.

\paragraph{Sharpening.}
We enforce sharp targets by tuning the teacher softmax temperature parameter $\tau_t$.
In this table, we observe that a temperature lower than $0.06$ is required to avoid collapse.
\begin{table}[h!]
\vspace{-0.8em}
  \centering
  \setlength{\tabcolsep}{4pt}
  \begin{tabular}{@{}l c c c c c c@{}}
	  $\tau_t$ & $0$ & $0.02$ & $0.04$ & $0.06$ & $0.08$ & $0.04 \rightarrow 0.07$ \\
    \midrule
	  $k$-NN top-1 & 43.9 & 66.7 & 69.6 & 68.7 & 0.1 & \colorbox{Light}{69.7} \\
  \end{tabular}
\vspace{-0.8em}
\end{table}
When the temperature is higher than $0.06$, the training loss consistently converges to $ln(K)$.
However, we have observed that using higher temperature than $0.06$ does not collapse if we start the training from a smaller value and increase it during the first epochs.
In practice, we use a linear warm-up for $\tau_t$ from $0.04$ to $0.07$ during the first $30$ epochs of training.
Finally, note that $\tau \rightarrow 0$ (extreme sharpening) correspond to the \texttt{argmax} operation and leads to one-hot hard distributions.

\paragraph{Longer training.}
We observe in this table that longer training improves the performance of \OURS applied to ViT-Small.
\vspace{-0.8em}
\begin{table}[h!]
  \centering
\begin{tabular}{@{}l c c c@{}}
	\OURS ViT-S & 100-ep & 300-ep & 800-ep \\
\midrule
	$k$-NN top-1 & 70.9 & 72.8 & \colorbox{Light}{74.5} \\
\end{tabular}
\vspace{-0.8em}
\end{table}
This observation is consistent with self-supervised results obtained with convolutional architectures~\cite{chen2020simple}.
We note that in our experiments with BYOL on ViT-S, training longer than $300$ epochs has been leading to worse performance compare our $300$ epochs run.
For this reason we report BYOL for 300 epochs in Tab.~\ref{tab:sota} while SwAV, MoCo-v2 and DINO are trained for 800 epochs.

\paragraph{The teacher outperforms the student.}
We have shown in Fig.~\ref{fig:mom} that the momentum teacher outperforms the student with ViT and we show in this Figure that it is also the case with ResNet-50.
\begin{figure}[h!]
\vspace{-0.8em}
\centering
\includegraphics[width=0.48\linewidth]{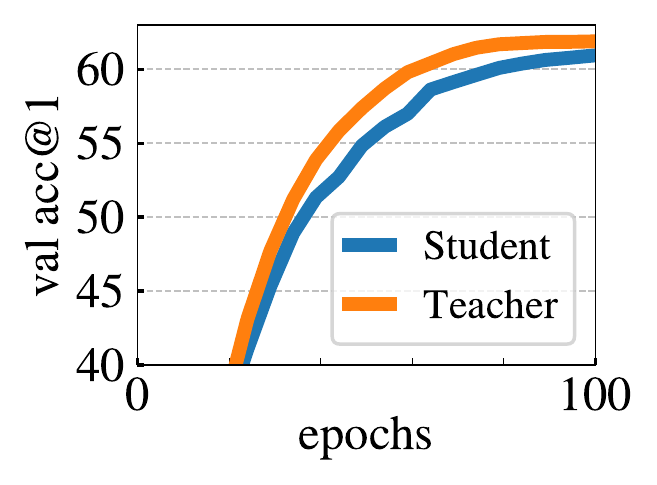}
\vspace{-0.8em}
\end{figure}
The fact that the teacher continually outperforms the student further encourages the interpretation of \OURS as a form of Mean Teacher~\cite{tarvainen2017mean} self-distillation.
Indeed, as motivated in Tarvainen et al.~\cite{tarvainen2017mean}, weight averaging usually produces a better model than the individual models from each iteration~\cite{polyak1992acceleration}.
By aiming a target obtained with a teacher better than the student, the student's representations improve.
Consequently, the teacher also improves since it is built directly from the student weights.

\paragraph{Self-attention maps from supervised versus self-supervised learning.}
We evaluate the masks obtained by thresholding the self-attention maps to keep 80\% of the mass.
\begin{table}[h!]
\vspace{-0.8em}
  \centering
  \begin{tabular}{@{}l c@{}}
	  \toprule
	  ViT-S/16 weights & \\
	  \midrule
	  Random weights & 22.0 \\
	  Supervised & 27.3 \\
	  \midrule
	  \OURS & 45.9 \\
	  \OURS w/o multicrop & 45.1 \\
	  MoCo-v2 & 46.3 \\
	  BYOL & 47.8 \\
	  SwAV & 46.8 \\
	\bottomrule
  \end{tabular}
\vspace{-0.8em}
\end{table}
We compare the Jaccard similarity between the ground truth and these masks on the validation images of PASCAL VOC12 dataset for different ViT-S trained with different frameworks.
The properties that self-attention maps from ViT explicitly contain the scene layout and, in particular, object boundaries is observed across different self-supervised methods.

\paragraph{Impact of the number of heads in ViT-S.}
We study the impact of the number of heads in ViT-S on the accuracy and throughput (images processed per second at inference time on a singe V100 GPU).
\begin{table}[h!]
\vspace{-0.8em}
  \centering
  \begin{tabular}{@{}l c c c c c@{}}
	  \# heads & dim & dim/head & \# params & im/sec & $k$-NN \\
    \midrule
    \rowcolor{Light}
	  6 & 384 & 64 & 21 & 1007 & 72.8 \\
	  8 & 384 & 48 & 21 & 971 & 73.1 \\
	  12 & 384 & 32 & 21 & 927 & 73.7 \\
	  16 & 384 & 24 & 21 & 860 & 73.8 \\
  \end{tabular}
\vspace{-0.8em}
\end{table}
We find that increasing the number of heads improves the performance, at the cost of a slighlty worse throughput.
In our paper, all experiments are run with the default model DeiT-S~\cite{touvron2020training}, i.e. with $6$ heads only.

\subsection{Multi-crop}
\label{ap:mc}
In this Appendix, we study a core component of \OURS: multi-crop training~\cite{caron2020unsupervised}.

\paragraph{Range of scales in multi-crop.}
For generating the different views, we use the \texttt{RandomResizedCrop} method from \texttt{torchvision.transforms} module in PyTorch.
\begin{table}[h!]
\vspace{-0.8em}
\centering
  \begin{tabular}{@{}l c c c c c@{}}
	  (0.05, $s$), ($s$, 1), $s$: & 0.08 & 0.16 & 0.24 & 0.32 & 0.48 \\
    \midrule
	  $k$-NN top-1 & 65.6 & 68.0 & 69.7 & 69.8 & 69.5 \\
  \end{tabular}
\vspace{-0.8em}
\end{table}
We sample two global views with scale range $(s, 1)$ before resizing them to $224^2$ and $6$ local views with scale sampled in the range $(0.05, s)$ resized to $96^2$ pixels.
Note that we arbitrarily choose to have non-overlapping scaling range for the global and local views following the original design of SwAV.
However, the ranges could definitely be overlapping and experimenting with finer hyperparameters search could lead to a more optimal setting.
In this table, we vary the parameter $s$ that controls the range of scales used in multi-crop and find the optimum to be around $0.3$ in our experiments.
We note that this is higher than the parameter used in SwAV which is of $0.14$.

\paragraph{Multi-crop in different self-supervised frameworks.}
We compare different recent self-supervised learning frameworks, namely MoCo-v2~\cite{chen2020improved}, BYOL~\cite{grill2020bootstrap} and SwAV~\cite{caron2020unsupervised} with ViT-S/16 architecture.
\begin{table}[h!]
\vspace{-0.8em}
  \centering
\begin{tabular}{@{}l cc c cc@{}}
\toprule
crops	& \multicolumn{2}{c}{$2 \times 224^2$} && \multicolumn{2}{c}{$2 \times 224^2 + 6 \times 96^2$} \\
    \cmidrule{2-3}
    \cmidrule{5-6}
	eval & $k$-NN & linear && $k$-NN & linear  \\
\midrule
	BYOL & 66.6 & 71.4 && 59.8 & 64.8 \\
	SwAV & 60.5 & 68.5 && 64.7 & 71.8 \\
	MoCo-v2 & 62.0 & 71.6 && 65.4 & 73.4 \\
        \rowcolor{Light}
	\OURS & \bf 67.9 & \bf 72.5 && \bf 72.7 & \bf 75.9 \\
\bottomrule
\vspace{-0.8em}
\end{tabular}
\end{table}
For fair comparisons, all models are pretrained either with two $224^2$ crops or with multi-crop~\cite{caron2020unsupervised} training, i.e. two $224^2$ crops and six $96^2$ crops for each image.
We report $k$-NN and linear probing evaluations after 300 epochs of training.
Multi-crop does not benefit all frameworks equally, which has been ignored in benchmarks considering only the two crops setting~\cite{chen2020exploring}.
The effectiveness of multi-crop depends on the considered framework, which positions multi-crop as a core component of a model and not a simple ``add-ons'' that will boost any framework the same way.
Without multi-crop, \OURS has better accuracy than other frameworks, though by a moderate margin (1\%).
Remarkably, \OURS benefits the most from multi-crop training ($+3.4\%$ in linear eval).
Interestingly, we also observe that the ranking of the frameworks depends on the evaluation protocol considered.

\paragraph{Training BYOL with multi-crop.}
When applying multi-crop to BYOL with ViT-S, we observe the transfer performance is higher than the baseline without multi-crop for the first training epochs.
\begin{figure}[h]
\vspace{-0.8em}
\centering
\includegraphics[width=0.5\linewidth]{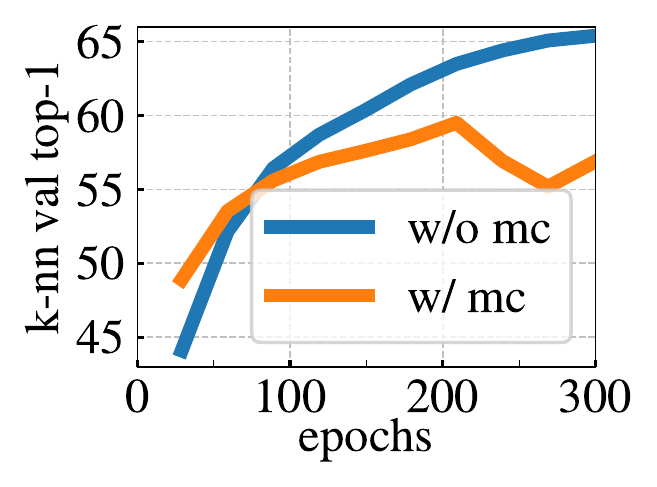}
\vspace{-0.8em}
\end{figure}
However, the transfer performance growth rate is slowing down and declines after a certain amount of training.
We have performed learning rate, weight decay, multi-crop parameters sweeps for this setting and systematically observe the same pattern.
More precisely, we experiment with \{$1e^{-5}$, $3e^{-5}$, $1e^{-4}$, $3e^{-4}$, $1e^{-3}$, $3e^{-3}$\} for learning rate base values, with \{$0.02$, $0.05$, $0.1$\} for weight decay and with different number of small crops: \{2, 4, 6\}.
All our runs are performed with synchronized batch normalizations in the heads.
When using a low learning rate, we did not observe the performance break point, i.e. the transfer performance was improving continually during training, but the overall accuracy was low.
We have tried a run with multi-crop training on ResNet-50 where we also observe the same behavior.
Since integrating multi-crop training to BYOL is not the focus of this study we did not push that direction further.
However, we believe this is worth investigating why multi-crop does not combine well with BYOL in our experiments and leave this for future work.

\subsection{Evaluation Protocols}
\subsubsection{$k$-NN classification}
Following the setting of Wu~\etal~\cite{wu2018unsupervised}, we evaluate the quality of features with a simple weighted $k$ Nearest Neighbor classifier.
We freeze the pretrained model to compute and store the features of the training data of the downstream task.
To classify a test image $x$, we compute its representation and compare it against all stored training features $T$.
The representation of an image is given by the output \texttt{[CLS]} token: it has dimensionality $d=384$ for ViT-S and $d=768$ for ViT-B.
The top $k$ NN (denoted $\mathcal{N}_k$) are used to make a prediction via weighted voting.
Specifically, the class $c$ gets a total weight of $\sum_{i \in \mathcal{N}_k} \alpha_i \mathbf{1}_{c_i = c}$, where $\alpha_i$ is a contribution weight.
We use $\alpha_i = \exp(T_i x / \tau)$ with $\tau$ equals to $0.07$ as in~\cite{wu2018unsupervised} which we do not tune.
We evaluate different values for $k$ and find that $k=20$ is consistently leading to the best accuracy across our runs.
This evaluation protocol does not require hyperparameter tuning, nor data augmentation and can be run with only one pass over the downstream dataset.

\subsubsection{Linear classification}
Following common practice in self-supervised learning, we evaluate the representation quality with a linear classifier.
The projection head is removed, and we train a supervised linear classifier on top of frozen features.
This linear classifier is trained with SGD and a batch size of $1024$ during $100$ epochs on ImageNet.
We do not apply weight decay.
For each model, we sweep the learning rate value.
During training, we apply only random resizes crops (with default parameters from PyTorch \texttt{RandomResizedCrop}) and horizontal flips as data augmentation.
We report central-crop top-1 accuracy.
When evaluating convnets, the common practice is to perform global average pooling on the final feature map before the linear classifier.
In the following, we describe how we adapt this design when evaluating ViTs.

\paragraph{ViT-S representations for linear eval.}
Following the \emph{feature-based} evaluations in BERT~\cite{devlin2018bert}, we concatenate the \texttt{[CLS]} tokens from the $l$ last layers.
\begin{table}[h!]
\vspace{-0.8em}
\small
  \centering
  \begin{tabular}{@{}l c c c c@{}}
	  concatenate $l$ last layers & $1$ & $2$ & $4$ & $6$\\
    \midrule
	  representation dim & 384 & 768 & 1536 & 2304 \\
	  ViT-S/16 linear eval & 76.1 & 76.6 & \colorbox{Light}{77.0} & 77.0 \\
  \end{tabular}
\vspace{-0.8em}
\end{table}
We experiment with the concatenation of a different number $l$ of layers and similarly to~\cite{devlin2018bert} we find $l=4$ to be optimal.

\paragraph{ViT-B representations for linear eval.}
With ViT-B we did not find that concatenating the representations from the last $l$ layers to provide any performance gain, and consider the final layer only ($l=1$).
\begin{table}[h!]
\small
\vspace{-0.8em}
  \centering
  \setlength{\tabcolsep}{4pt}
  \begin{tabular}{@{}l c c@{}}
	  pooling strategy & \texttt{[CLS]} tok. & concatenate \texttt{[CLS]} tok.\\
	  & only &  and avgpooled patch tok. \\
    \midrule
	  representation dim & 768 & 1536 \\
	  ViT-B/16 linear eval & 78.0 & \colorbox{Light}{78.2} \\
  \end{tabular}
\vspace{-0.8em}
\end{table}
In this setting, we adapt the pipeline used in convnets with global average pooling on the output patch tokens.
We concatenate these pooled features to the final \texttt{[CLS]} output token.

\subsection{Self-Attention Visualizations}
\label{ap:visu}
\begin{figure*}
\centering
\setlength{\tabcolsep}{0.5pt}
\begin{tabular}{c cc ccc cc ccc cc c cc ccc cc ccc}
	&&& \multicolumn{3}{c}{\OURS} &&& \multicolumn{3}{c}{Supervised} &&& &&& \multicolumn{3}{c}{\OURS} &&& \multicolumn{3}{c}{Supervised} \\
    \cmidrule{4-6}
    \cmidrule{9-11}
    \cmidrule{17-19}
    \cmidrule{22-24}
\includegraphics[width=0.07\linewidth]{3478.png} &&&
\includegraphics[width=0.07\linewidth]{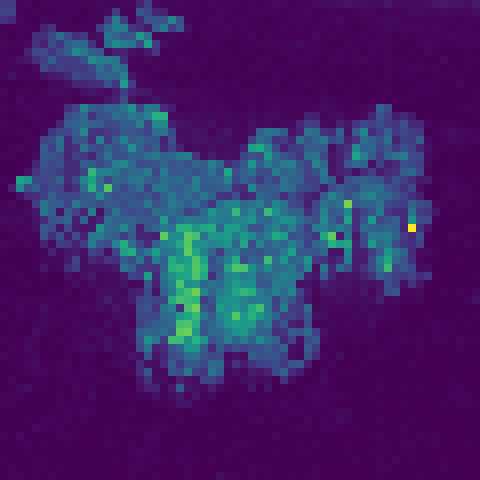} &
\includegraphics[width=0.07\linewidth]{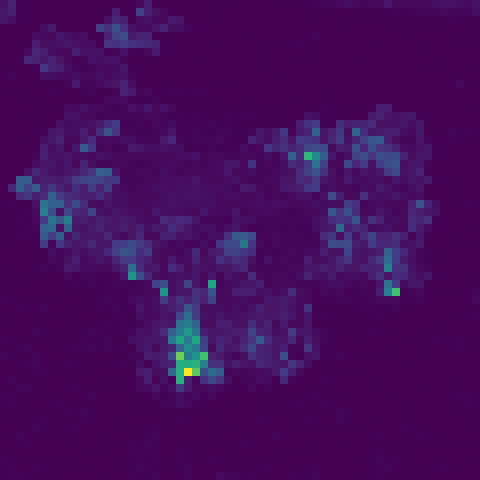} &
\includegraphics[width=0.07\linewidth]{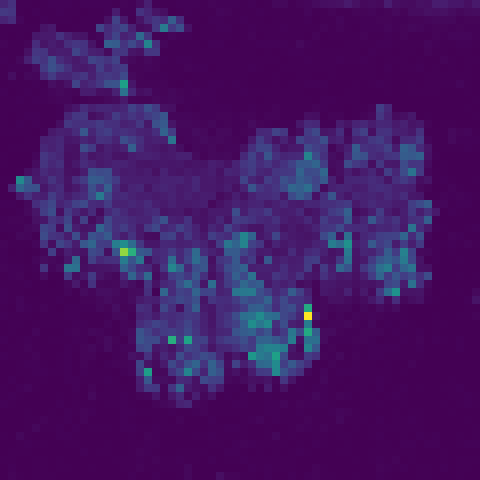} &&&
\includegraphics[width=0.07\linewidth]{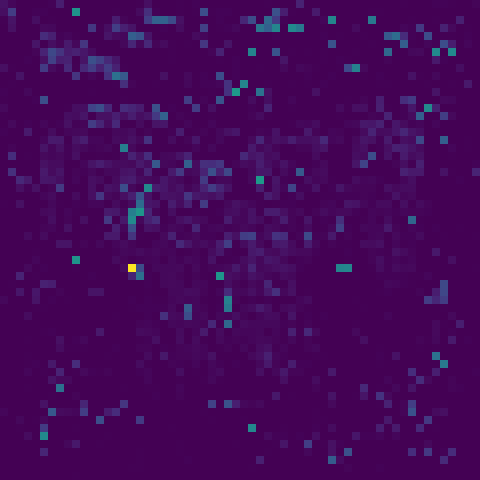} &
\includegraphics[width=0.07\linewidth]{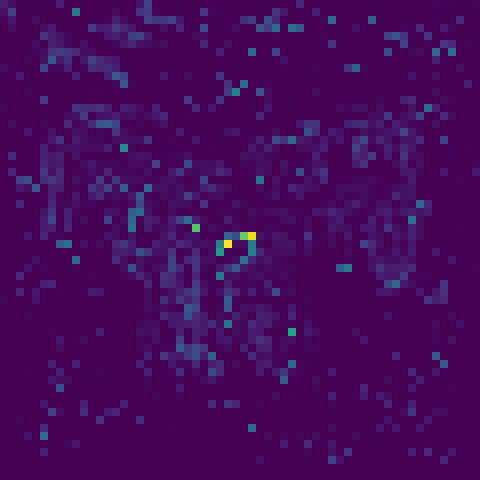} &
\includegraphics[width=0.07\linewidth]{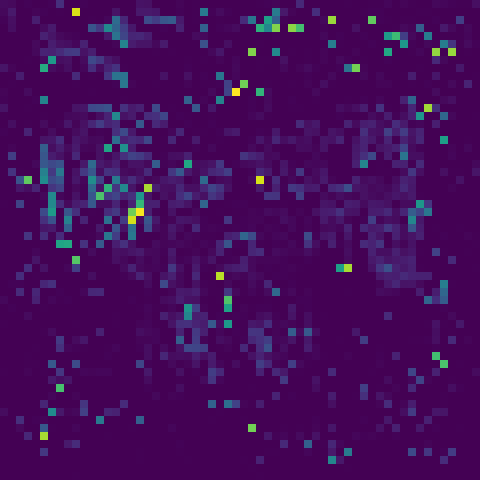}
&&&
\includegraphics[width=0.07\linewidth]{944.png} &&&
\includegraphics[width=0.07\linewidth]{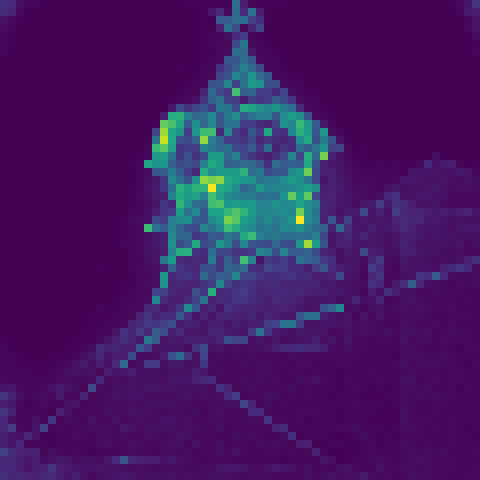} &
\includegraphics[width=0.07\linewidth]{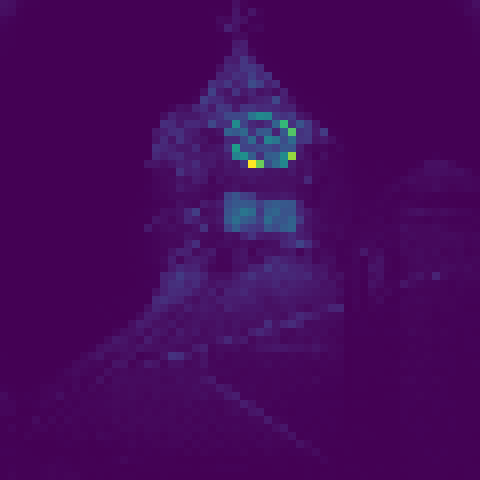} &
\includegraphics[width=0.07\linewidth]{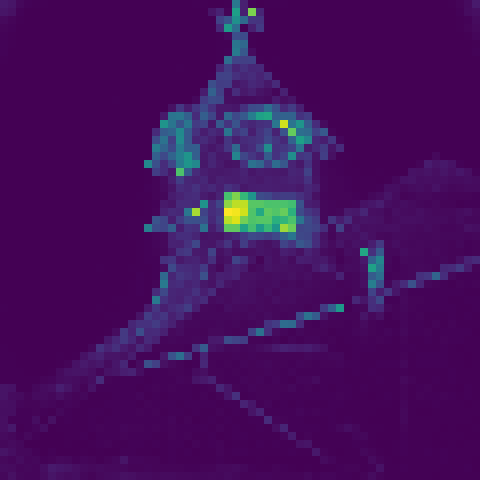} &&&
\includegraphics[width=0.07\linewidth]{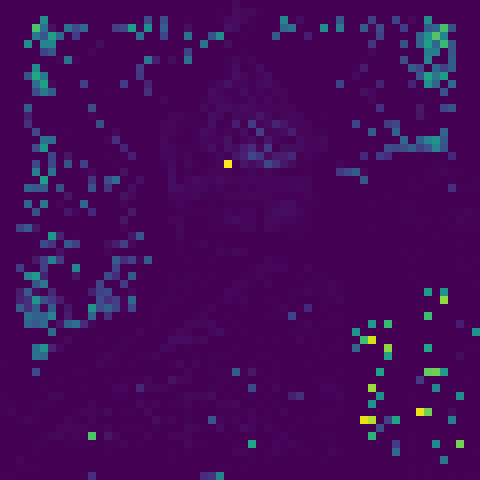} &
\includegraphics[width=0.07\linewidth]{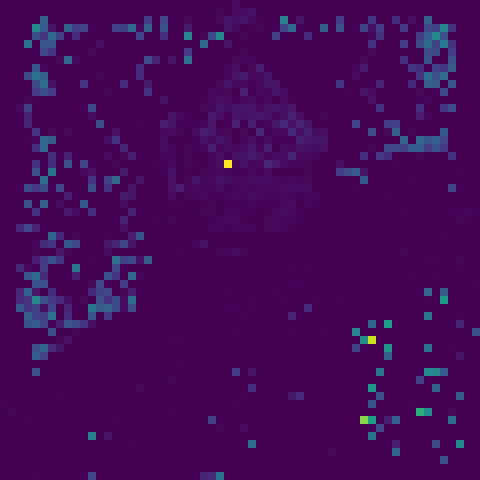} &
\includegraphics[width=0.07\linewidth]{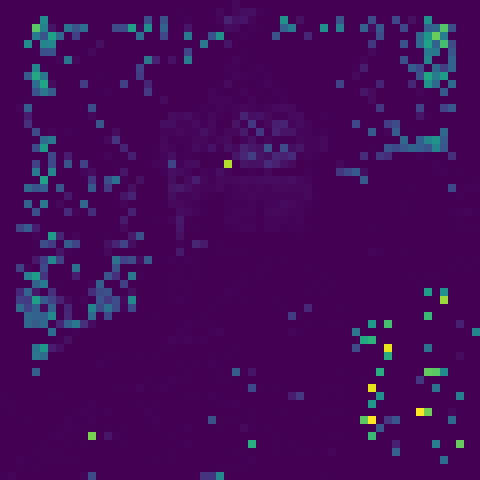}
\\
\includegraphics[width=0.07\linewidth]{4323.png} &&&
\includegraphics[width=0.07\linewidth]{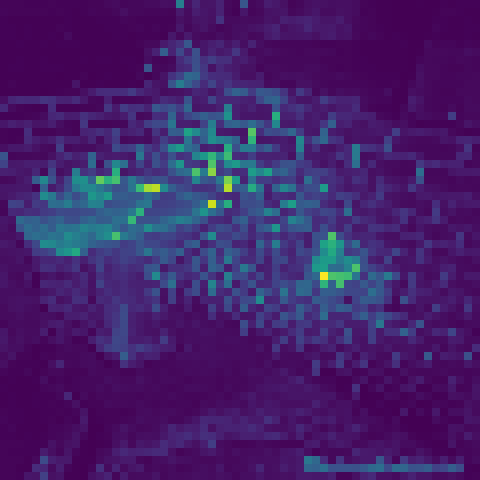} &
\includegraphics[width=0.07\linewidth]{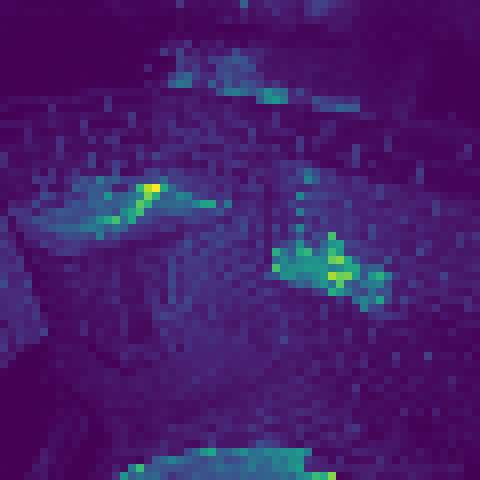} &
\includegraphics[width=0.07\linewidth]{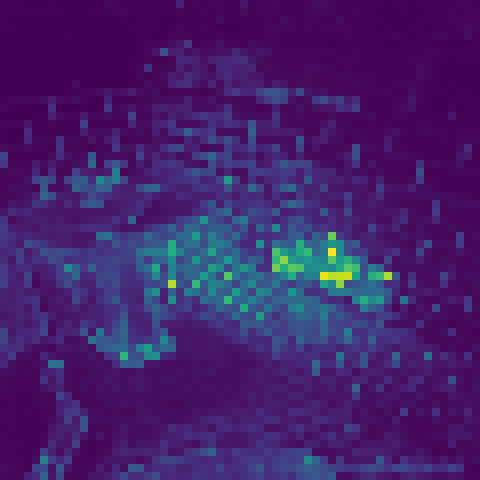} &&&
\includegraphics[width=0.07\linewidth]{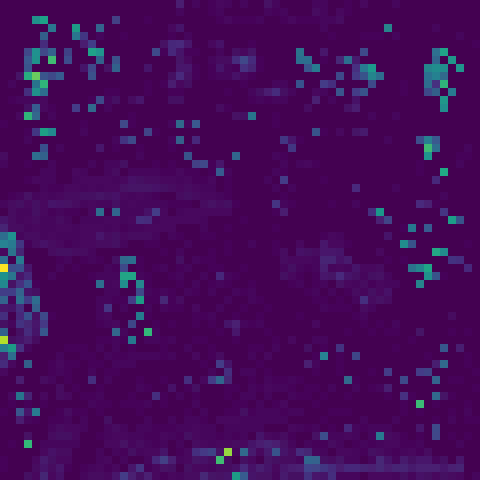} &
\includegraphics[width=0.07\linewidth]{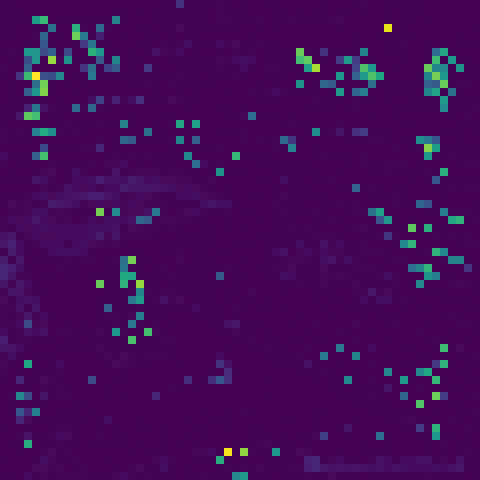} &
\includegraphics[width=0.07\linewidth]{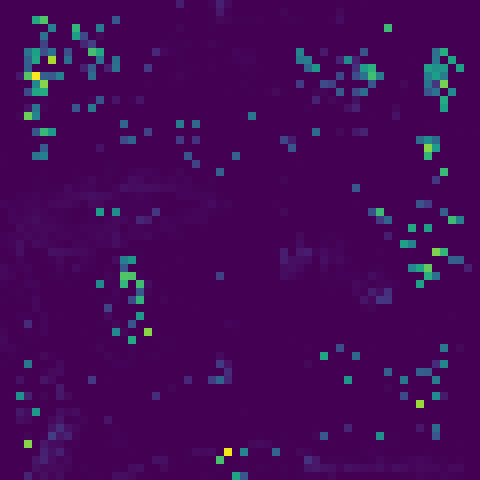}
&&&
\includegraphics[width=0.07\linewidth]{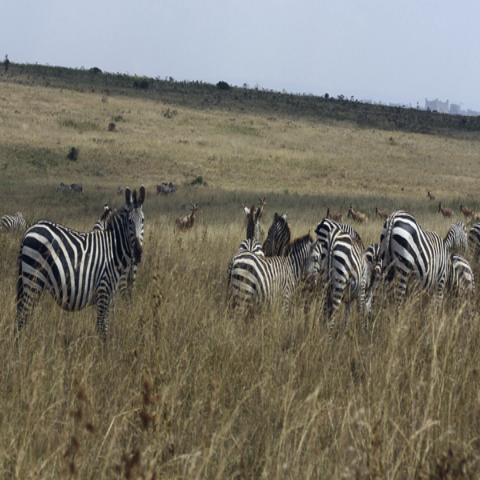} &&&
\includegraphics[width=0.07\linewidth]{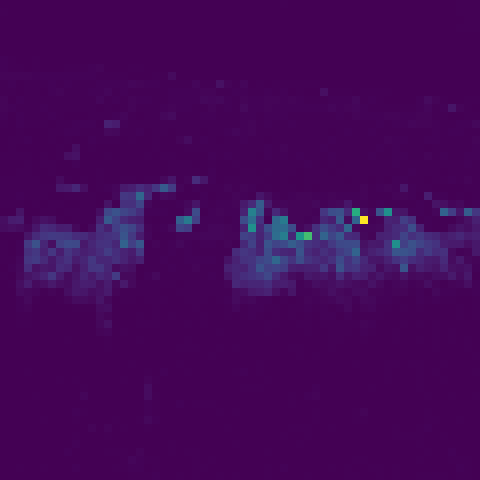} &
\includegraphics[width=0.07\linewidth]{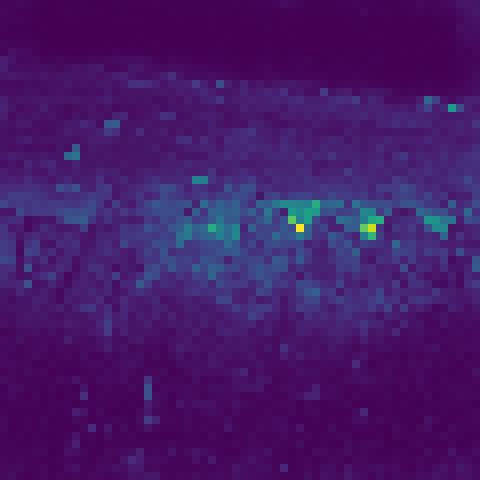} &
\includegraphics[width=0.07\linewidth]{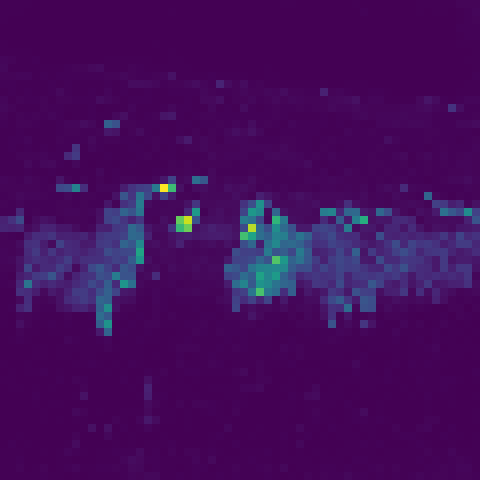} &&&
\includegraphics[width=0.07\linewidth]{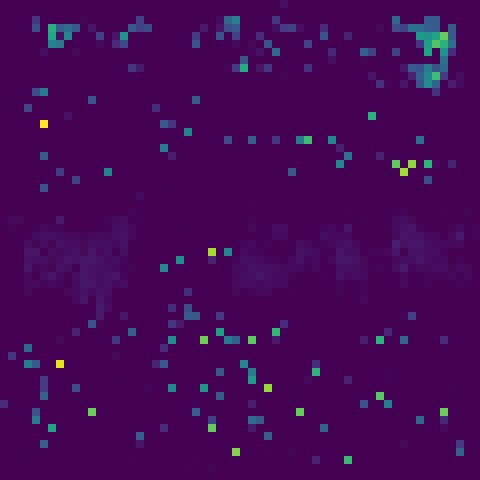} &
\includegraphics[width=0.07\linewidth]{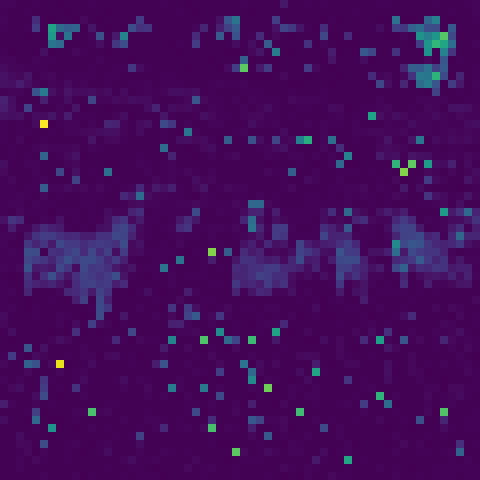} &
\includegraphics[width=0.07\linewidth]{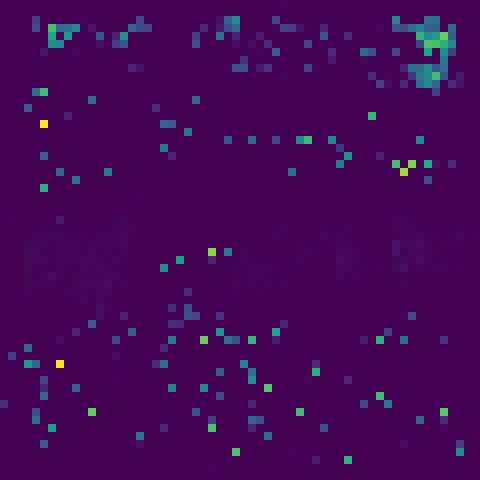}
\\
\includegraphics[width=0.07\linewidth]{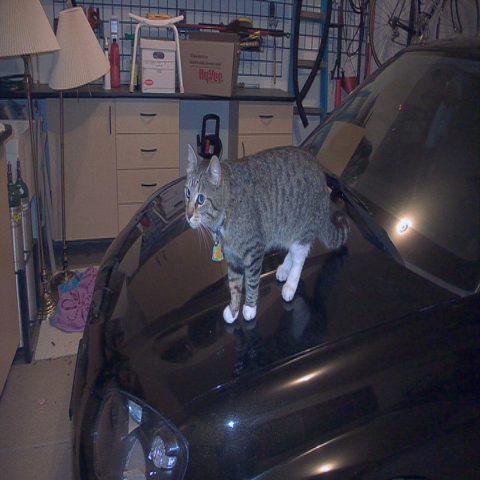} &&&
\includegraphics[width=0.07\linewidth]{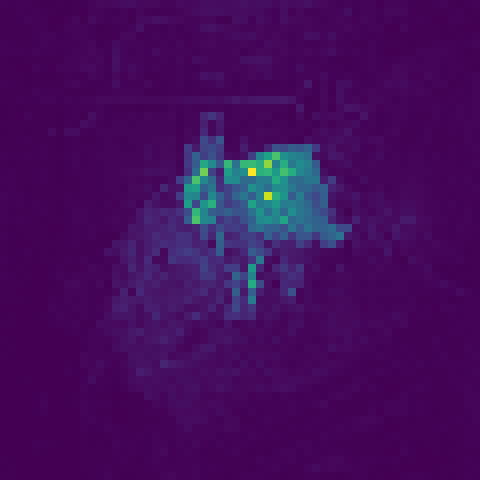} &
\includegraphics[width=0.07\linewidth]{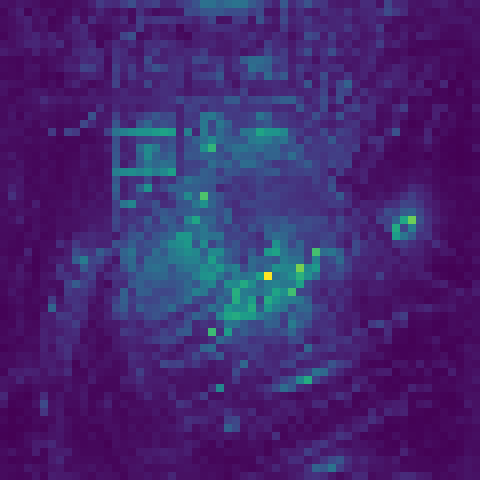} &
\includegraphics[width=0.07\linewidth]{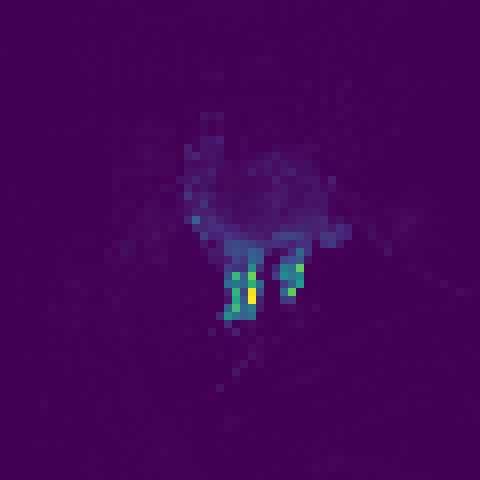} &&&
\includegraphics[width=0.07\linewidth]{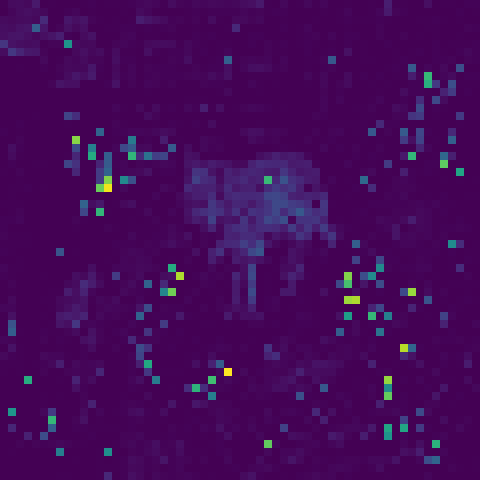} &
\includegraphics[width=0.07\linewidth]{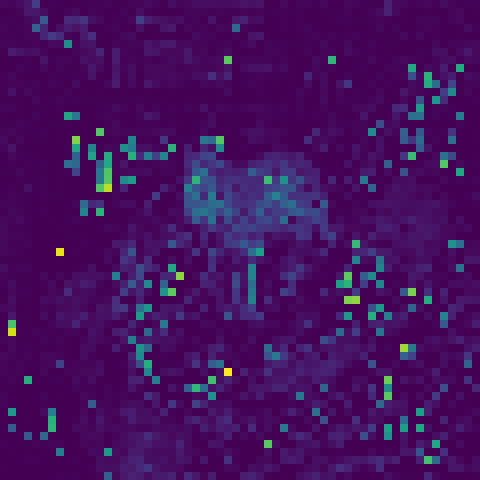} &
\includegraphics[width=0.07\linewidth]{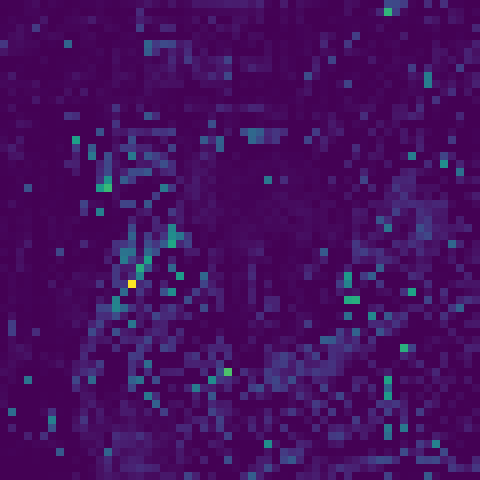}
&&&
\includegraphics[width=0.07\linewidth]{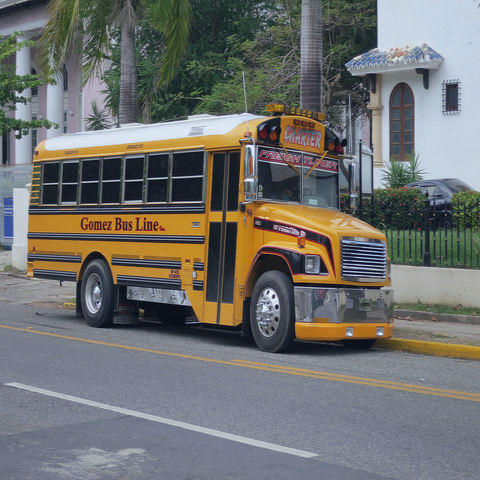} &&&
\includegraphics[width=0.07\linewidth]{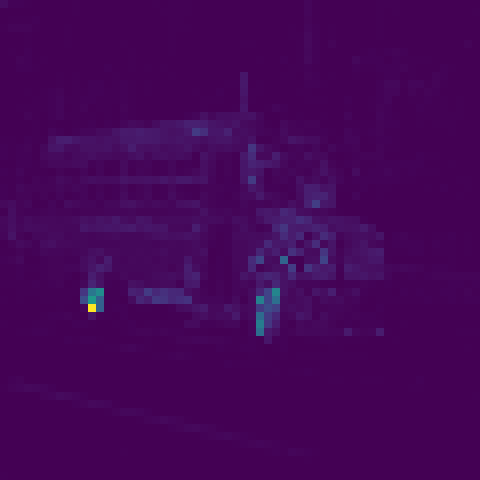} &
\includegraphics[width=0.07\linewidth]{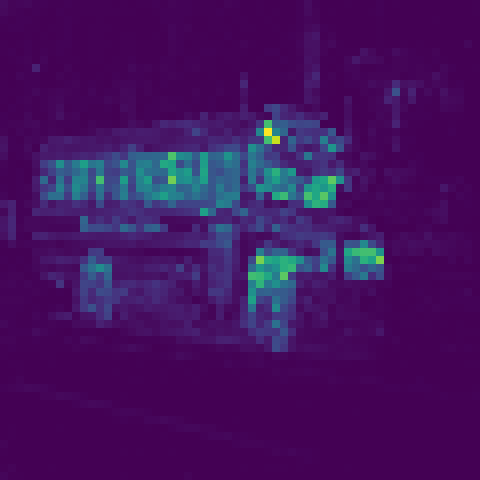} &
\includegraphics[width=0.07\linewidth]{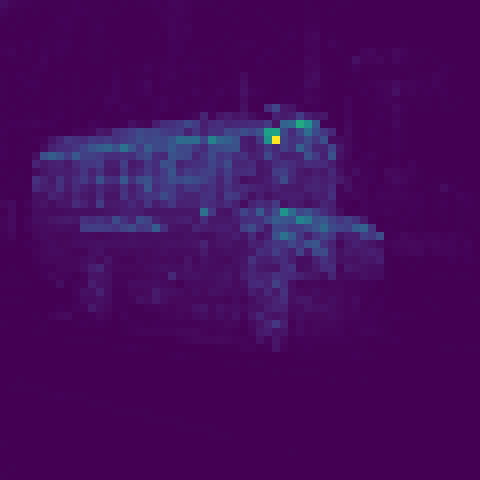} &&&
\includegraphics[width=0.07\linewidth]{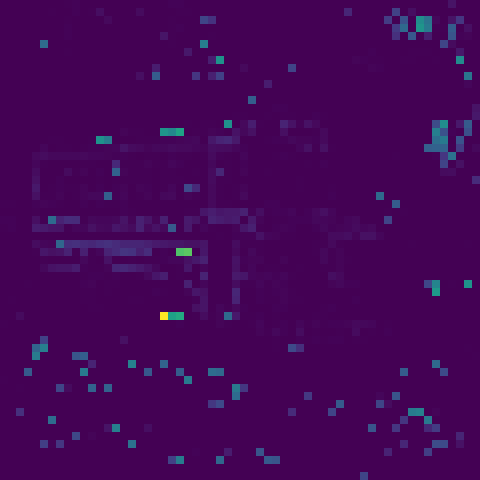} &
\includegraphics[width=0.07\linewidth]{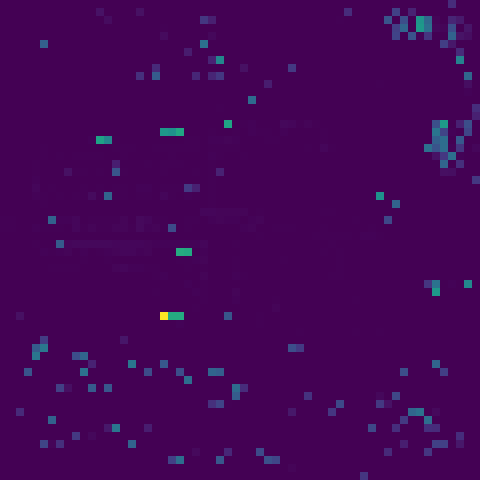} &
\includegraphics[width=0.07\linewidth]{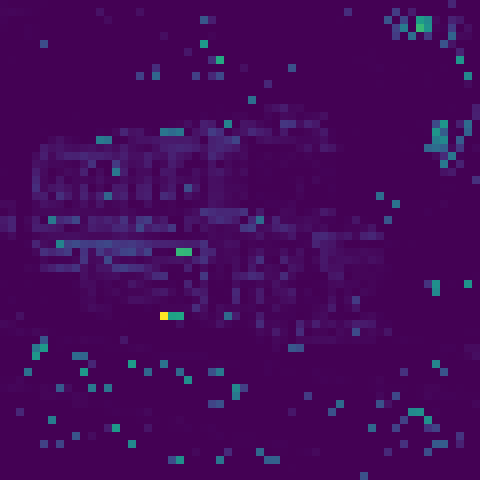}
\\
\includegraphics[width=0.07\linewidth]{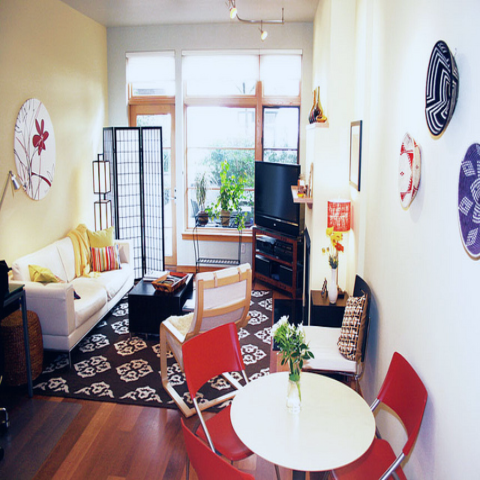} &&&
\includegraphics[width=0.07\linewidth]{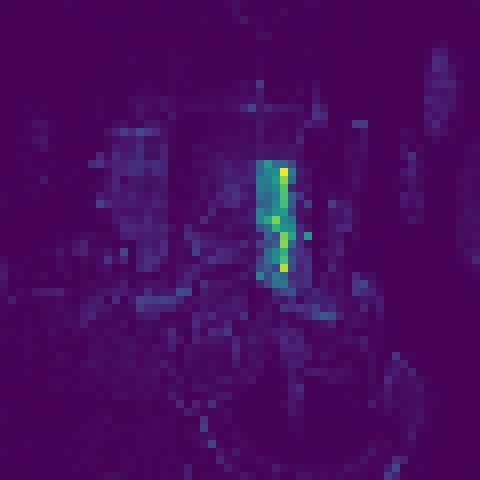} &
\includegraphics[width=0.07\linewidth]{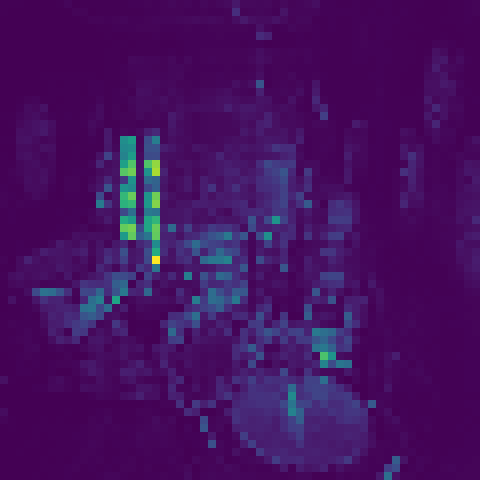} &
\includegraphics[width=0.07\linewidth]{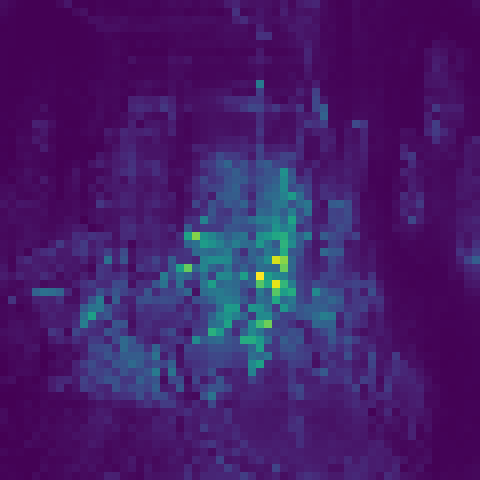} &&&
\includegraphics[width=0.07\linewidth]{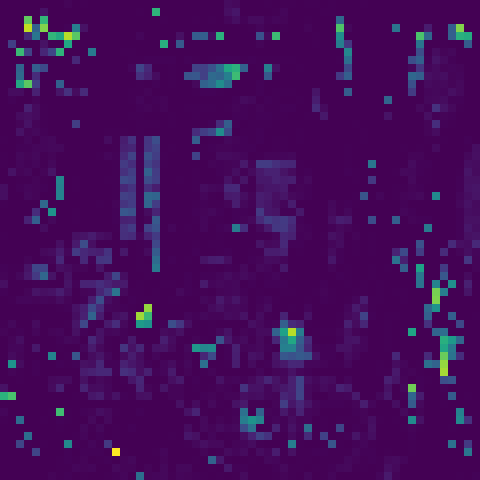} &
\includegraphics[width=0.07\linewidth]{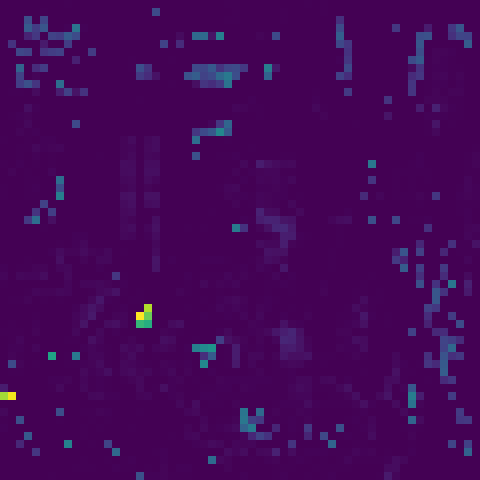} &
\includegraphics[width=0.07\linewidth]{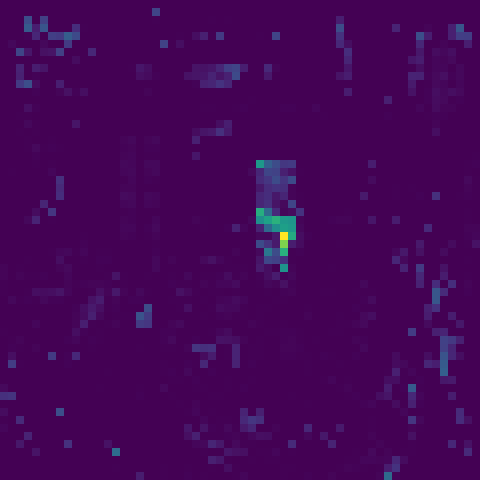}
&&&
\includegraphics[width=0.07\linewidth]{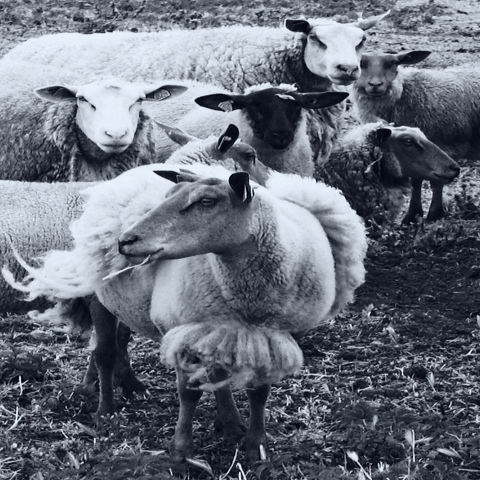} &&&
\includegraphics[width=0.07\linewidth]{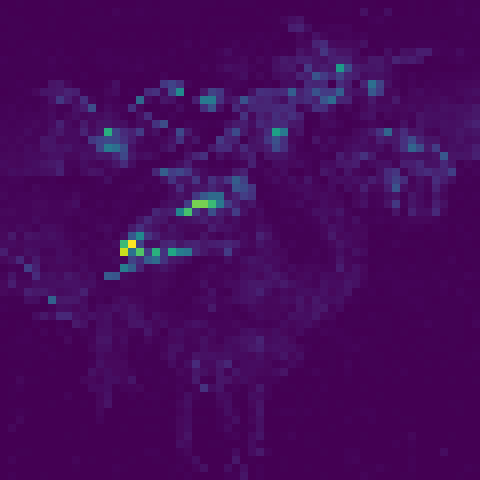} &
\includegraphics[width=0.07\linewidth]{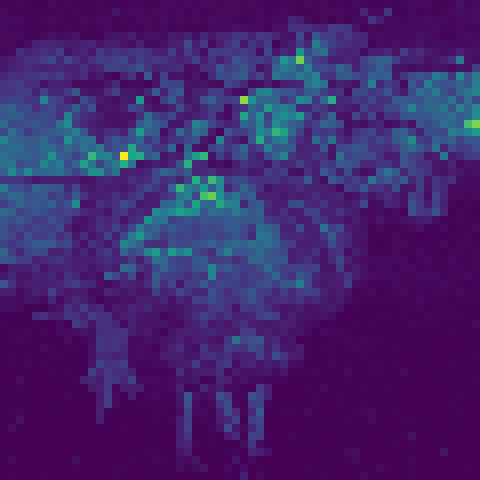} &
\includegraphics[width=0.07\linewidth]{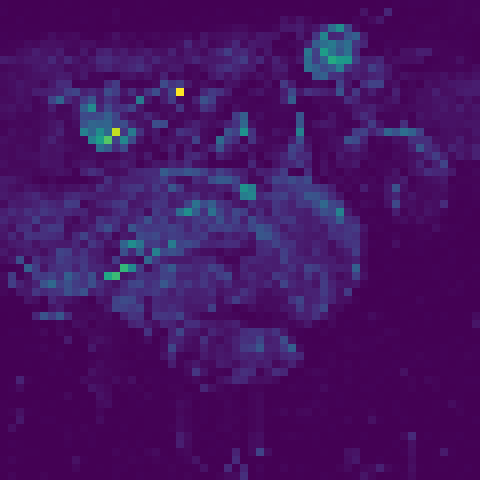} &&&
\includegraphics[width=0.07\linewidth]{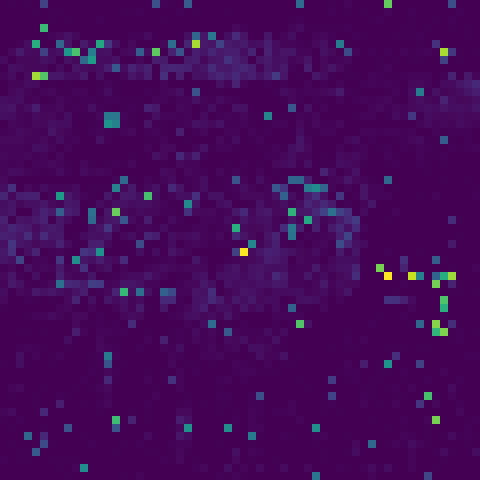} &
\includegraphics[width=0.07\linewidth]{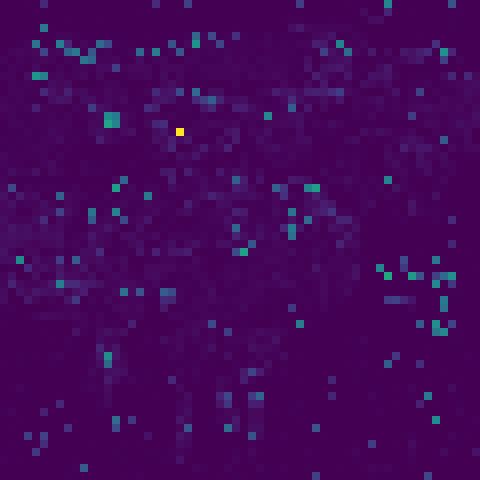} &
\includegraphics[width=0.07\linewidth]{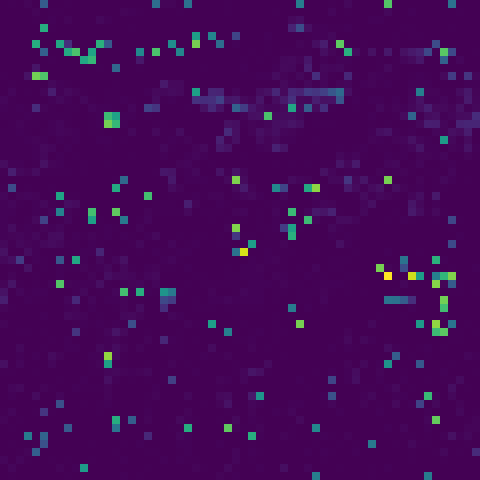}
\\
\includegraphics[width=0.07\linewidth]{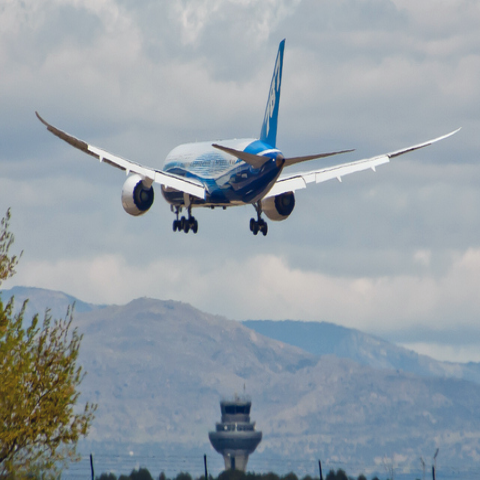} &&&
\includegraphics[width=0.07\linewidth]{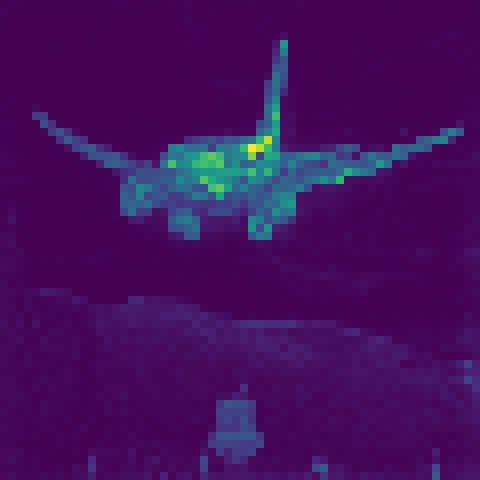} &
\includegraphics[width=0.07\linewidth]{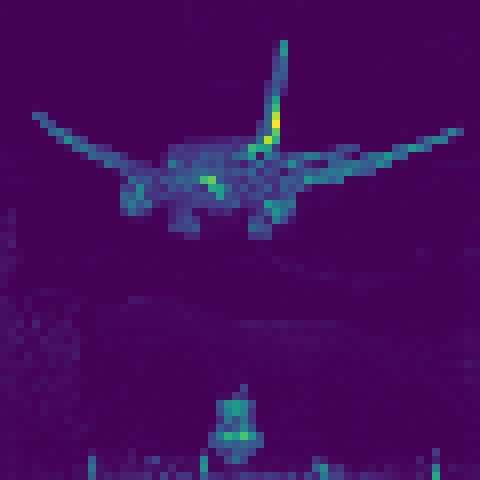} &
\includegraphics[width=0.07\linewidth]{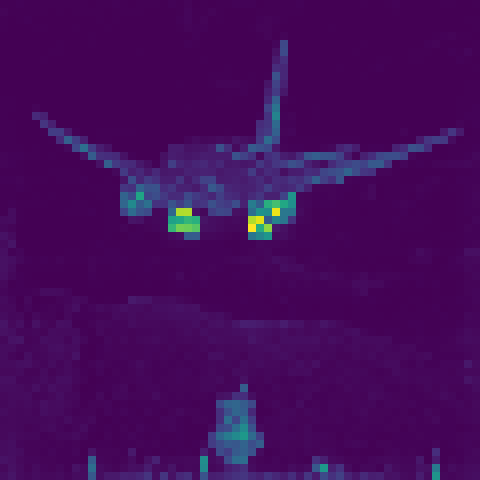} &&&
\includegraphics[width=0.07\linewidth]{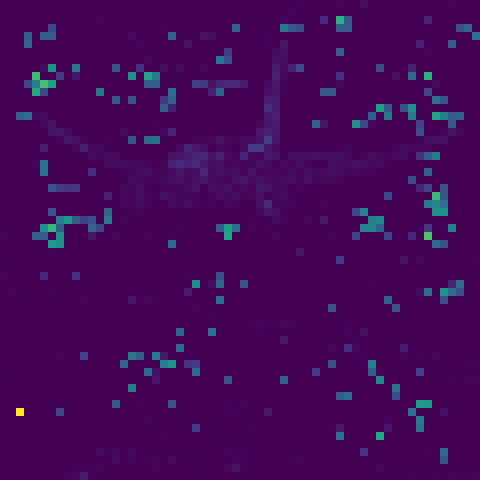} &
\includegraphics[width=0.07\linewidth]{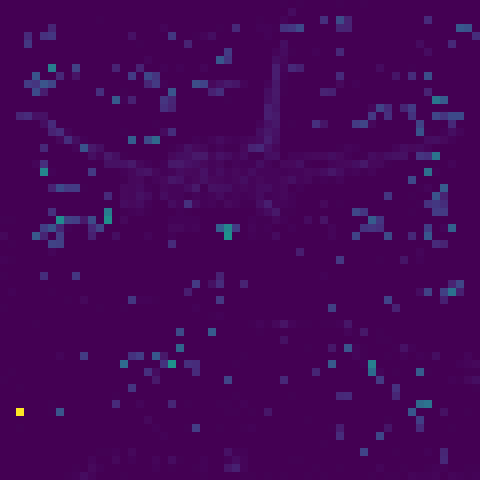} &
\includegraphics[width=0.07\linewidth]{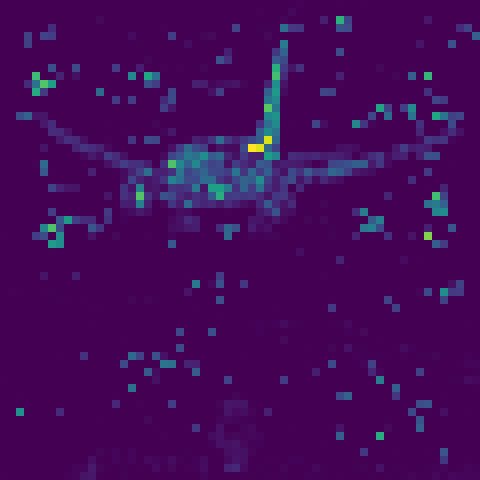}
&&&
\includegraphics[width=0.07\linewidth]{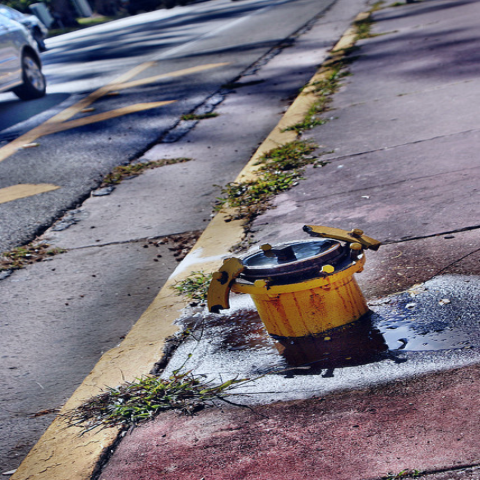} &&&
\includegraphics[width=0.07\linewidth]{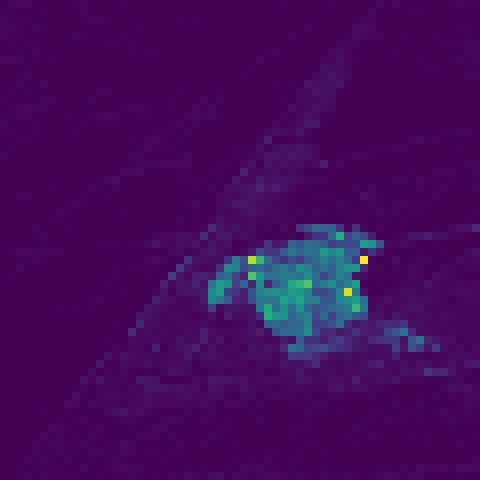} &
\includegraphics[width=0.07\linewidth]{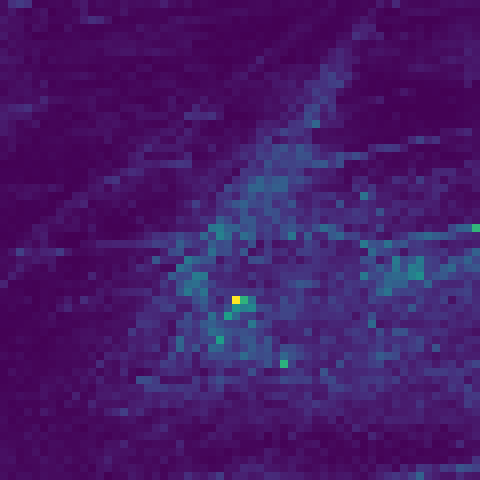} &
\includegraphics[width=0.07\linewidth]{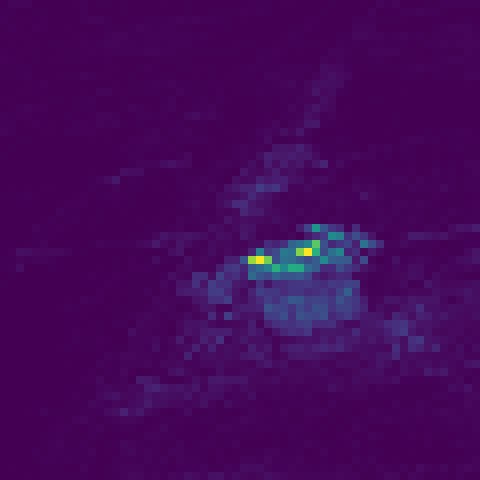} &&&
\includegraphics[width=0.07\linewidth]{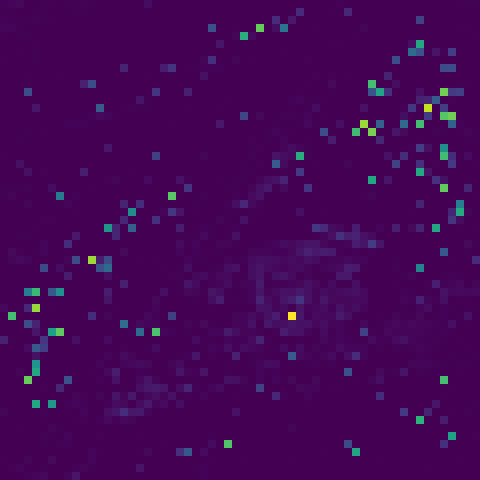} &
\includegraphics[width=0.07\linewidth]{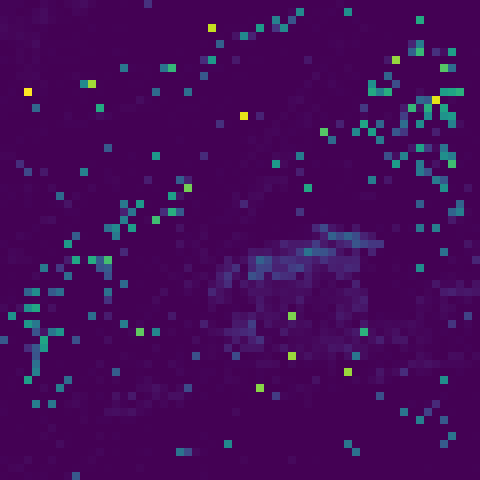} &
\includegraphics[width=0.07\linewidth]{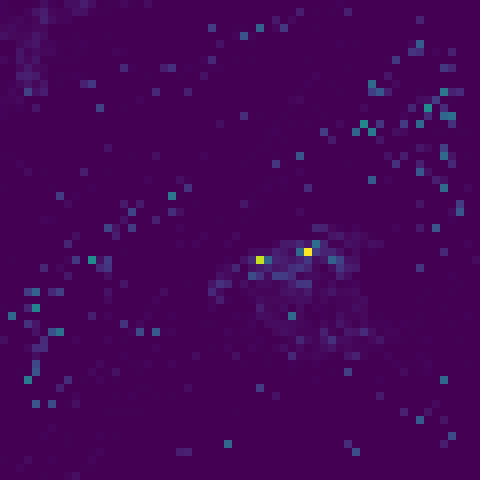}
\\
\includegraphics[width=0.07\linewidth]{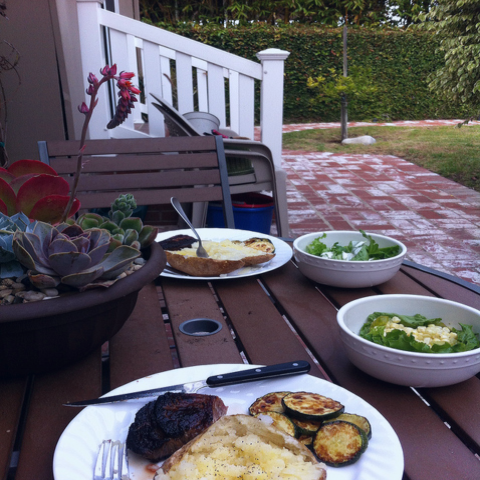} &&&
\includegraphics[width=0.07\linewidth]{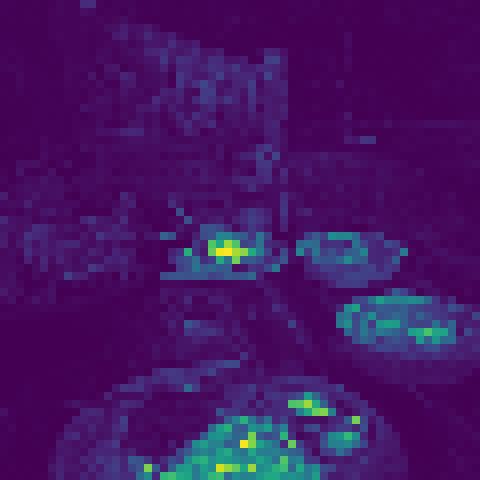} &
\includegraphics[width=0.07\linewidth]{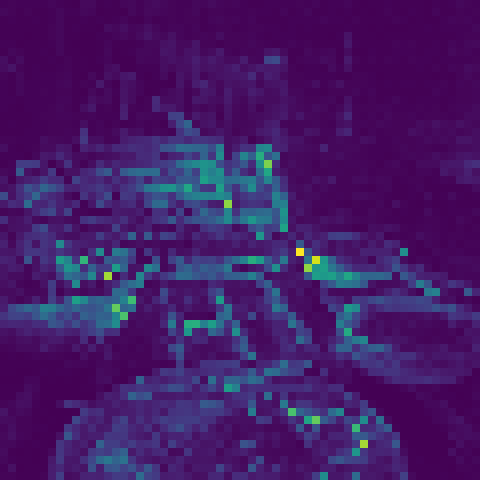} &
\includegraphics[width=0.07\linewidth]{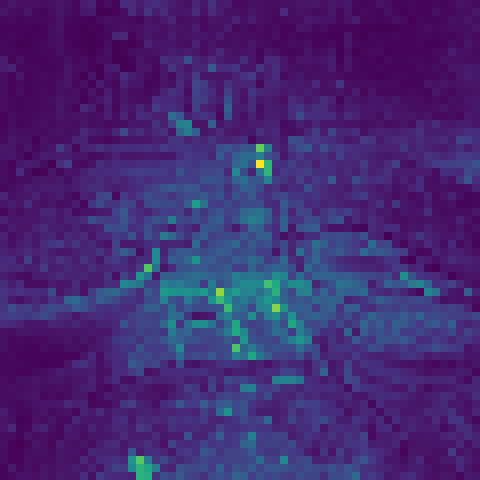} &&&
\includegraphics[width=0.07\linewidth]{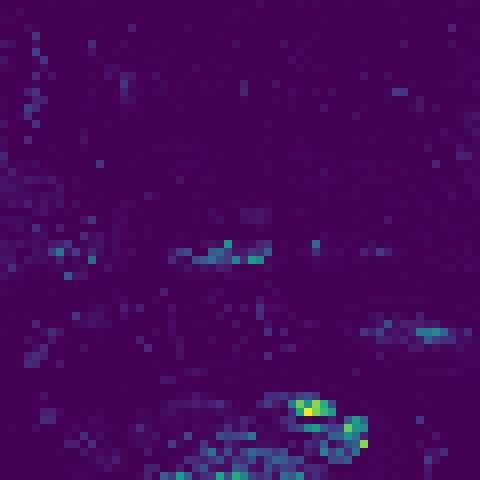} &
\includegraphics[width=0.07\linewidth]{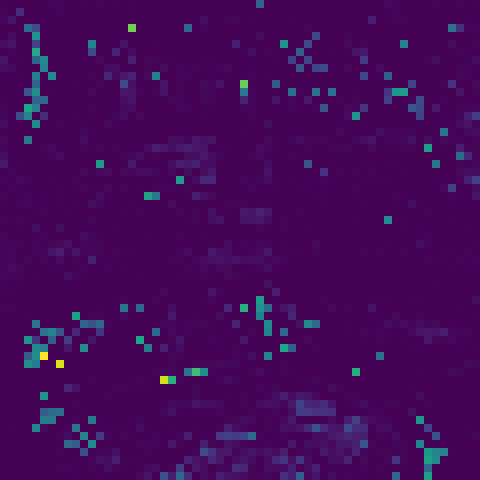} &
\includegraphics[width=0.07\linewidth]{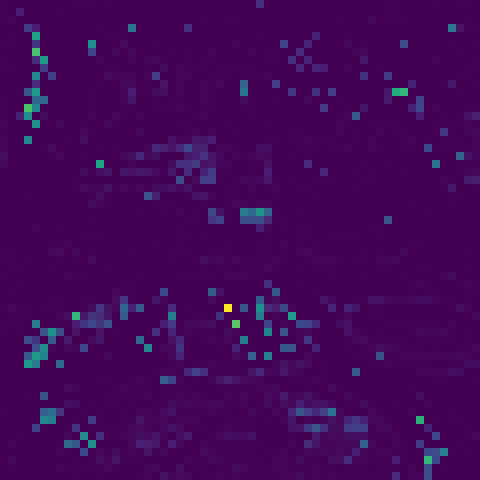}
&&&
\includegraphics[width=0.07\linewidth]{1508.png} &&&
\includegraphics[width=0.07\linewidth]{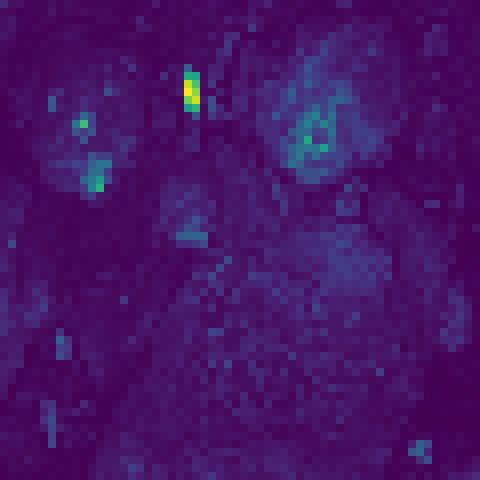} &
\includegraphics[width=0.07\linewidth]{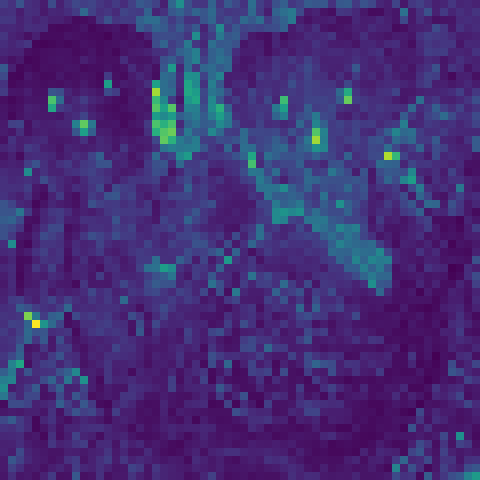} &
\includegraphics[width=0.07\linewidth]{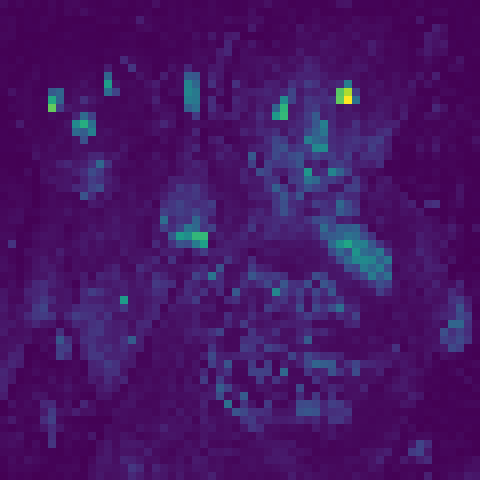} &&&
\includegraphics[width=0.07\linewidth]{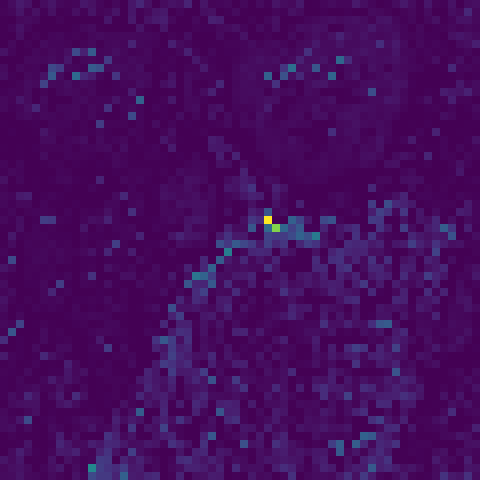} &
\includegraphics[width=0.07\linewidth]{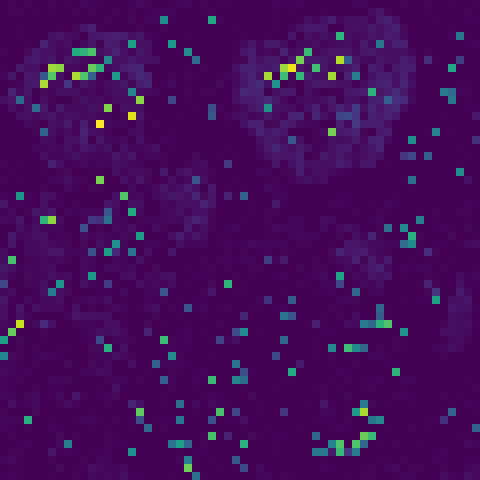} &
\includegraphics[width=0.07\linewidth]{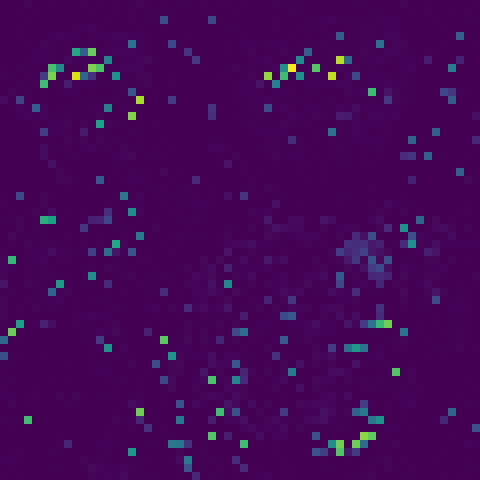}
\\
\includegraphics[width=0.07\linewidth]{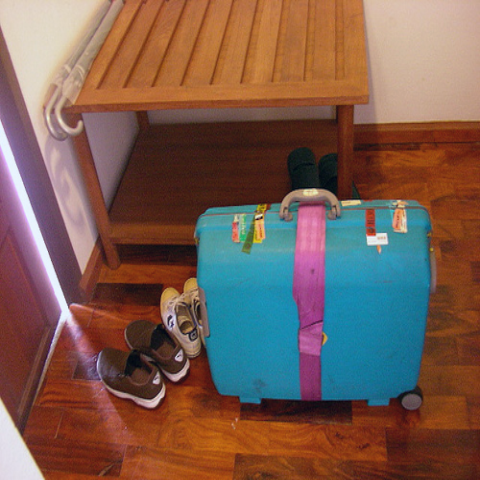} &&&
\includegraphics[width=0.07\linewidth]{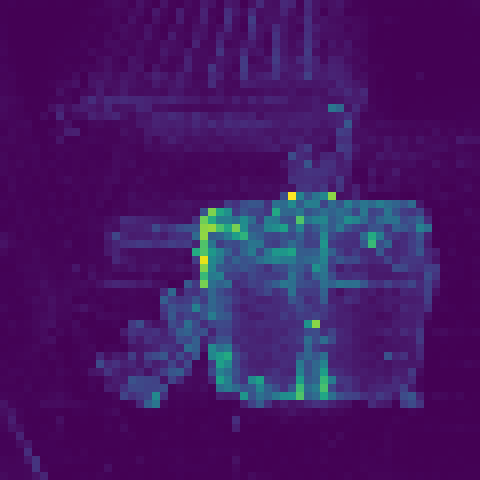} &
\includegraphics[width=0.07\linewidth]{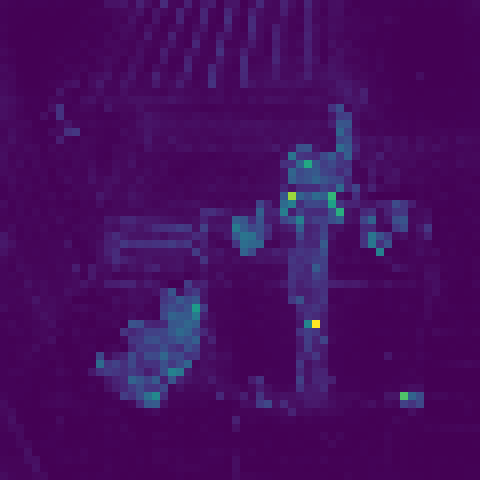} &
\includegraphics[width=0.07\linewidth]{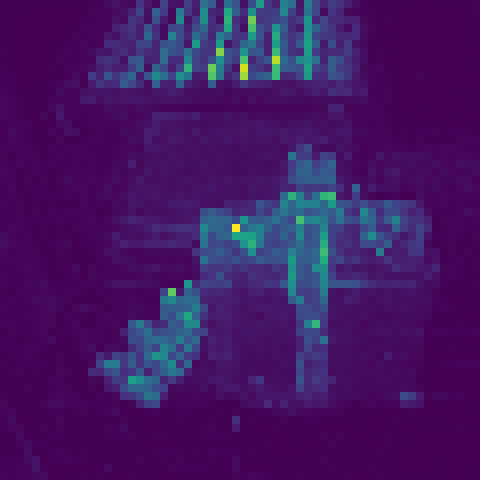} &&&
\includegraphics[width=0.07\linewidth]{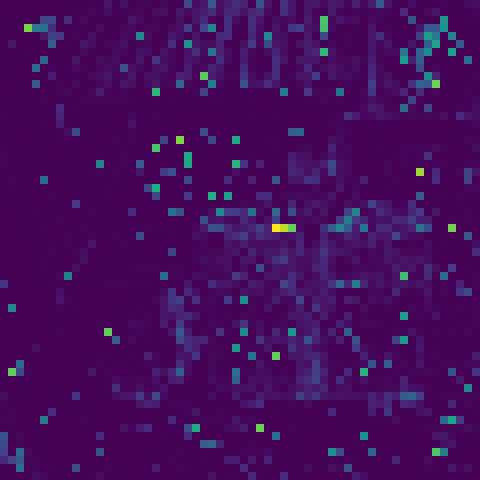} &
\includegraphics[width=0.07\linewidth]{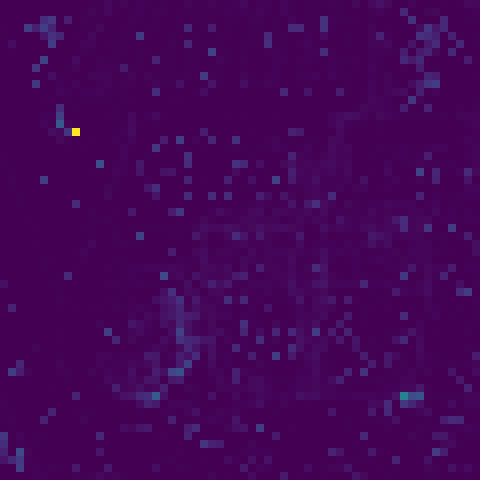} &
\includegraphics[width=0.07\linewidth]{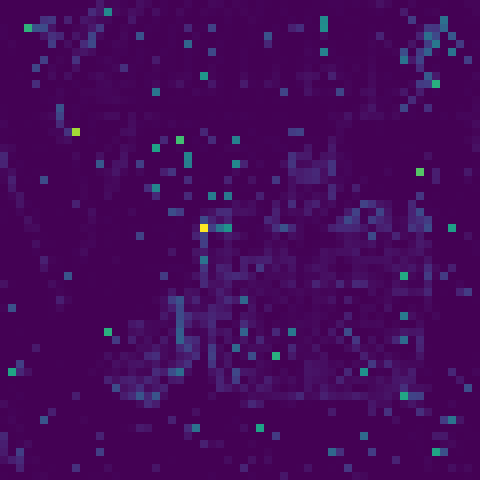}
&&&
\includegraphics[width=0.07\linewidth]{644.png} &&&
\includegraphics[width=0.07\linewidth]{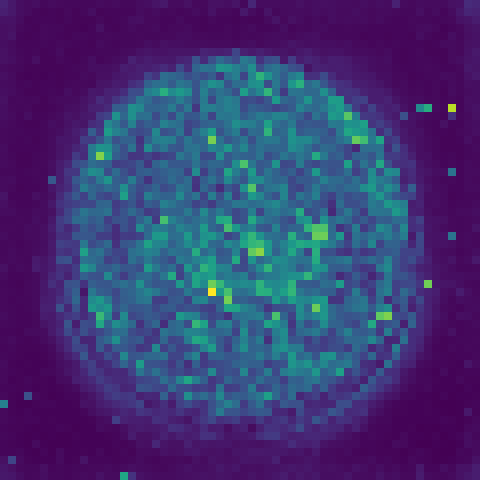} &
\includegraphics[width=0.07\linewidth]{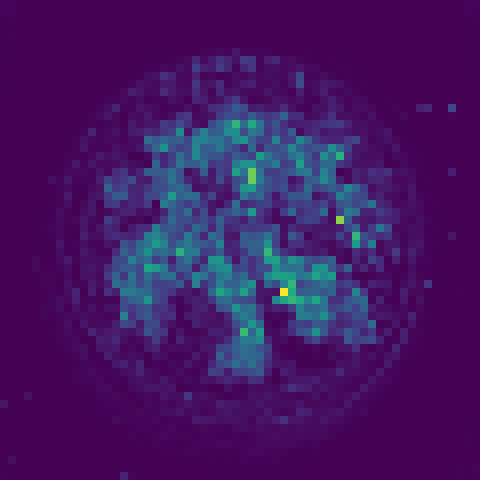} &
\includegraphics[width=0.07\linewidth]{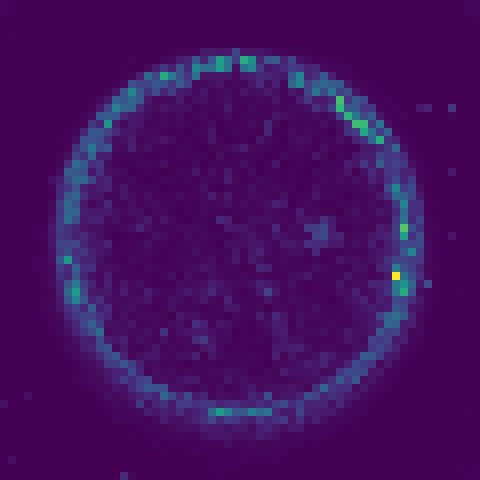} &&&
\includegraphics[width=0.07\linewidth]{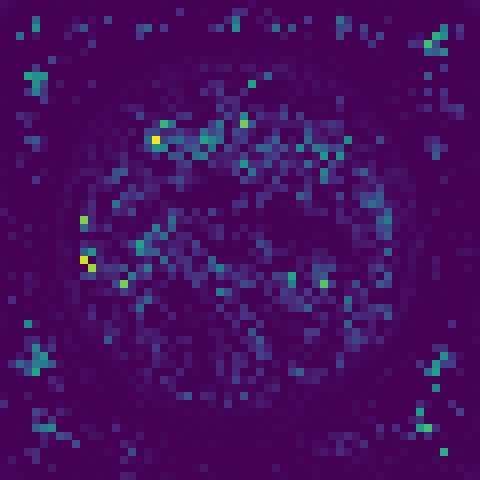} &
\includegraphics[width=0.07\linewidth]{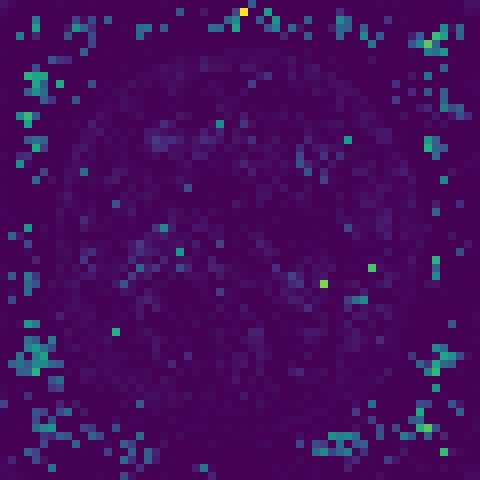} &
\includegraphics[width=0.07\linewidth]{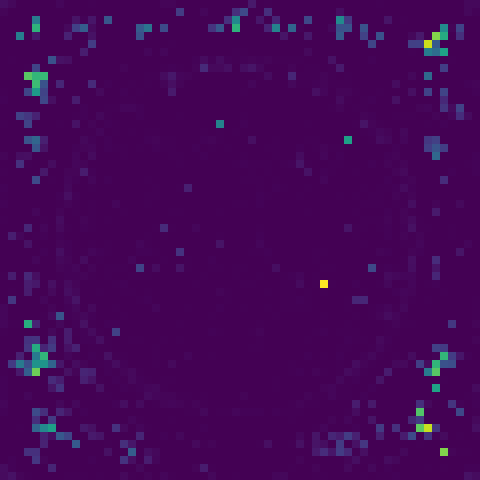}
\\
\includegraphics[width=0.07\linewidth]{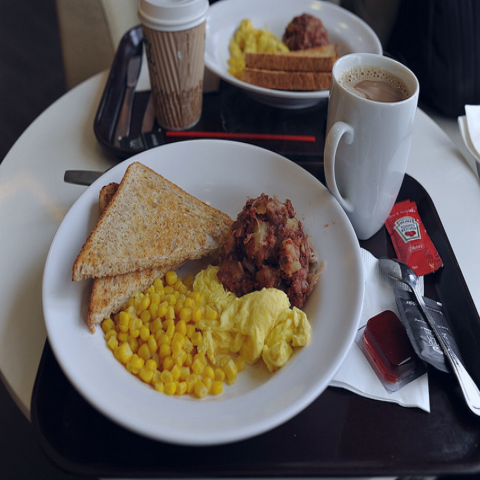} &&&
\includegraphics[width=0.07\linewidth]{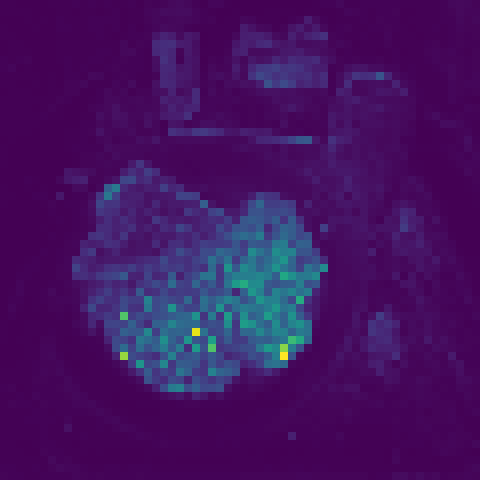} &
\includegraphics[width=0.07\linewidth]{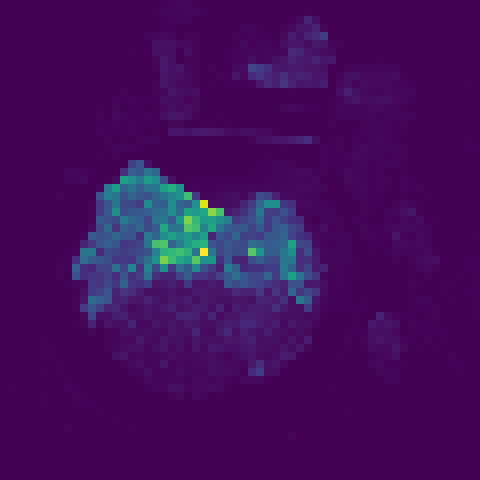} &
\includegraphics[width=0.07\linewidth]{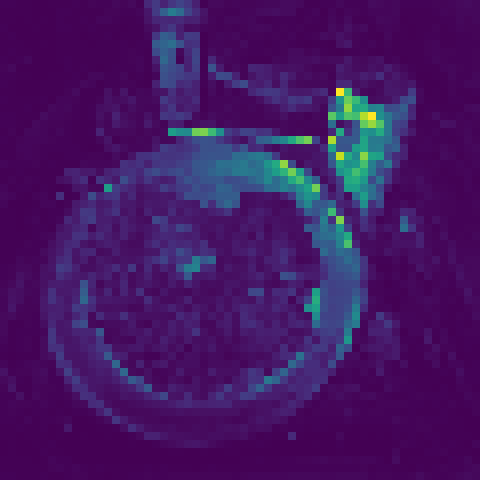} &&&
\includegraphics[width=0.07\linewidth]{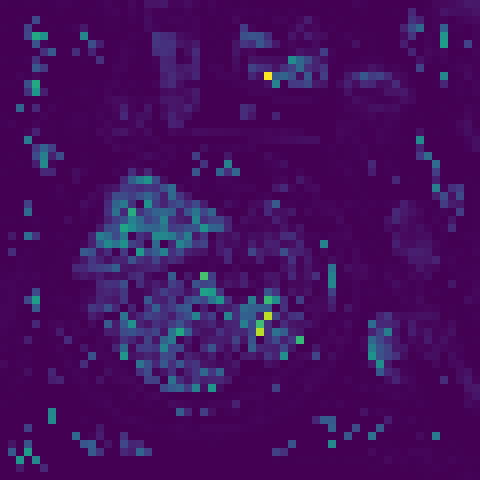} &
\includegraphics[width=0.07\linewidth]{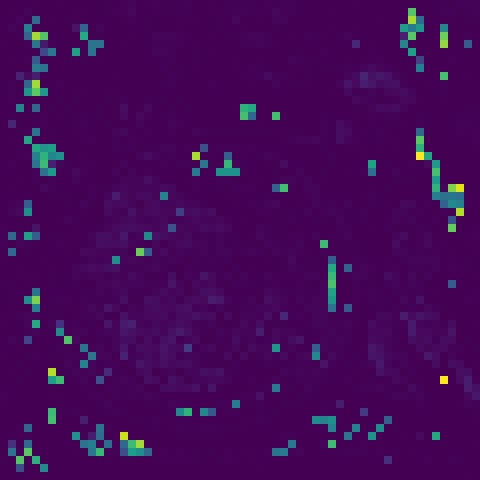} &
\includegraphics[width=0.07\linewidth]{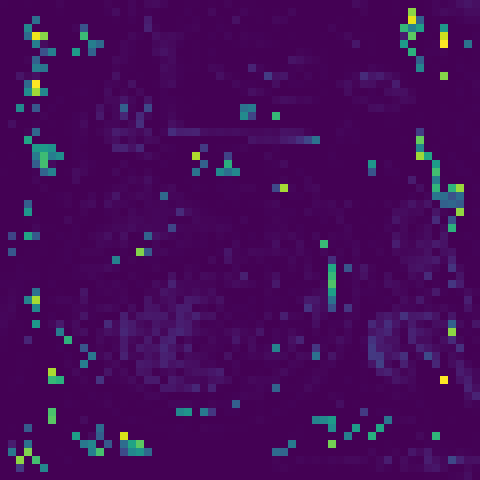}
&&&
\includegraphics[width=0.07\linewidth]{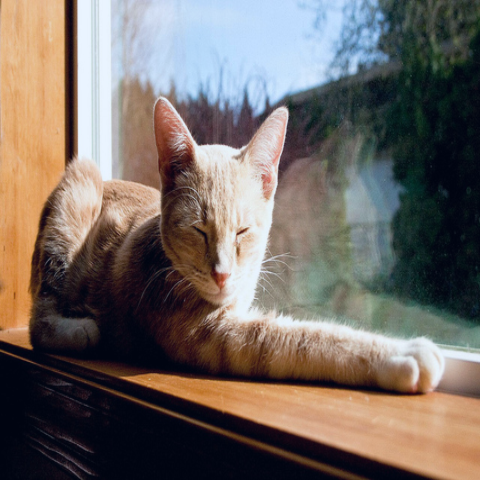} &&&
\includegraphics[width=0.07\linewidth]{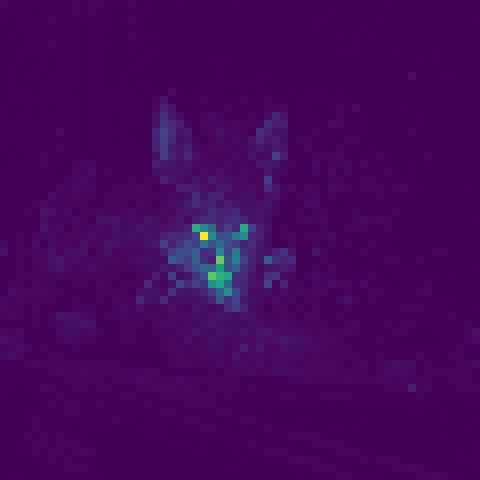} &
\includegraphics[width=0.07\linewidth]{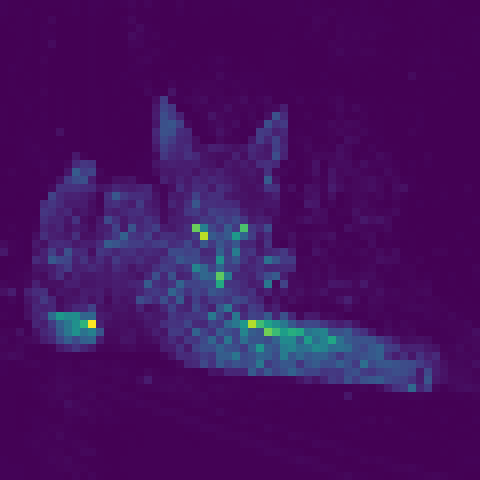} &
\includegraphics[width=0.07\linewidth]{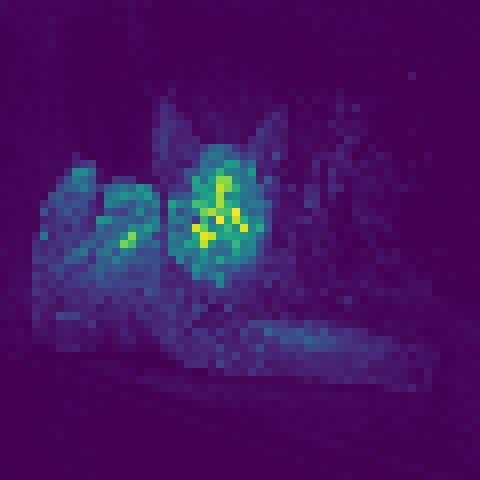} &&&
\includegraphics[width=0.07\linewidth]{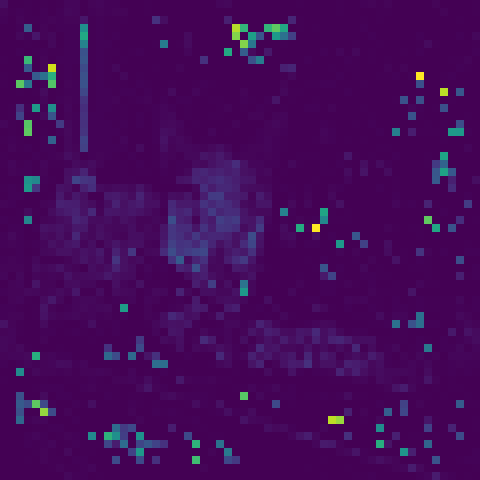} &
\includegraphics[width=0.07\linewidth]{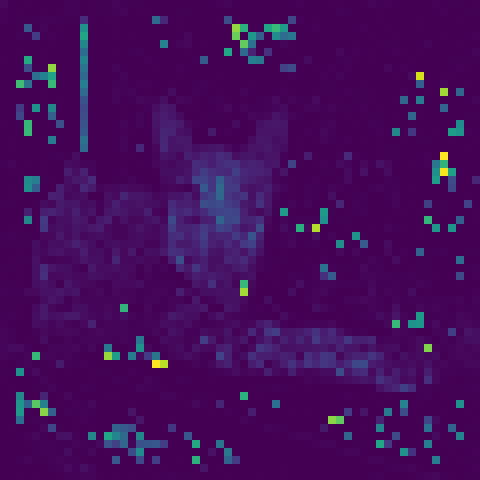} &
\includegraphics[width=0.07\linewidth]{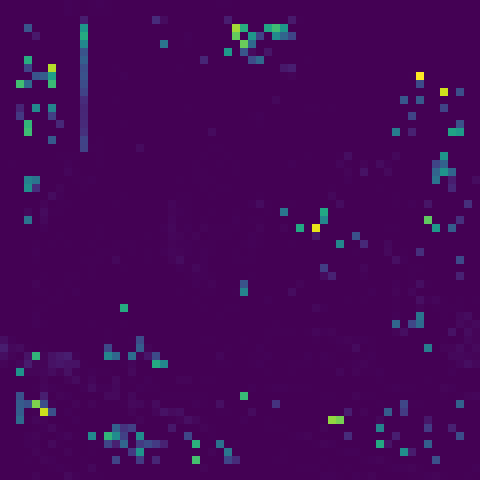}
\\
\includegraphics[width=0.07\linewidth]{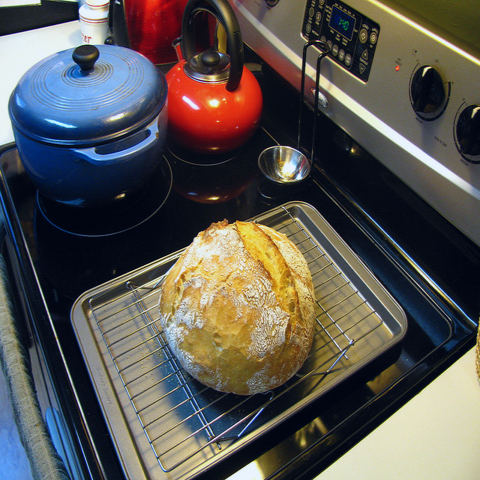} &&&
\includegraphics[width=0.07\linewidth]{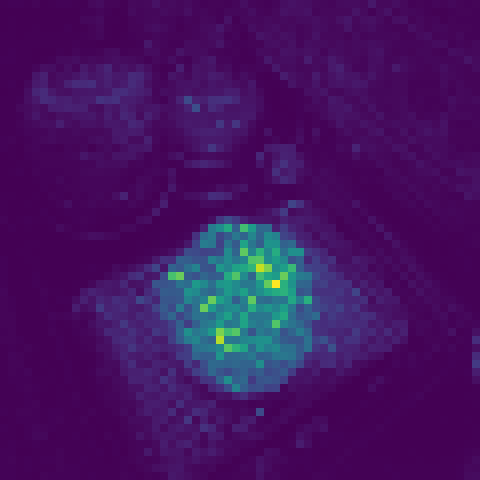} &
\includegraphics[width=0.07\linewidth]{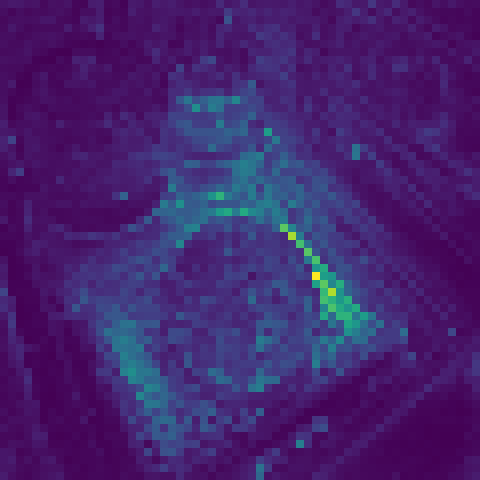} &
\includegraphics[width=0.07\linewidth]{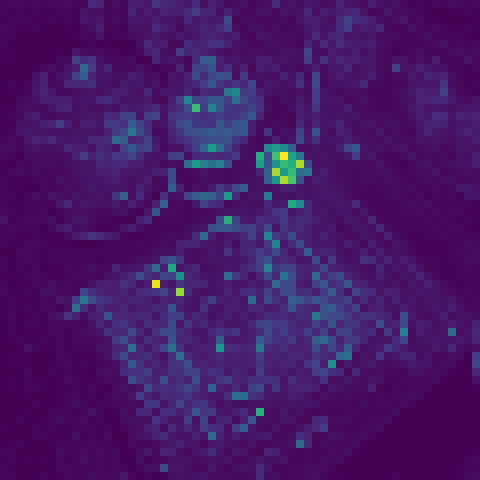} &&&
\includegraphics[width=0.07\linewidth]{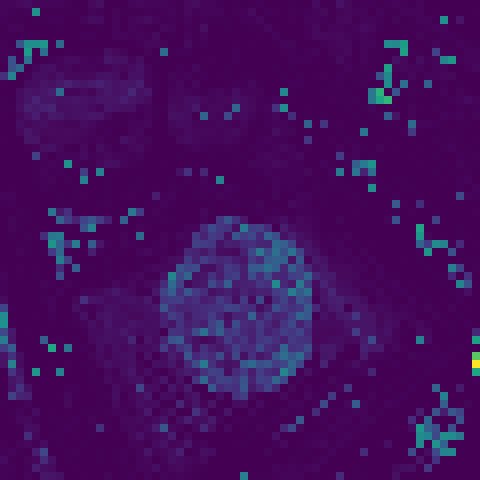} &
\includegraphics[width=0.07\linewidth]{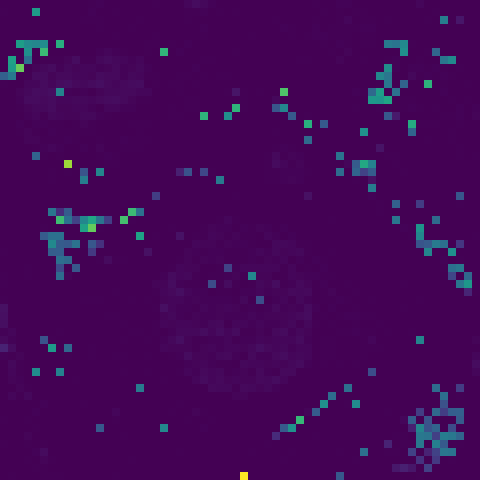} &
\includegraphics[width=0.07\linewidth]{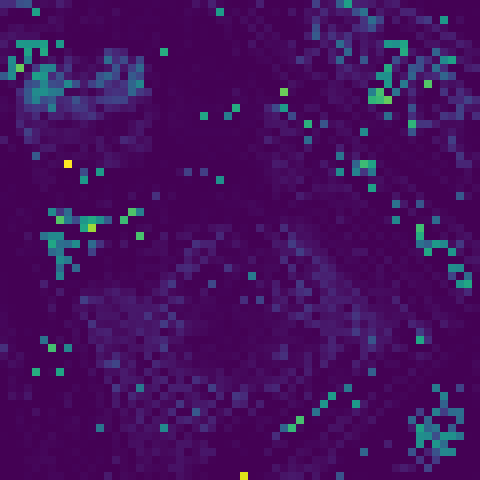}
&&&
\includegraphics[width=0.07\linewidth]{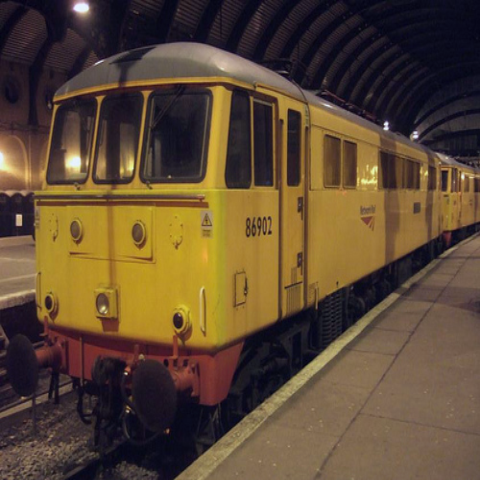} &&&
\includegraphics[width=0.07\linewidth]{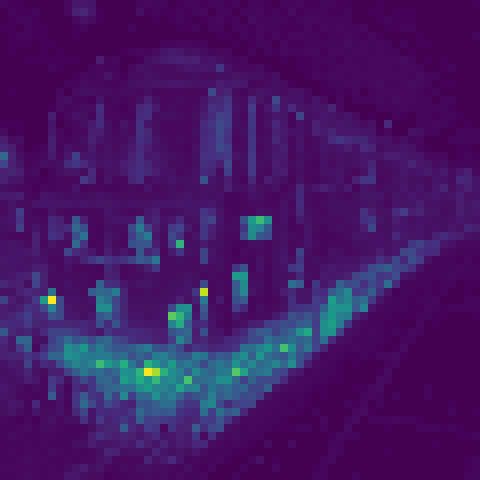} &
\includegraphics[width=0.07\linewidth]{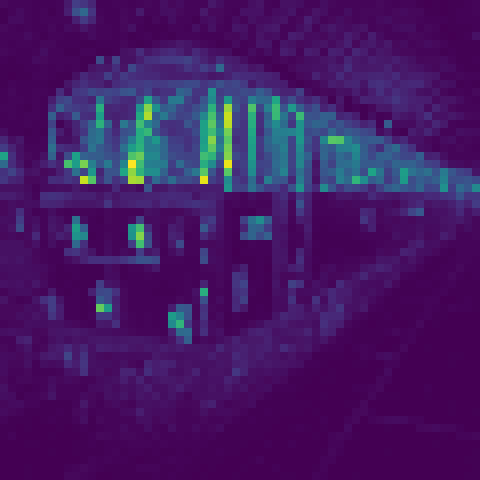} &
\includegraphics[width=0.07\linewidth]{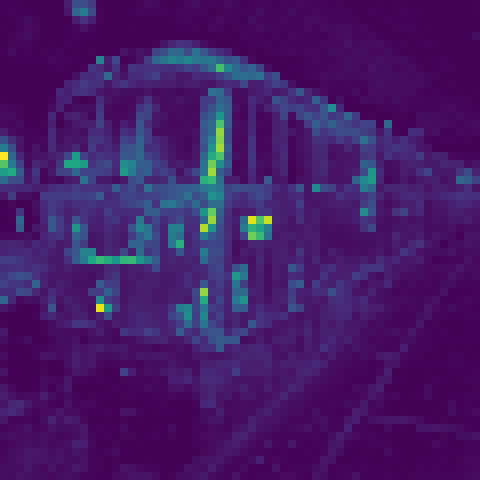} &&&
\includegraphics[width=0.07\linewidth]{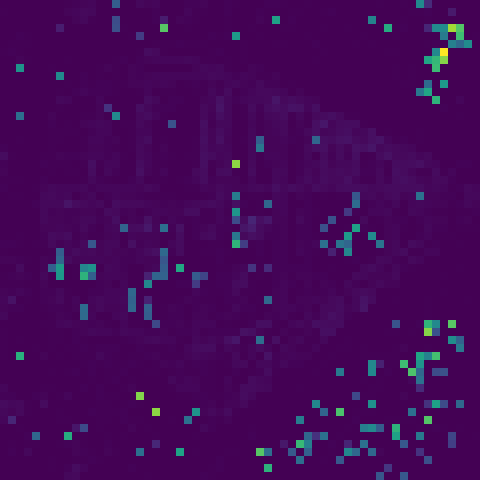} &
\includegraphics[width=0.07\linewidth]{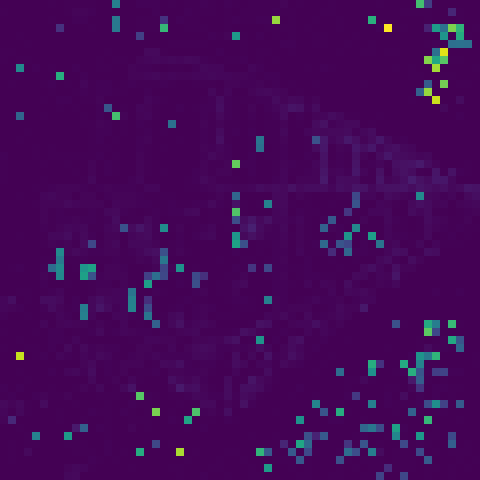} &
\includegraphics[width=0.07\linewidth]{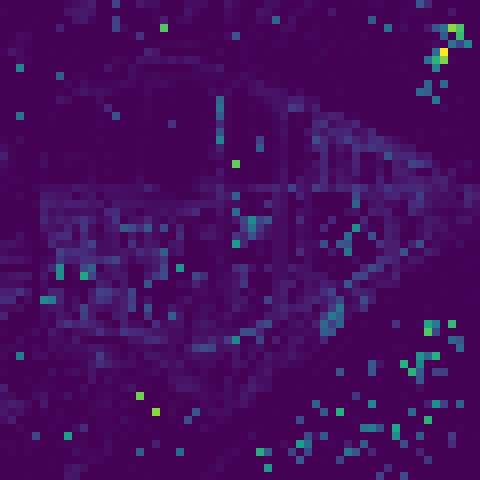}
\\
\includegraphics[width=0.07\linewidth]{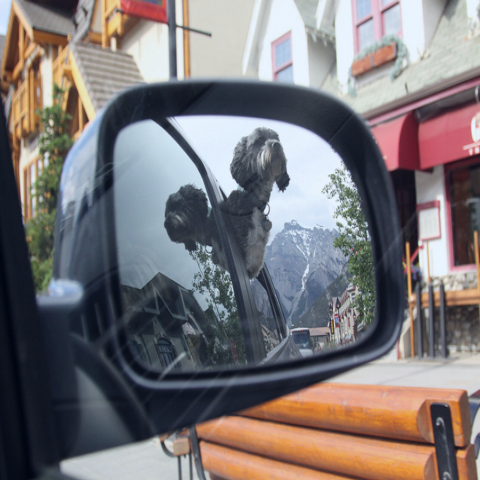} &&&
\includegraphics[width=0.07\linewidth]{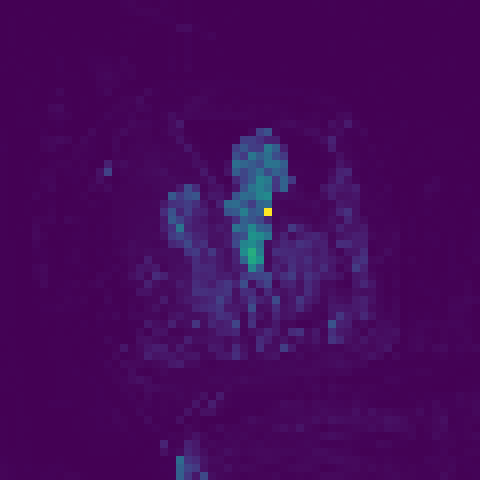} &
\includegraphics[width=0.07\linewidth]{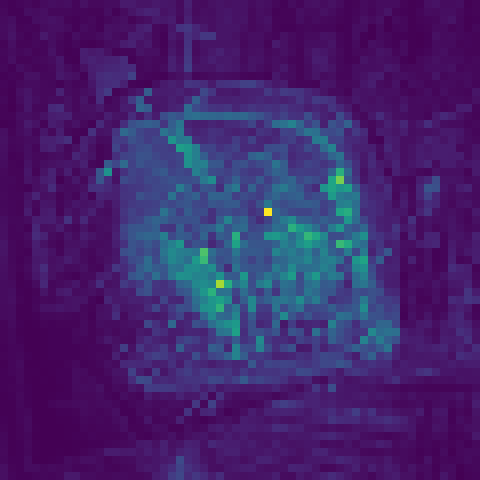} &
\includegraphics[width=0.07\linewidth]{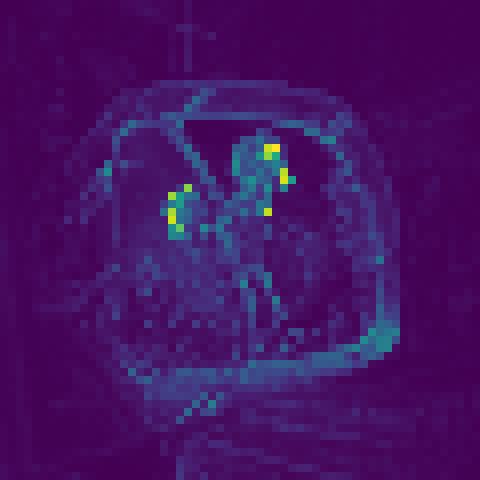} &&&
\includegraphics[width=0.07\linewidth]{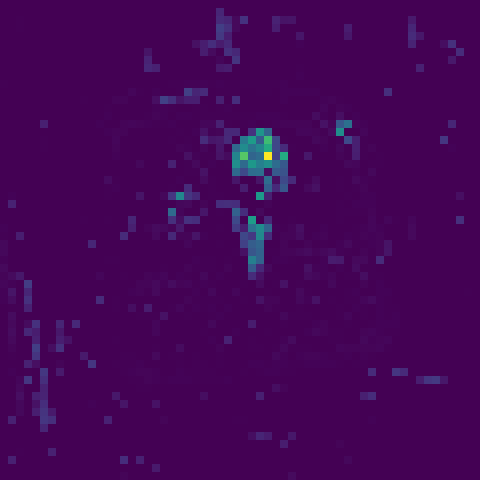} &
\includegraphics[width=0.07\linewidth]{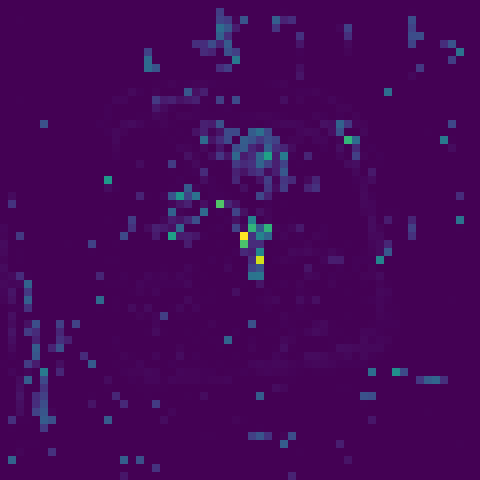} &
\includegraphics[width=0.07\linewidth]{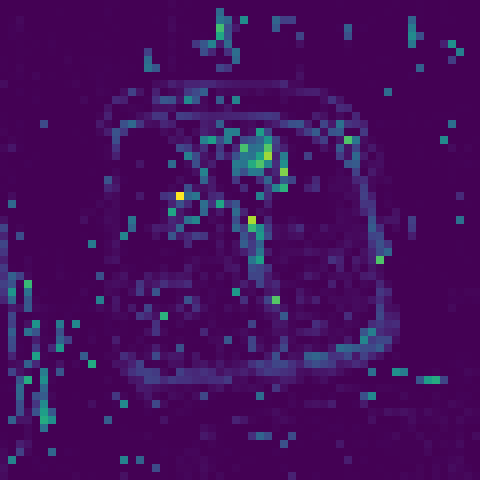}
&&&
\includegraphics[width=0.07\linewidth]{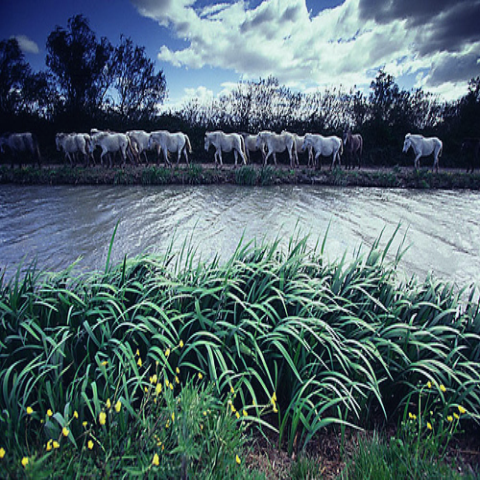} &&&
\includegraphics[width=0.07\linewidth]{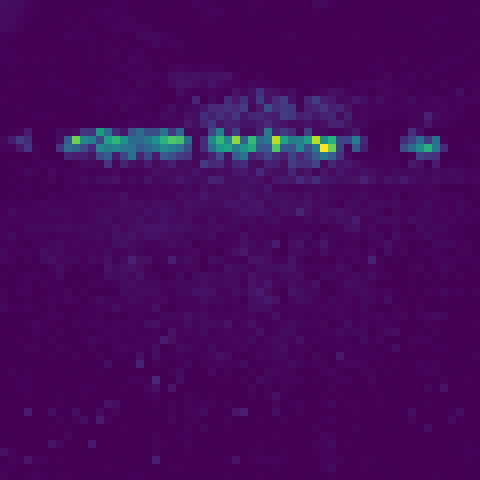} &
\includegraphics[width=0.07\linewidth]{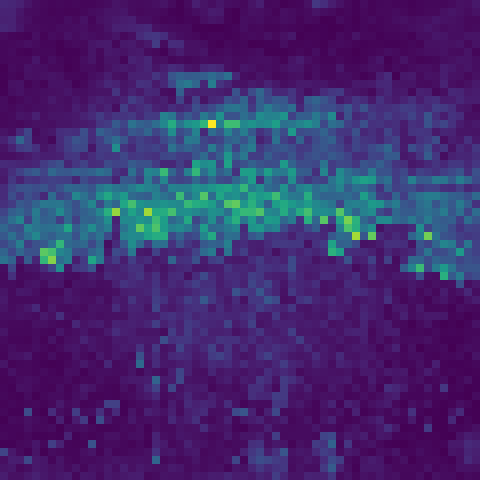} &
\includegraphics[width=0.07\linewidth]{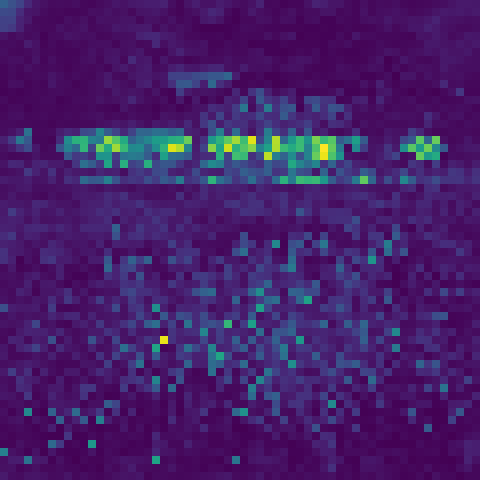} &&&
\includegraphics[width=0.07\linewidth]{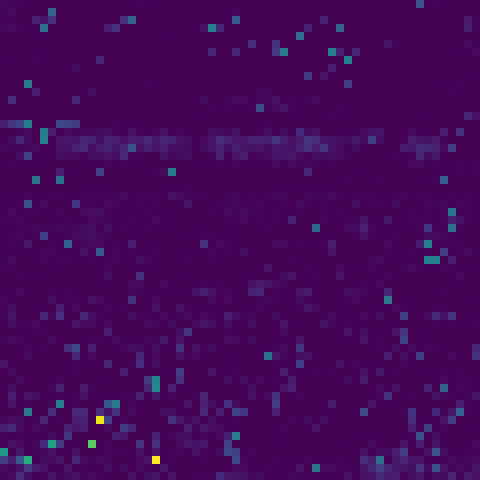} &
\includegraphics[width=0.07\linewidth]{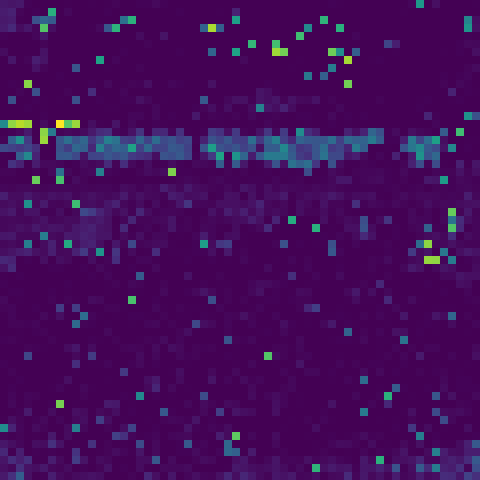} &
\includegraphics[width=0.07\linewidth]{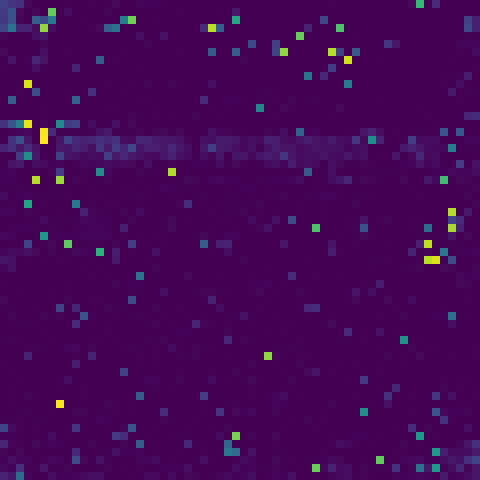}
\\
\includegraphics[width=0.07\linewidth]{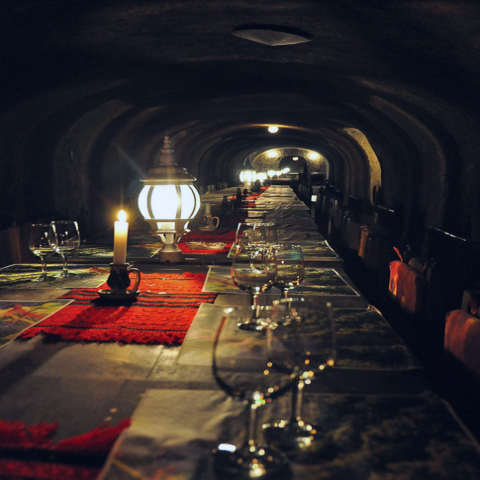} &&&
\includegraphics[width=0.07\linewidth]{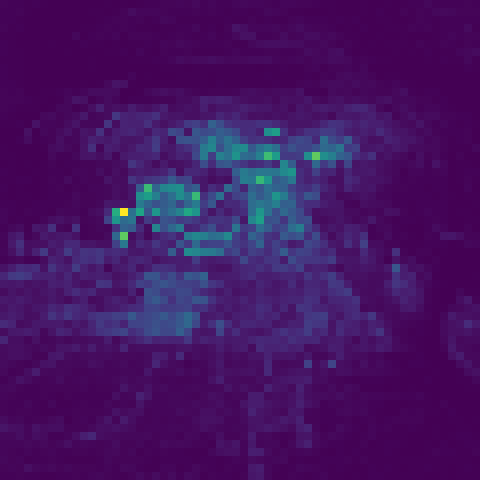} &
\includegraphics[width=0.07\linewidth]{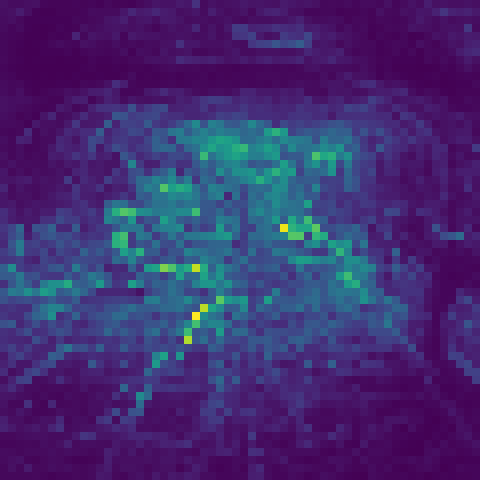} &
\includegraphics[width=0.07\linewidth]{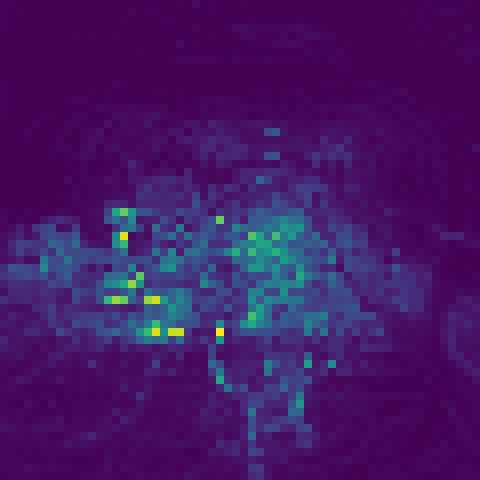} &&&
\includegraphics[width=0.07\linewidth]{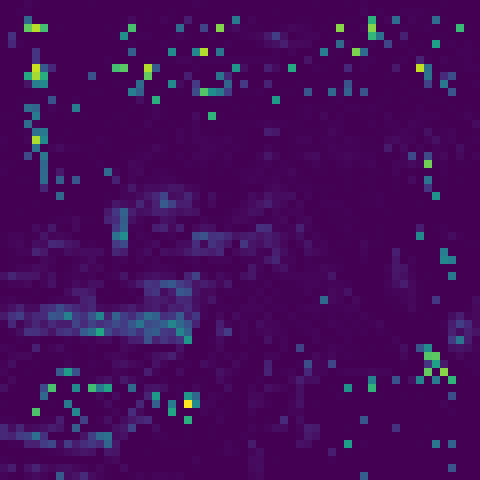} &
\includegraphics[width=0.07\linewidth]{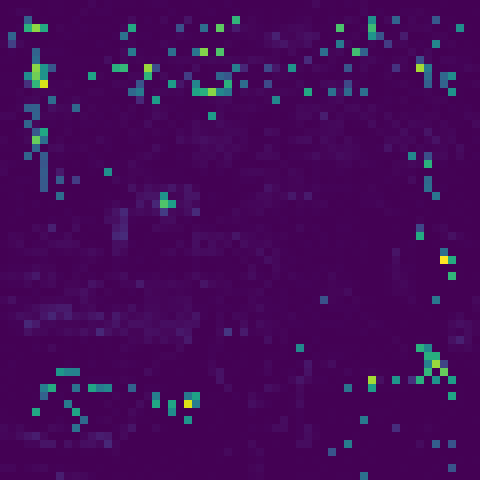} &
\includegraphics[width=0.07\linewidth]{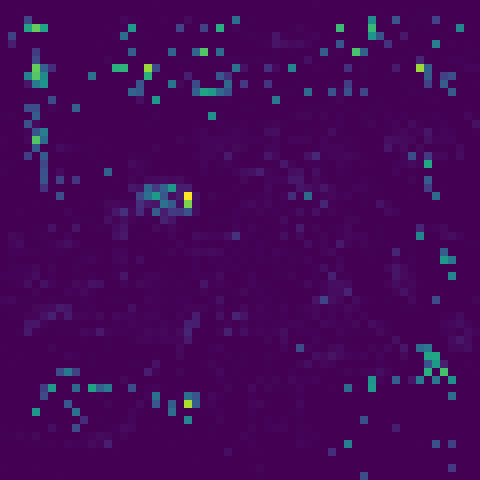}
&&&
\includegraphics[width=0.07\linewidth]{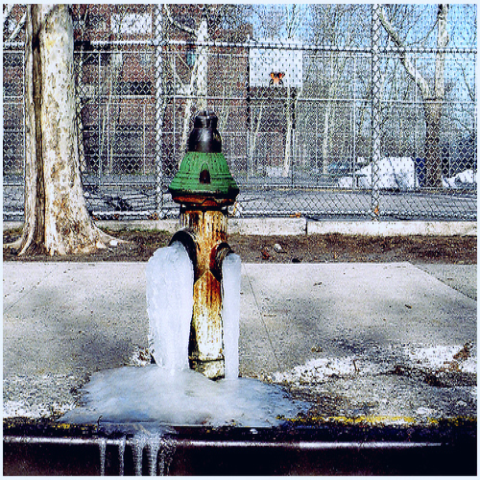} &&&
\includegraphics[width=0.07\linewidth]{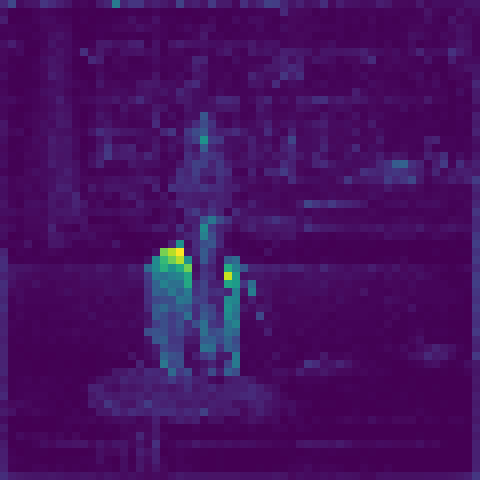} &
\includegraphics[width=0.07\linewidth]{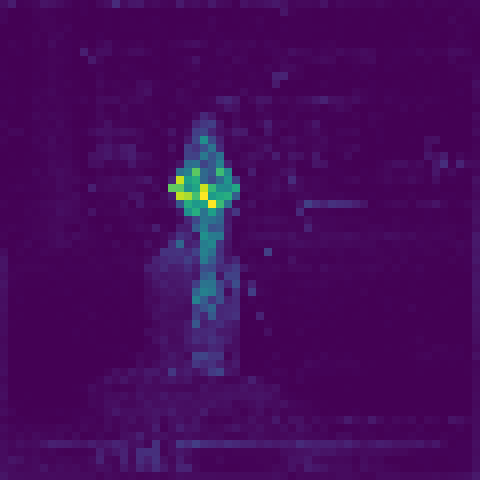} &
\includegraphics[width=0.07\linewidth]{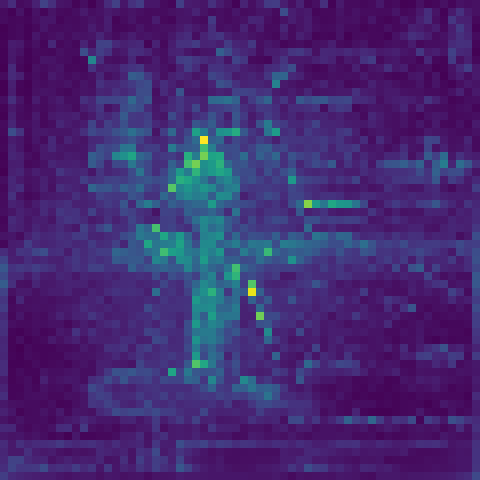} &&&
\includegraphics[width=0.07\linewidth]{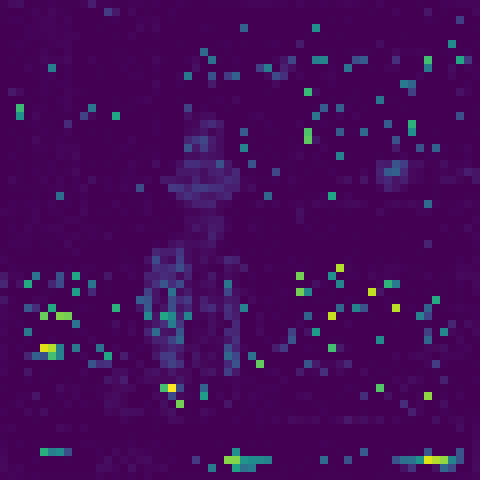} &
\includegraphics[width=0.07\linewidth]{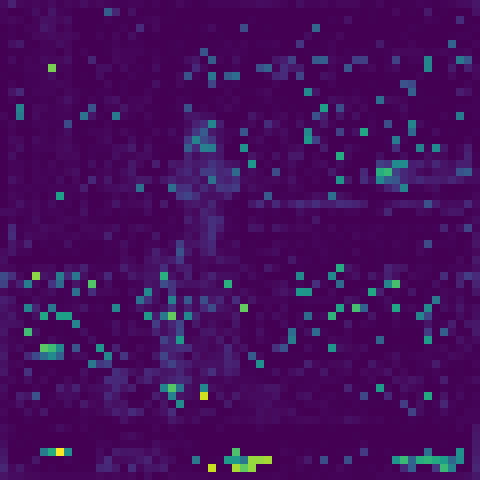} &
\includegraphics[width=0.07\linewidth]{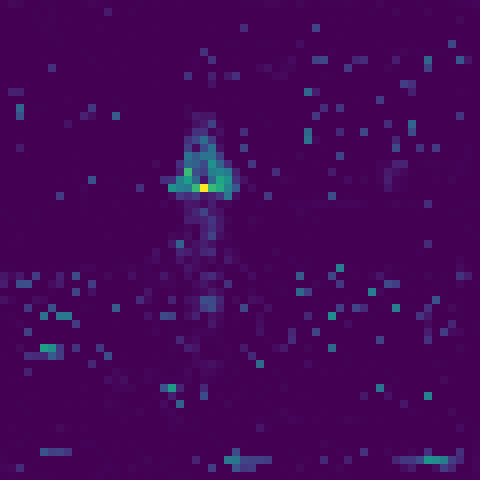}
\\
\includegraphics[width=0.07\linewidth]{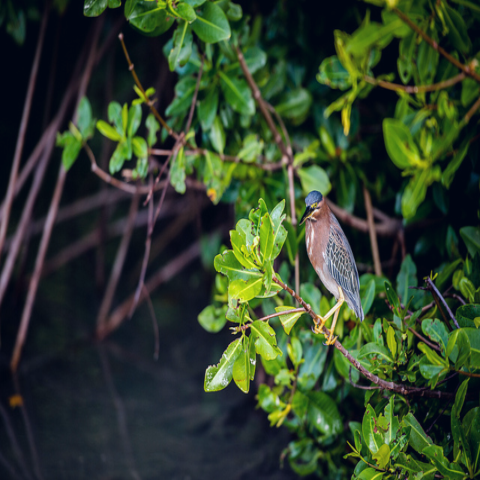} &&&
\includegraphics[width=0.07\linewidth]{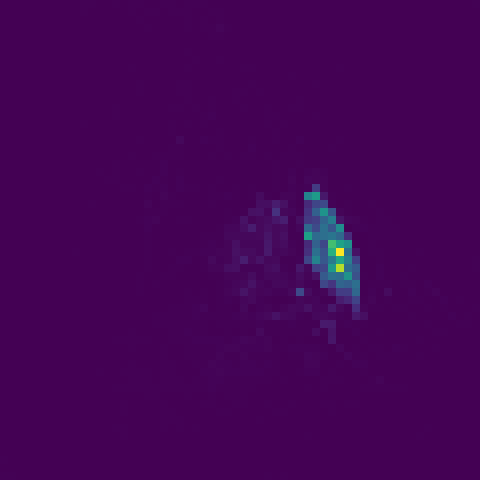} &
\includegraphics[width=0.07\linewidth]{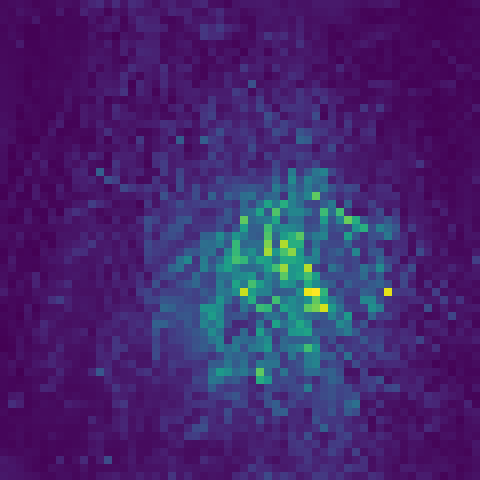} &
\includegraphics[width=0.07\linewidth]{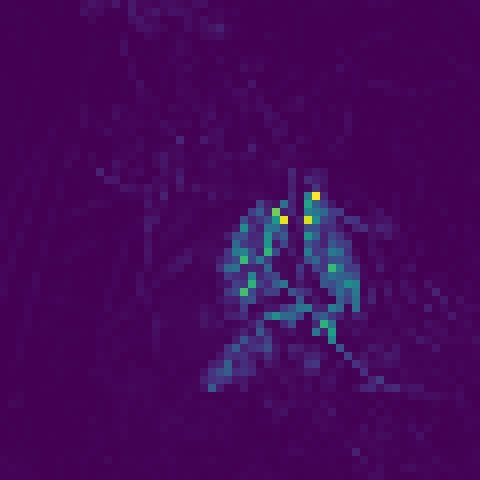} &&&
\includegraphics[width=0.07\linewidth]{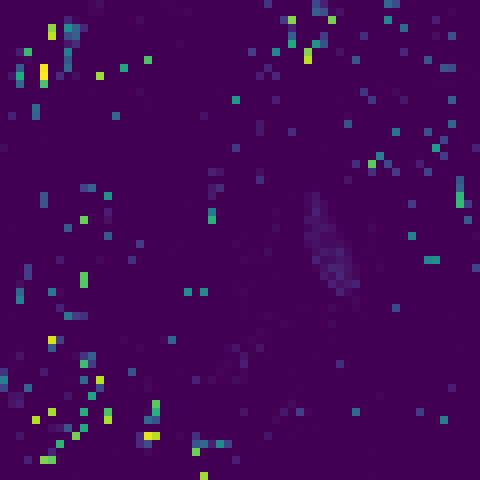} &
\includegraphics[width=0.07\linewidth]{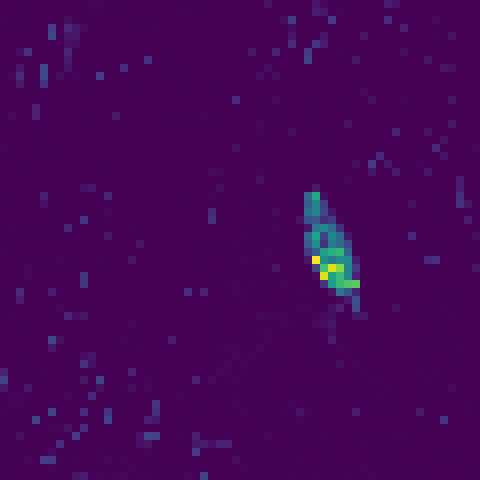} &
\includegraphics[width=0.07\linewidth]{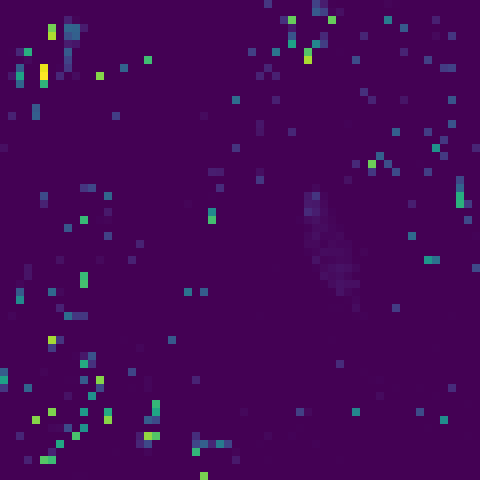}
&&&
\includegraphics[width=0.07\linewidth]{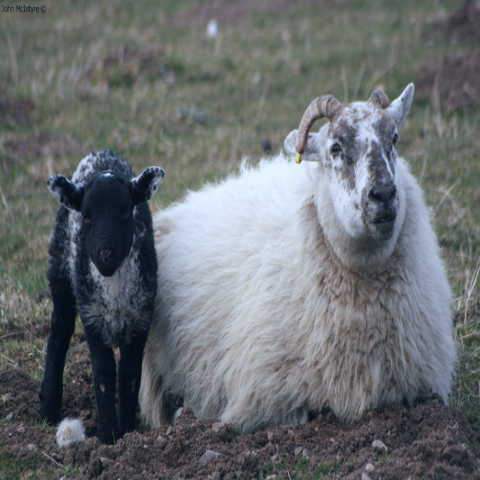} &&&
\includegraphics[width=0.07\linewidth]{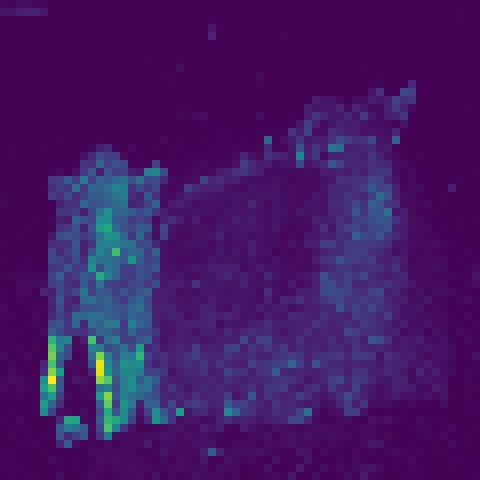} &
\includegraphics[width=0.07\linewidth]{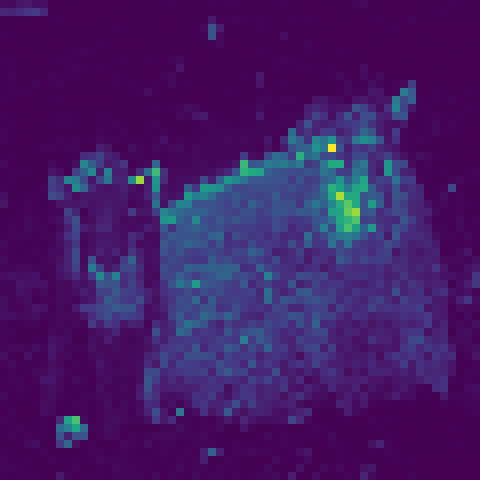} &
\includegraphics[width=0.07\linewidth]{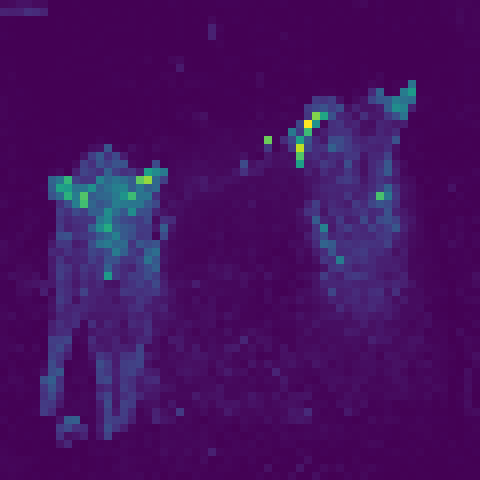} &&&
\includegraphics[width=0.07\linewidth]{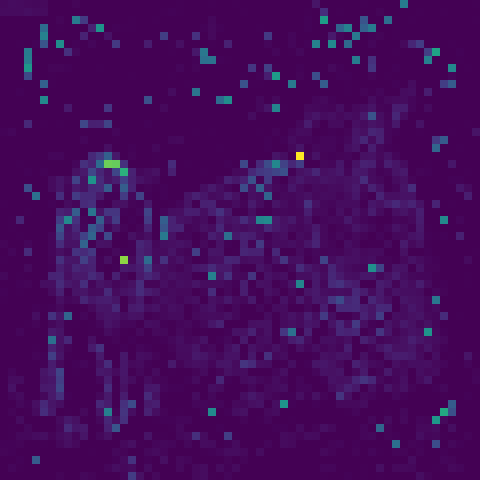} &
\includegraphics[width=0.07\linewidth]{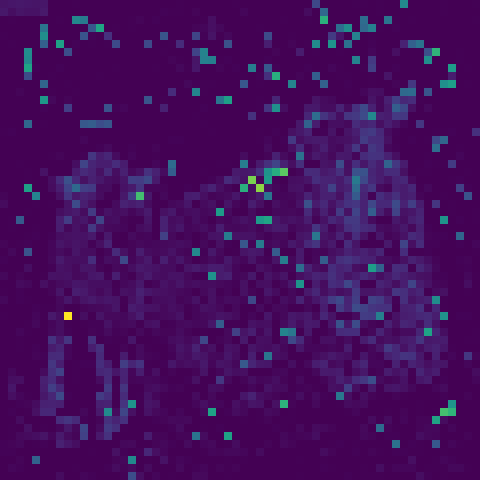} &
\includegraphics[width=0.07\linewidth]{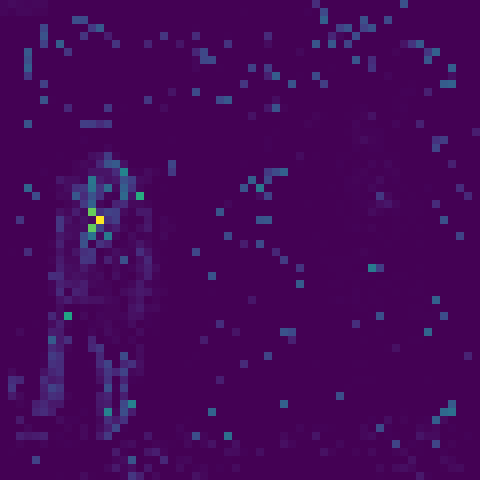}
\\
\includegraphics[width=0.07\linewidth]{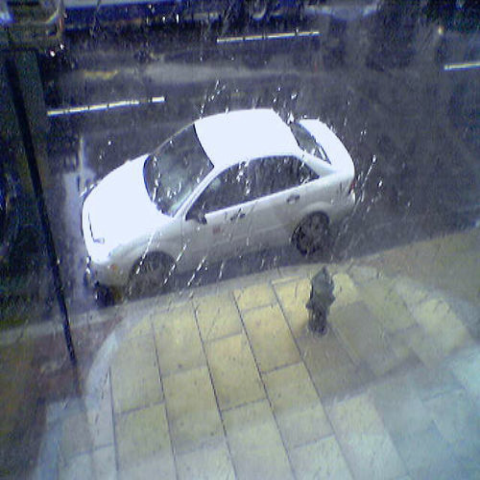} &&&
\includegraphics[width=0.07\linewidth]{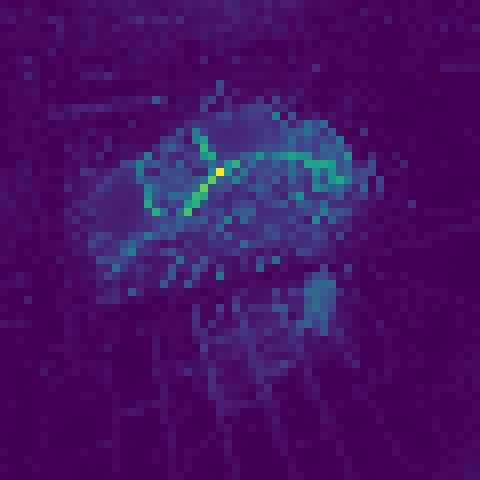} &
\includegraphics[width=0.07\linewidth]{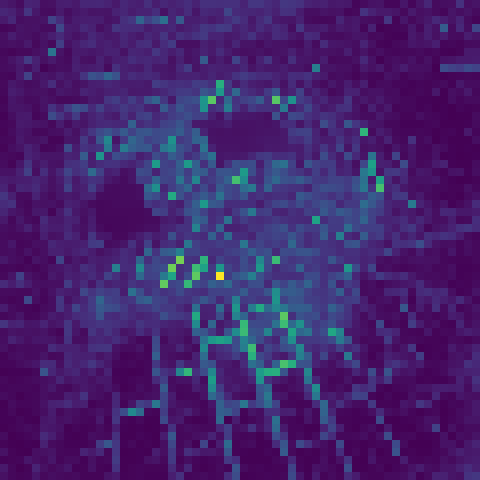} &
\includegraphics[width=0.07\linewidth]{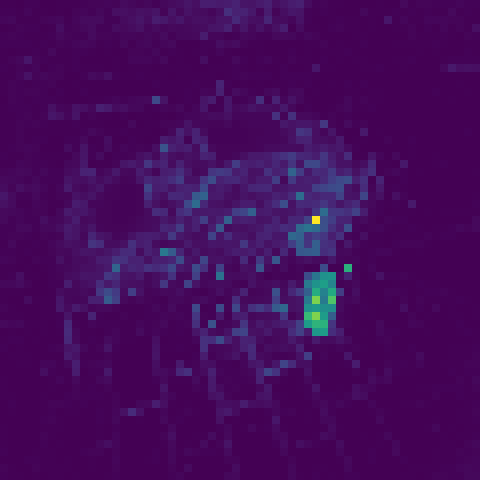} &&&
\includegraphics[width=0.07\linewidth]{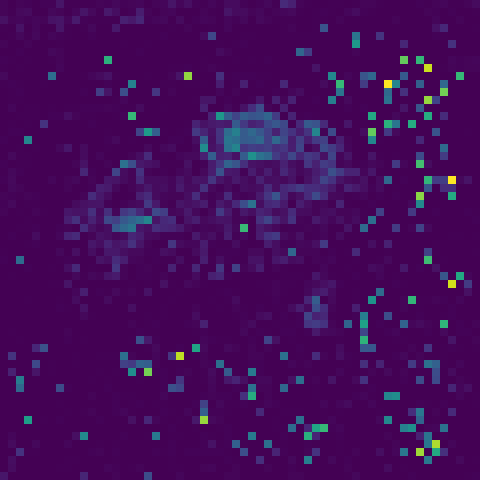} &
\includegraphics[width=0.07\linewidth]{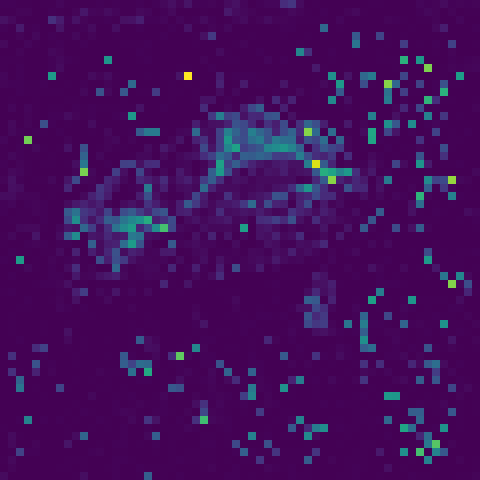} &
\includegraphics[width=0.07\linewidth]{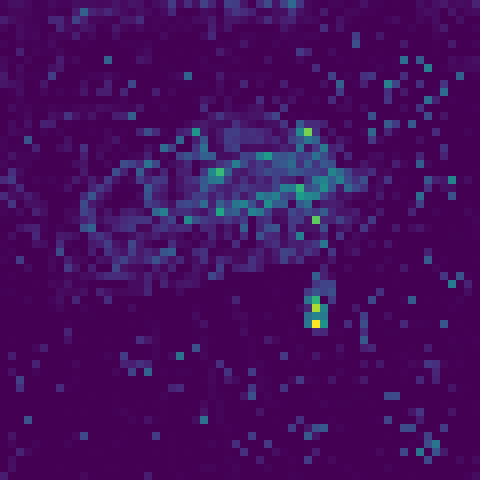}
&&&
\includegraphics[width=0.07\linewidth]{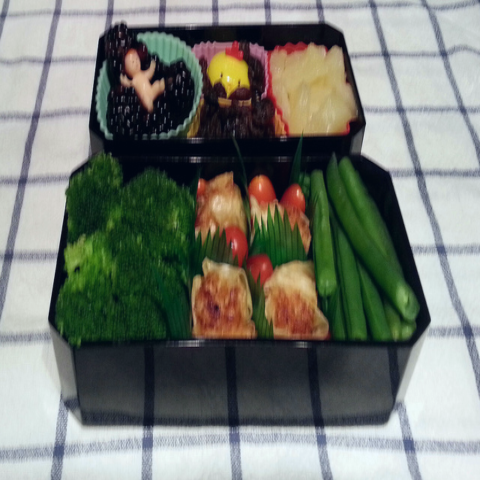} &&&
\includegraphics[width=0.07\linewidth]{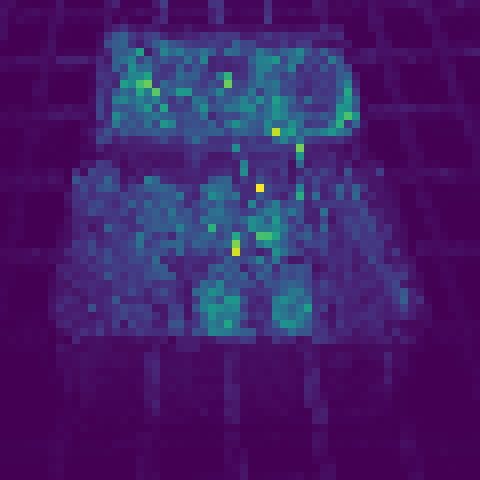} &
\includegraphics[width=0.07\linewidth]{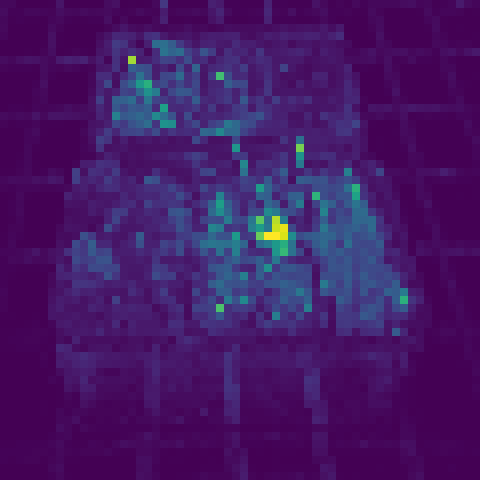} &
\includegraphics[width=0.07\linewidth]{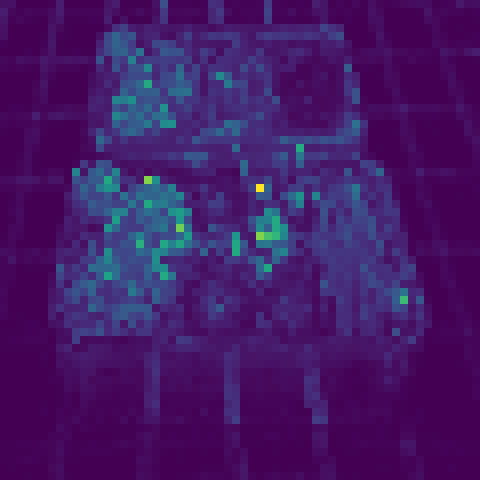} &&&
\includegraphics[width=0.07\linewidth]{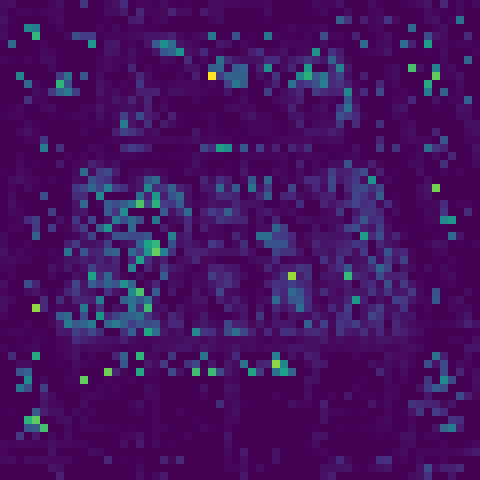} &
\includegraphics[width=0.07\linewidth]{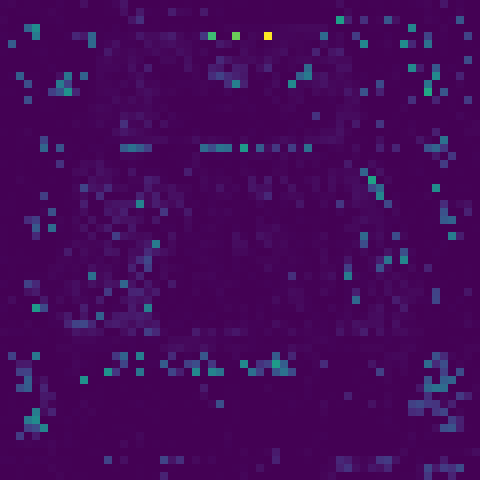} &
\includegraphics[width=0.07\linewidth]{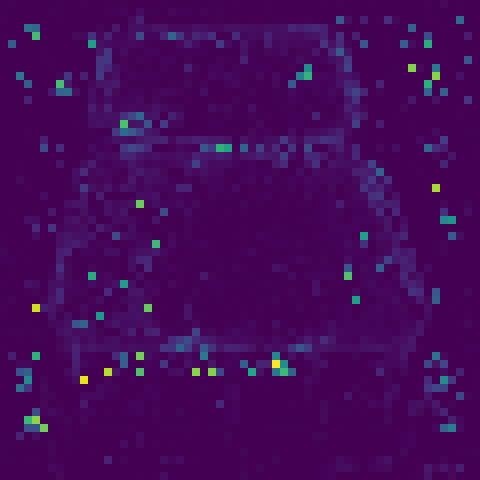}
\\
\includegraphics[width=0.07\linewidth]{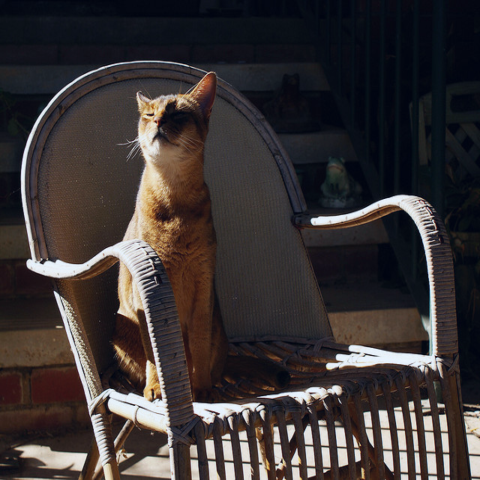} &&&
\includegraphics[width=0.07\linewidth]{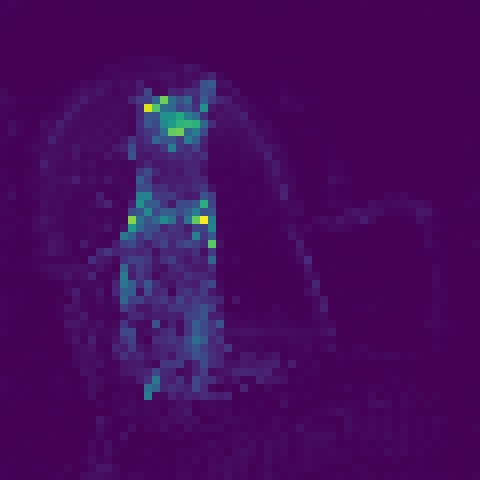} &
\includegraphics[width=0.07\linewidth]{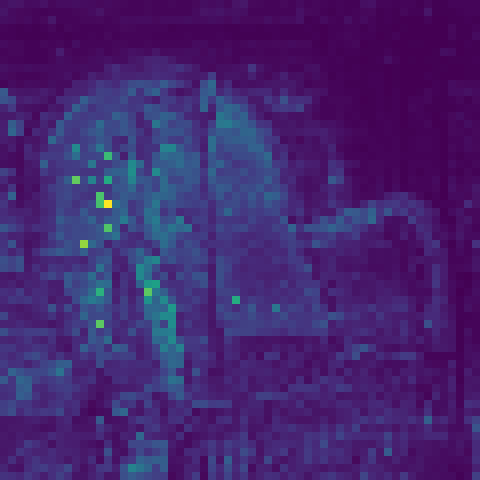} &
\includegraphics[width=0.07\linewidth]{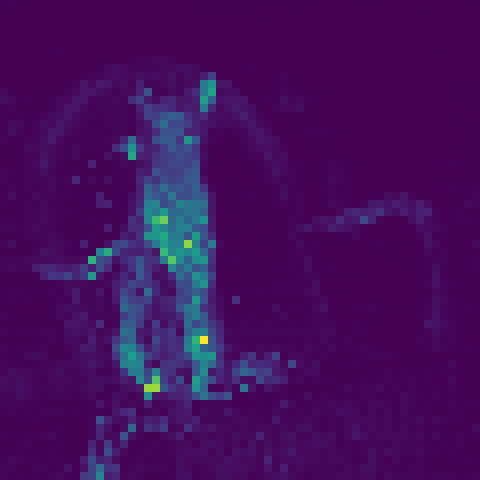} &&&
\includegraphics[width=0.07\linewidth]{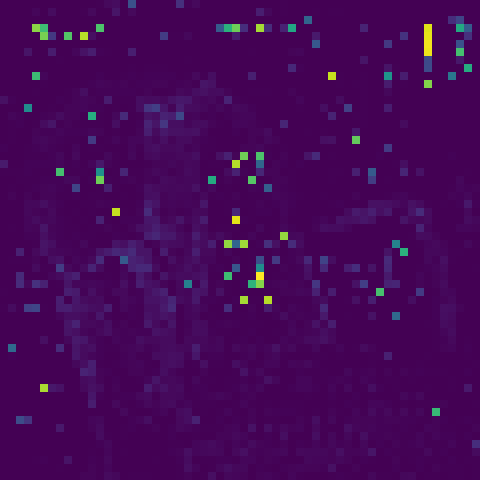} &
\includegraphics[width=0.07\linewidth]{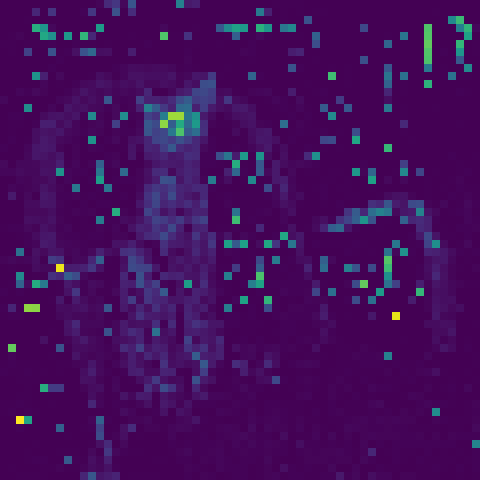} &
\includegraphics[width=0.07\linewidth]{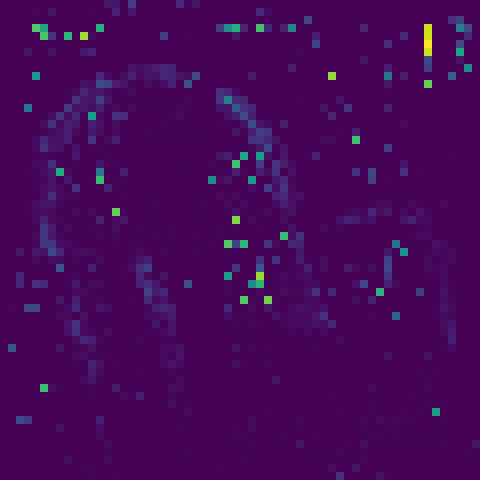}
&&&
\includegraphics[width=0.07\linewidth]{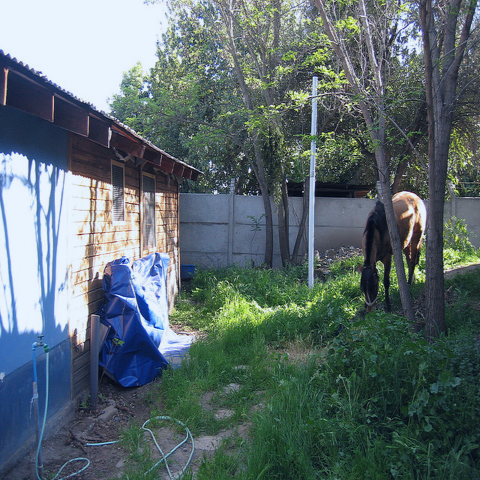} &&&
\includegraphics[width=0.07\linewidth]{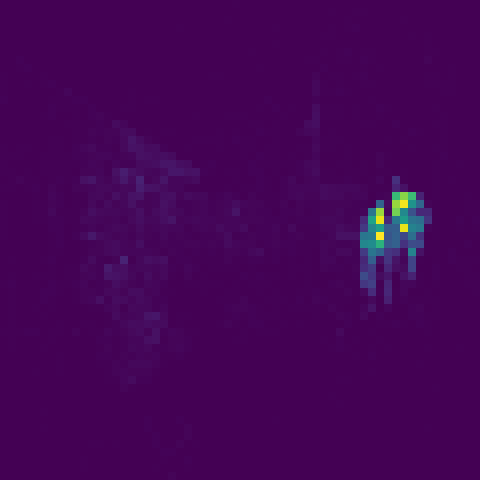} &
\includegraphics[width=0.07\linewidth]{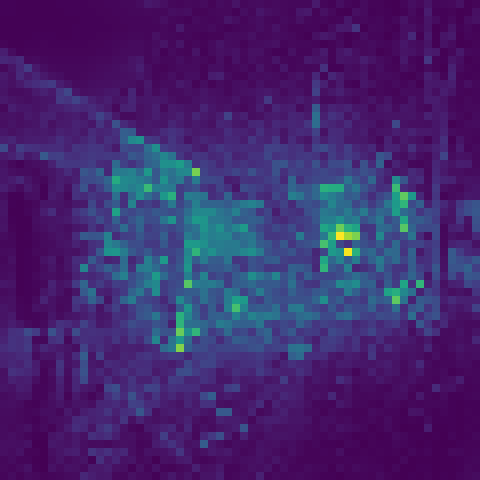} &
\includegraphics[width=0.07\linewidth]{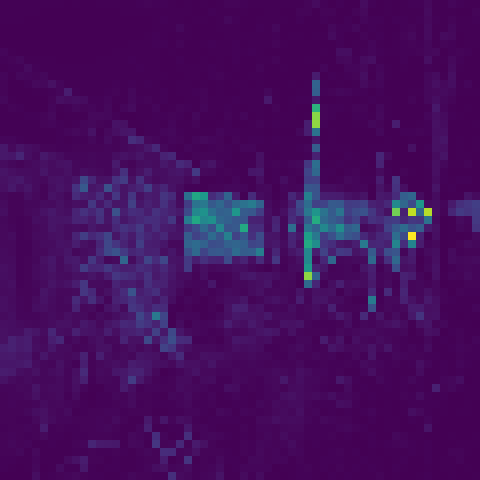} &&&
\includegraphics[width=0.07\linewidth]{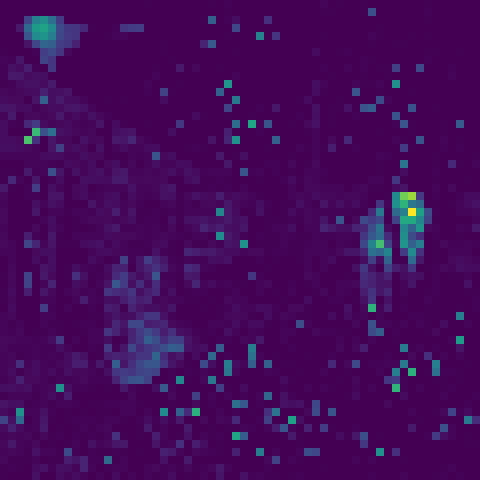} &
\includegraphics[width=0.07\linewidth]{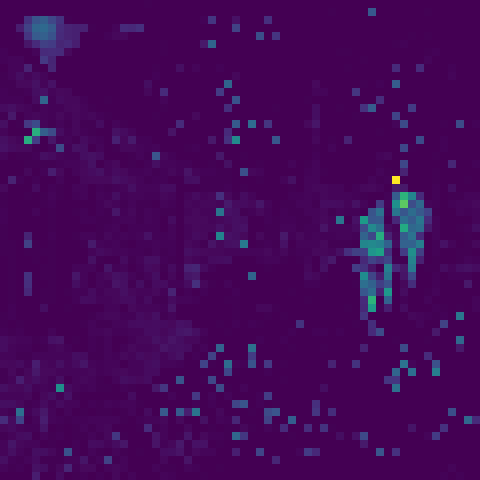} &
\includegraphics[width=0.07\linewidth]{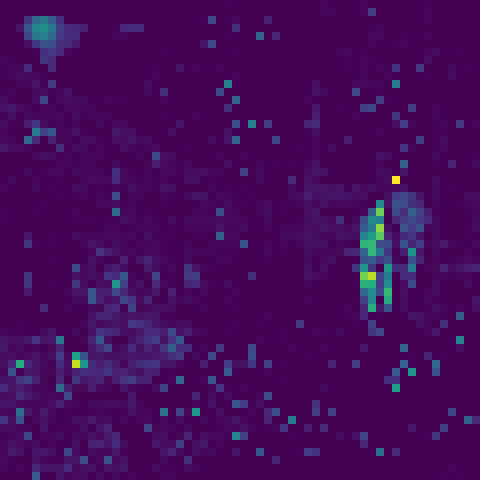}
\\
\includegraphics[width=0.07\linewidth]{3224.png} &&&
\includegraphics[width=0.07\linewidth]{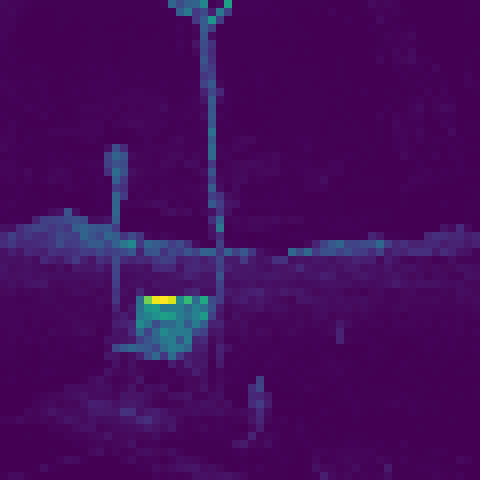} &
\includegraphics[width=0.07\linewidth]{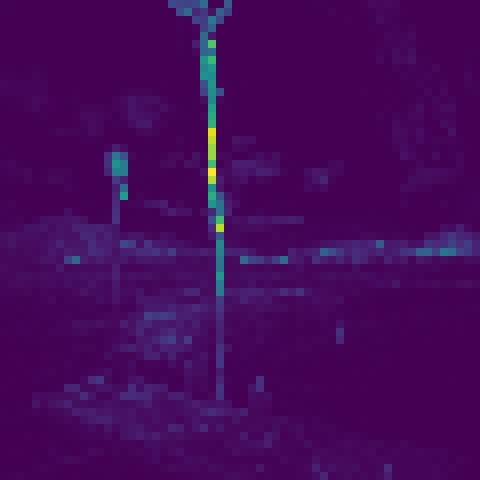} &
\includegraphics[width=0.07\linewidth]{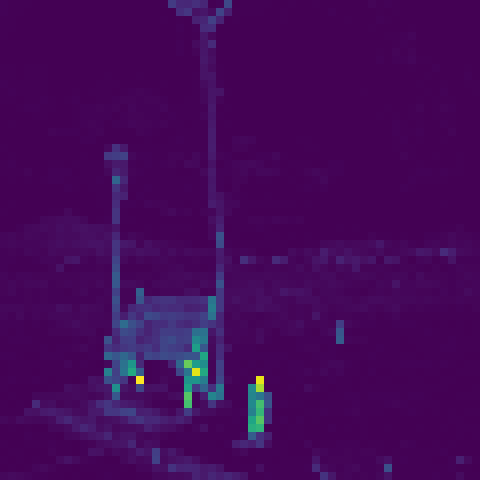} &&&
\includegraphics[width=0.07\linewidth]{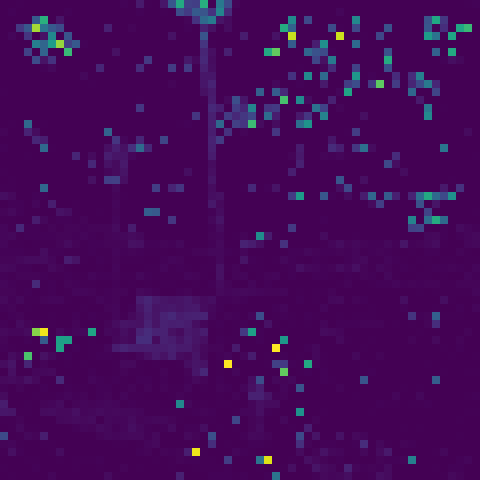} &
\includegraphics[width=0.07\linewidth]{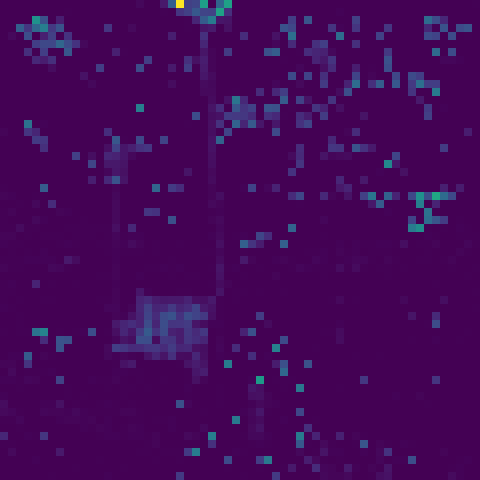} &
\includegraphics[width=0.07\linewidth]{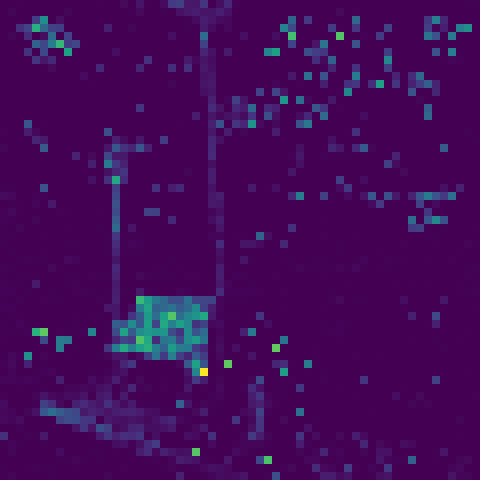}
&&&
\includegraphics[width=0.07\linewidth]{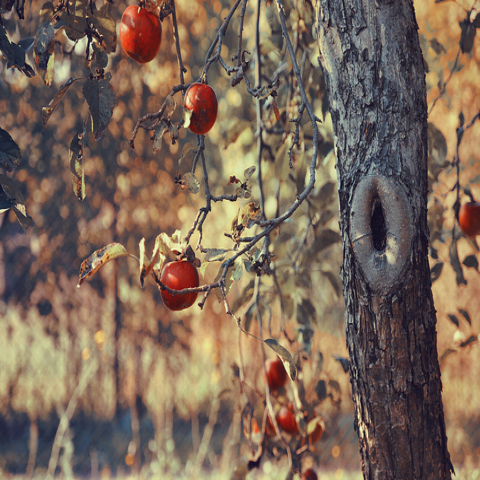} &&&
\includegraphics[width=0.07\linewidth]{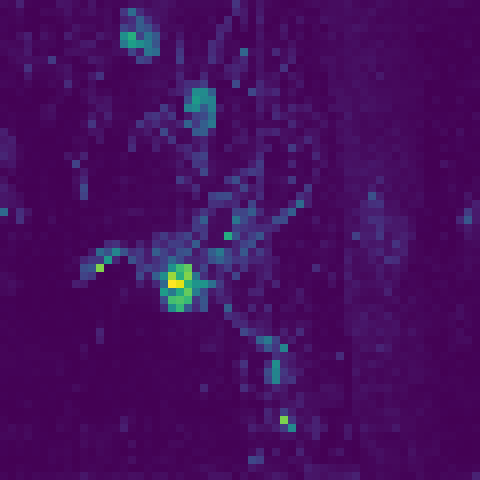} &
\includegraphics[width=0.07\linewidth]{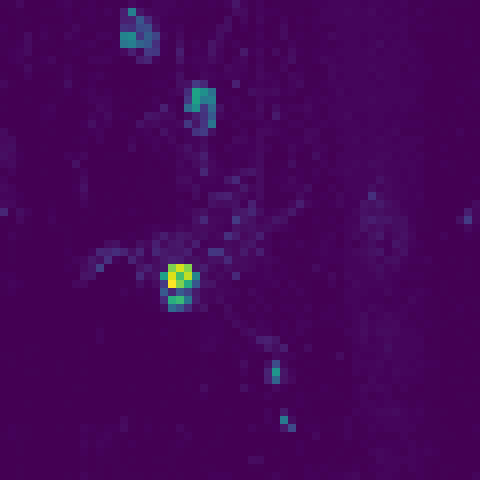} &
\includegraphics[width=0.07\linewidth]{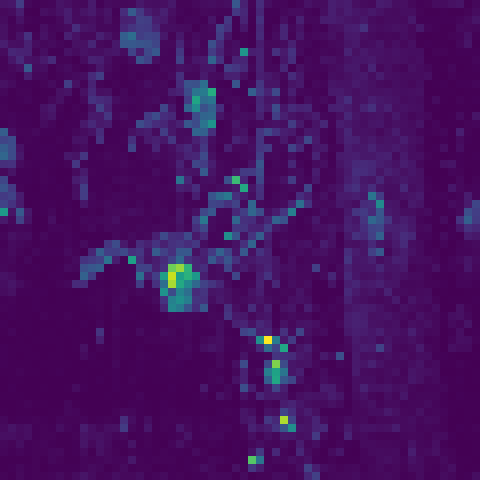} &&&
\includegraphics[width=0.07\linewidth]{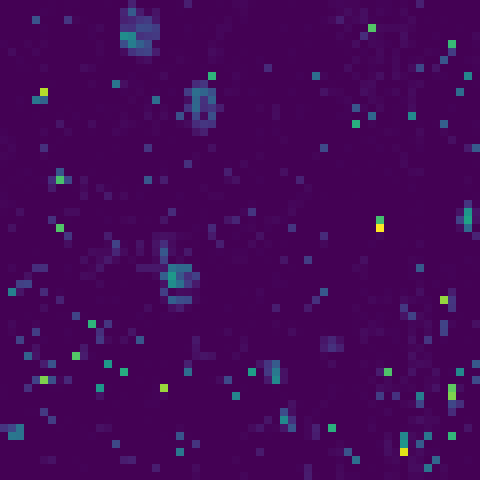} &
\includegraphics[width=0.07\linewidth]{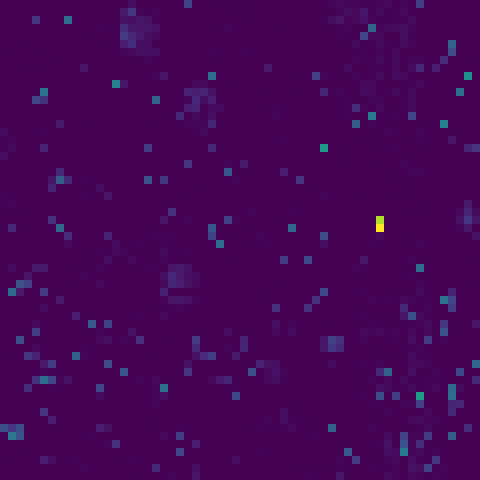} &
\includegraphics[width=0.07\linewidth]{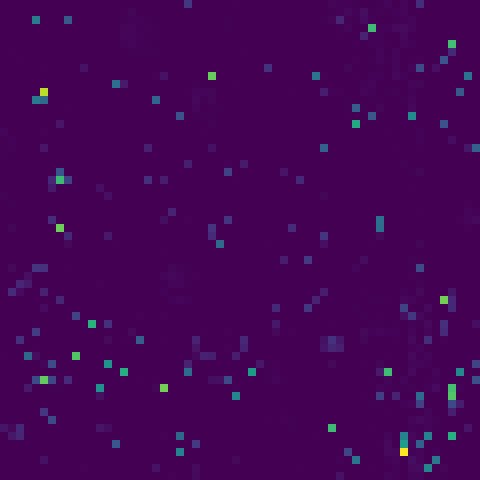}
\\
\includegraphics[width=0.07\linewidth]{1525.png} &&&
\includegraphics[width=0.07\linewidth]{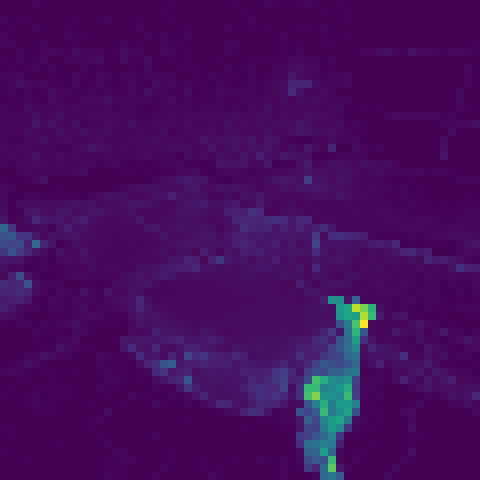} &
\includegraphics[width=0.07\linewidth]{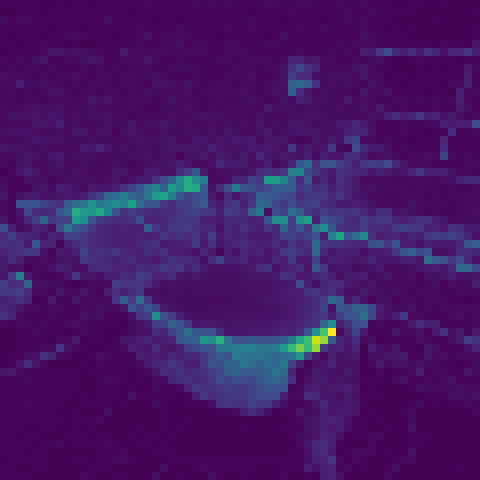} &
\includegraphics[width=0.07\linewidth]{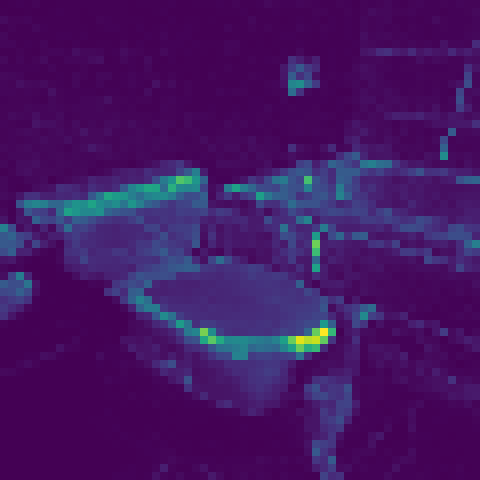} &&&
\includegraphics[width=0.07\linewidth]{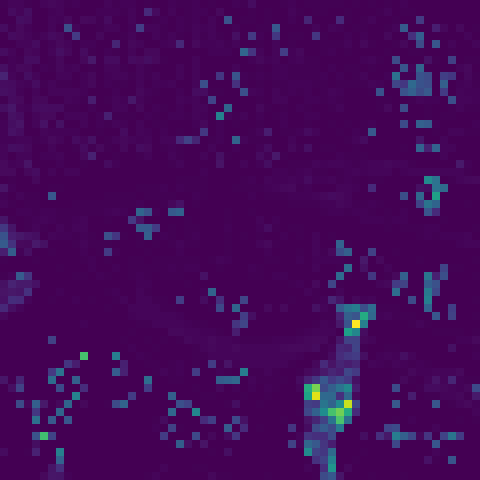} &
\includegraphics[width=0.07\linewidth]{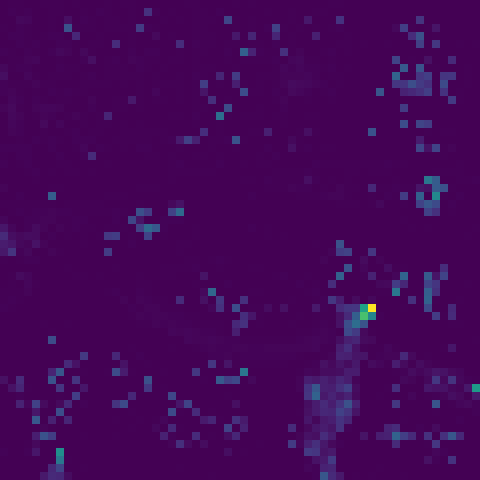} &
\includegraphics[width=0.07\linewidth]{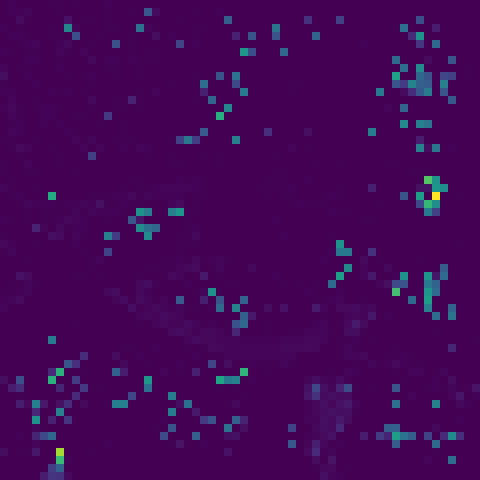}
&&&
\includegraphics[width=0.07\linewidth]{4859.png} &&&
\includegraphics[width=0.07\linewidth]{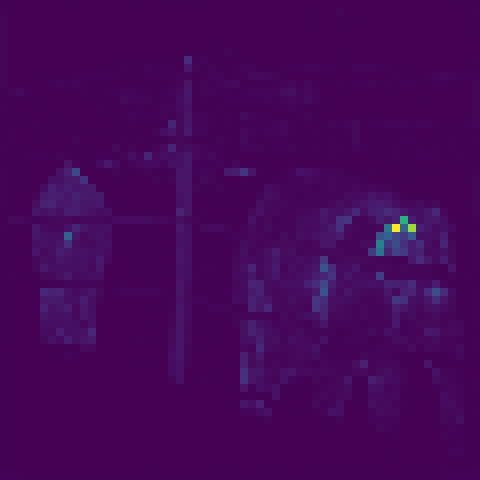} &
\includegraphics[width=0.07\linewidth]{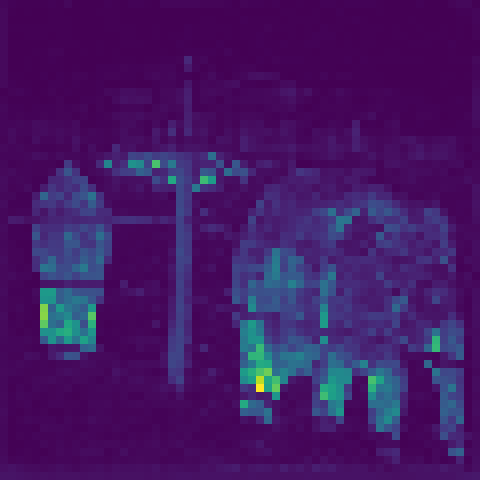} &
\includegraphics[width=0.07\linewidth]{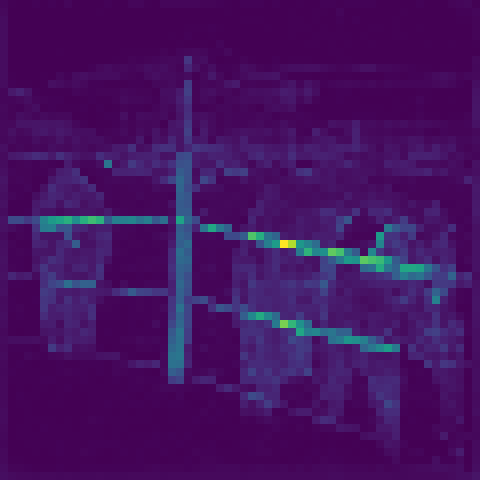} &&&
\includegraphics[width=0.07\linewidth]{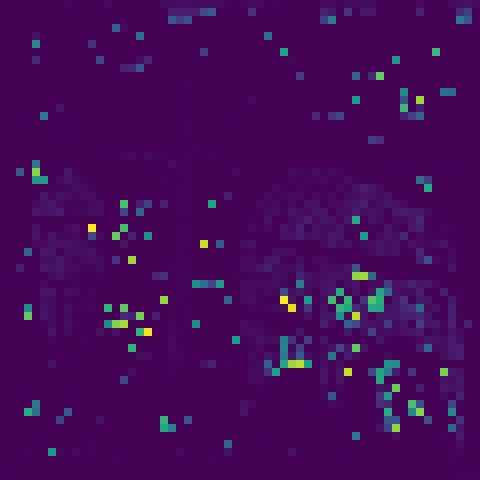} &
\includegraphics[width=0.07\linewidth]{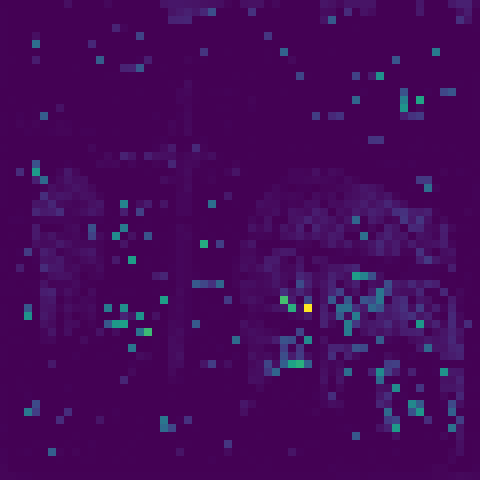} &
\includegraphics[width=0.07\linewidth]{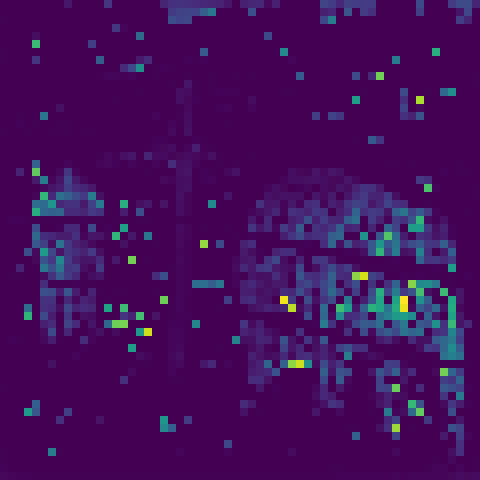}
\end{tabular}
	\caption{\textbf{Self-attention heads from the last layer.} We look at the attention map when using the \texttt{[CLS]} token as a query for the different heads in the last layer. Note that the \texttt{[CLS]} token is not attached to any label or supervision.}
\label{fig:all}
\end{figure*}

We provide more self-attention visualizations in Fig.~\ref{fig:pointing} and in Fig.~\ref{fig:all}.
The images are randomly selected from COCO validation set, and are not used during training of \OURS.
In Fig.~\ref{fig:pointing}, we show the self-attention from the last layer of a \OURS ViT-S/8 for several reference points.

\subsection{Class Representation}
\begin{figure*}[t]
\centering
\includegraphics[width=\linewidth]{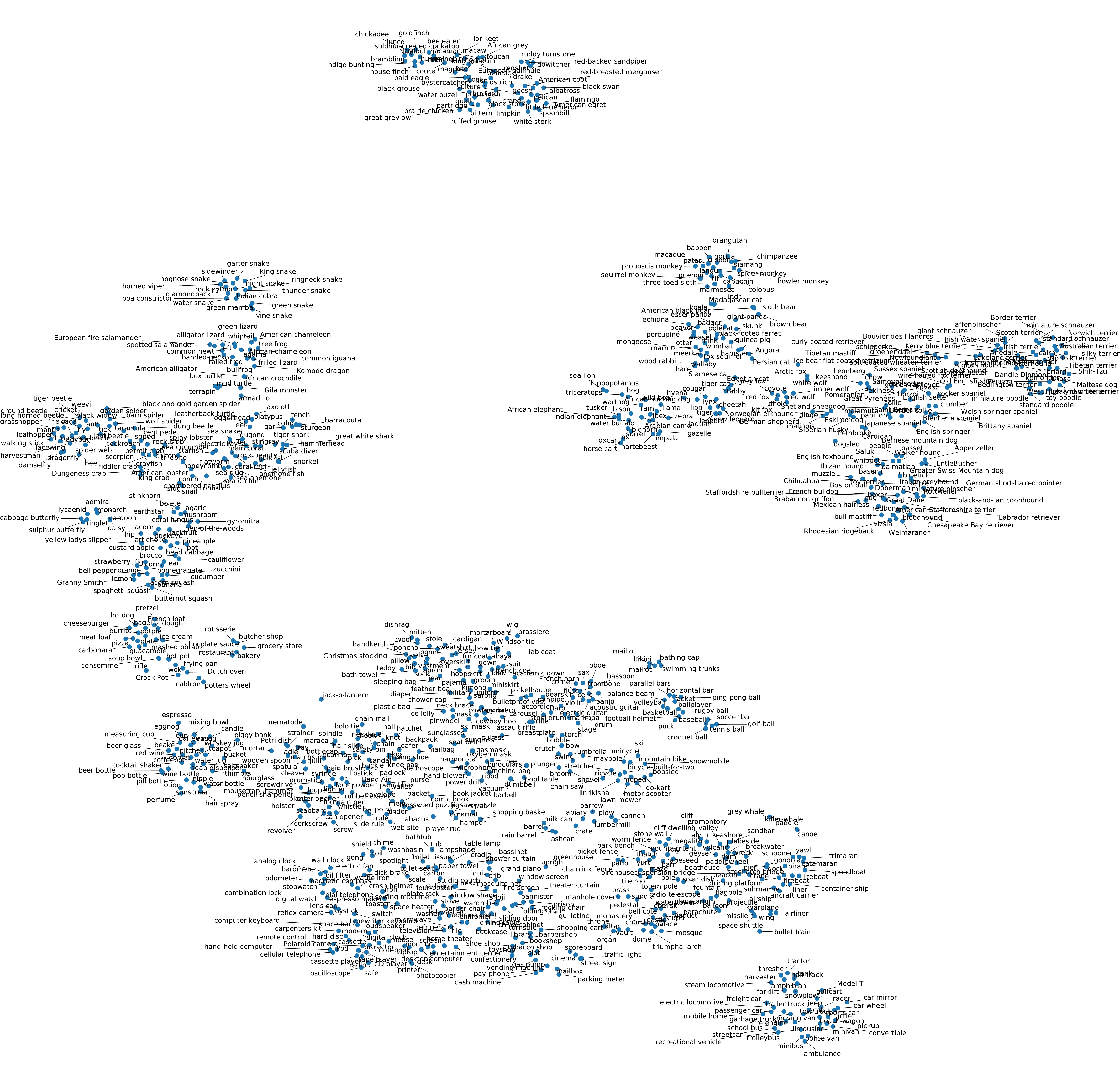}
  \caption{
    t-SNE visualization of ImageNet classes as represented using \OURS.
    For each class, we obtain the embedding by taking the average feature for all images of that class in the validation set.
  }
  \label{fig:tsne}
\end{figure*}

As a final visualization, we propose to look at the distribution of ImageNet concepts in the feature space from \OURS.
We represent each ImageNet class with the average feature vector for its validation images.
We reduce the dimension of these features to 30 with PCA, and run t-SNE with a perplexity of 20, a learning rate of 200 for 5000 iterations.
We present the resulting class embeddings in Fig.~\ref{fig:tsne}.
Our model recovers structures between classes: similar animal species are grouped together, forming coherent clusters of birds (top) or dogs, and especially terriers (far right).

\end{document}